\documentclass{article}

\usepackage{microtype}
\usepackage{graphicx}
\usepackage{subfigure}
\usepackage{booktabs} 

\usepackage{hyperref}

\usepackage[accepted]{icml2023}

\usepackage{amsmath}
\usepackage{amssymb}
\usepackage{mathtools}
\usepackage{amsthm}

\usepackage[capitalize,noabbrev]{cleveref}

\usepackage{multirow}

\theoremstyle{plain}
\newtheorem{theorem}{Theorem}[section]
\newtheorem{proposition}[theorem]{Proposition}
\newtheorem{lemma}[theorem]{Lemma}
\newtheorem{corollary}[theorem]{Corollary}
\theoremstyle{definition}
\newtheorem{definition}[theorem]{Definition}

\theoremstyle{remark}

\DeclareMathOperator*{\argmax}{arg\,max}
\newcommand{\olsi}[1]{\,\overline{\!{#1}}}
\newcommand{\diag}{\mathop{\mathrm{diag}}\nolimits}

\usepackage[textsize=tiny]{todonotes}

\icmltitlerunning{Building Neural Networks on Matrix Manifolds: A Gyrovector Space Approach}

\begin{document}

\twocolumn[
\icmltitle{Building Neural Networks on Matrix Manifolds: A Gyrovector Space Approach}



\icmlsetsymbol{equal}{}

\begin{icmlauthorlist}
\icmlauthor{Xuan Son Nguyen}{equal,yyy}
\icmlauthor{Shuo Yang}{equal,yyy}
\end{icmlauthorlist}

\icmlaffiliation{yyy}{ETIS, UMR 8051, CY Cergy Paris Universit{\'e}, ENSEA, CNRS, Cergy, France}

\icmlcorrespondingauthor{Xuan Son Nguyen}{xuan-son.nguyen@ensea.fr}

\icmlkeywords{SPD neural networks, Grassmann neural networks, Gyrovector spaces}

\vskip 0.3in
]



\printAffiliationsAndNotice{}  

\begin{abstract}
Matrix manifolds, such as manifolds of Symmetric Positive Definite (SPD) matrices and Grassmann manifolds,
appear in many applications. 
Recently, by applying the theory of gyrogroups and gyrovector spaces that is a powerful framework for
studying hyperbolic geometry, 
some works have attempted to build principled generalizations of Euclidean neural networks on matrix manifolds. 
However, due to the lack of many concepts in gyrovector spaces for the considered manifolds, 
e.g., the inner product and gyroangles,  
techniques and mathematical tools provided by these works are still limited compared to those developed 
for studying hyperbolic geometry.
In this paper, we generalize some notions in gyrovector spaces for SPD and Grassmann manifolds,  
and propose new models and layers for building neural networks on these manifolds. 
We show the effectiveness of our approach in two applications, i.e., human action recognition and knowledge graph completion.
\end{abstract}

\section{Introduction}
\label{sec:intro}

Deep neural networks (DNNs) usually assume Euclidean geometry in their computations.  
However, in many applications, data exhibit a strongly non-Euclidean latent structure 
such as those lying on Riemannian manifolds~\cite{BronsteinGDL}. 
Therefore, a lot of effort has been put into building DNNs on Riemannian manifolds in recent years.    
The most common representation spaces are Riemannian manifolds of constant non-zero curvature, 
e.g., spherical and hyperbolic spaces~\cite{NEURIPS2018_dbab2adc,skopek2020mixedcurvature}. 
Beside having closed-form expressions for the distance function, exponential and logarithmic maps, and parallel transport that
ease the task of learning parametric models,    
such spaces also have the nice algebraic structure of gyrovector spaces~\cite{UngarHyperbolicNDim} that enables
principled generalizations of DNNs to the manifold setting~\cite{NEURIPS2018_dbab2adc}.  
Another type of Riemannian manifolds referred to as matrix manifolds~\cite{Absil2007}, 
where elements can be represented in the form of matrix arrays, are also popular in representation learning. 
Typical examples are SPD and Grassmann manifolds.  
Unlike the works in~\citet{NEURIPS2018_dbab2adc,skopek2020mixedcurvature}, 
the works in~\cite{HuangGool17,DongSPD17,HuangAAAI18,NguyenHandRecgCVPR19,NguyenHandRecgFG19,NguyenHandRecgICPR20,Nguyen_2021_ICCV,WangSymNet21} 
approach the problem of generalizing DNNs to the considered manifolds in an unprincipled way. 
This makes it hard for them to generalize a broad class of DNNs to these manifolds.     
Another line of research~\cite{ChakrabortyYZBA18,ChakrabortyManifoldNet20,WeilerCNNRiemannian21,BanerjeeVolterraNetTPAMI22,XuUnifiedFourrierEquiHomoSpace22} that proposes analogs of convolutional neural networks (CNNs) on Riemannian manifolds 
relies on the notion of equivariant neural networks~\cite{BronsteinGDL}.
However, these works mainly focus on building CNNs where many essential building blocks of DNNs 
for solving a wide range of problems are missing. 

Recently, 
some works~\cite{KimSYMGyroOrder20,NguyenNeurIPS22} have attempted to explore analogies that SPD and 
Grassmann manifolds share with Euclidean and hyperbolic spaces.  
Although these works show how to construct some basic operations, e.g., the binary operation and scalar multiplication 
from the Riemannian geometry of the considered manifolds, 
it might be difficult to obtain closed-form expressions for such operations 
even if closed-form expressions for the exponential and logarithmic maps, and the parallel transport exist. 
This is the case of Grassmann manifolds from the ONB (orthonormal basis) perspective~\cite{Edelman98,bendokat2020grassmann},
where the expressions of the exponential map and parallel transport are based on singular value decomposition (SVD).  
In some applications, it is more advantageous~\cite{bendokat2020grassmann} to represent points on Grassmann manifolds 
with orthogonal matrices (the ONB perspective) 
than with projection matrices. 
Furthermore, due to the lack of some concepts in gyrovector spaces for the considered matrix manifolds, 
e.g., the inner product and gyroangles~\cite{UngarHyperbolicNDim},  
it is not trivial for these works to generalize many traditional machine learning models, e.g., 
multinomial logistic regression (MLR) to the considered manifolds.   

In this paper, we propose a new method for constructing the basic operations and gyroautomorphism of 
Grassmann manifolds from the ONB perspective.
We also improve existing works by generalizing some notions in gyrovector spaces for SPD and Grassmann manifolds. 
This leads to the development of 
MLR on SPD manifolds and isometric models on SPD and Grassmann manifolds, which we refer to as  
SPD and Grassmann gyroisometries. 
These are the counterparts of Euclidean isometries on SPD and Grassmann manifolds. 
Our motivation for studying such isometries is that they can be seen as transformations that deform the manifold without affecting its local structure. 
In the context of geometric deep learning (GDL)~\cite{BronsteinGDL} on manifolds, one aims to construct functions acting on signals defined on a manifold that are invariant to isometries. This invariance property, referred to as geometric stability, is one of the geometric principles in GDL for learning stable representations of high-dimensional data. 
Therefore, the characterization of SPD and Grassmann gyroisometries is one of the first steps to build such functions (e.g., neural networks) on the considered manifolds.

\section{Proposed Approach}
\label{sec:proposed_approach}

\subsection{Notations}
\label{subsec:notations}

We adopt the notations used in~\citet{NguyenNeurIPS22}. 
Let $\mathcal{M}$ be a homogeneous Riemannian manifold, 
$T_{\mathbf{P}}\mathcal{M}$ be the tangent space of $\mathcal{M}$ at $\mathbf{P} \in \mathcal{M}$.  
Denote by $\exp(\mathbf{P})$ and $\log(\mathbf{P})$ the usual matrix exponential and logarithm of $\mathbf{P}$,           
$\operatorname{Exp}_{\mathbf{P}}(\mathbf{W})$ the exponential map at $\mathbf{P}$ that associates to 
a tangent vector $\mathbf{W} \in T_{\mathbf{P}}\mathcal{M}$ 
a point of $\mathcal{M}$, $\operatorname{Log}_{\mathbf{P}}(\mathbf{Q})$ the logarithmic map of 
$\mathbf{Q} \in \mathcal{M}$ at $\mathbf{P}$,             
$\mathcal{T}_{\mathbf{P} \rightarrow \mathbf{Q}}(\mathbf{W})$ the parallel transport of
$\mathbf{W}$ from $\mathbf{P}$ to $\mathbf{Q}$ along geodesics connecting 
$\mathbf{P}$ and $\mathbf{Q}$,  
$D \phi_{\mathbf{P}}(\mathbf{W})$ the directional derivative of map $\phi$ at point $\mathbf{P}$ 
along direction $\mathbf{W}$.  
Denote by $\operatorname{M}_{n,m}$ the space of $n \times m$ matrices, 
$\operatorname{Sym}^{+}_n$ the space of $n \times n$ SPD matrices, 
$\operatorname{Sym}_n$ the space of $n \times n$ symmetric matrices, 
$\operatorname{Gr}_{n,p}$ the $p$-dimensional subspaces of $\mathbb{R}^n$ from the projector perspective~\cite{bendokat2020grassmann}.  
For clarity of presentation, let $\widetilde{\operatorname{Gr}}_{n,p}$ be the $p$-dimensional 
subspaces of $\mathbb{R}^n$ from the ONB perspective.    
We will use superscripts for the exponential and logarithmic maps, 
and the parallel transport to indicate their associated Riemannian metric (in the case of SPD manifolds) 
or the considered manifold (in the case of Grassmann manifolds). 
Other notations will be introduced in appropriate paragraphs of the paper.           

\subsection{Gyrovector Spaces Induced by Isometries}
\label{diffeo_gyrovector_spaces}

In this section, we study the connections of the basic operations and gyroautomorphisms of two homogeneous Riemannian manifolds 
that are related by an isometry. A review of gyrogroups and gyrovector spaces is given in Appendix~\ref{sec:gyrogroups_gyrovector_spaces}.

Let $M$ and $N$ be two homogeneous Riemannian manifolds.   
Assuming that there exists a bijective isometry between the two manifolds
\begin{equation*}
\phi: M \rightarrow N.
\end{equation*}
Assuming in addition that one can construct the binary operations, scalar multiplications, 
and gyroautomorphisms for the two manifolds 
that verify the axioms of gyrovector spaces from the following equations~\cite{NguyenNeurIPS22}: 
\begin{equation}\label{eq:matrix_matrix_addition}
\mathbf{P} \oplus \mathbf{Q} = \operatorname{Exp}_{\mathbf{P}}(\mathcal{T}_{\mathbf{I} \rightarrow \mathbf{P}}(\operatorname{Log}_{\mathbf{I}}(\mathbf{Q}))), 
\end{equation}
\begin{equation}\label{eq:scalar_matrix_multiplication}
t \otimes \mathbf{P} = \operatorname{Exp}_{\mathbf{I}}(t\operatorname{Log}_{\mathbf{I}} (\mathbf{P})),  
\end{equation}
\begin{equation}\label{eq:basic_gyroautomorphism}
\operatorname{gyr}[\mathbf{P},\mathbf{Q}]\mathbf{R} = \big( \ominus (\mathbf{P} \oplus \mathbf{Q}) \big) \oplus \big( \mathbf{P} \oplus (\mathbf{Q} \oplus \mathbf{R}) \big), 
\end{equation}
where $\mathbf{P},\mathbf{Q}$, and $\mathbf{R}$ are three points on the considered manifold, 
$\mathbf{I}$ is the identity element of the manifold, 
$t \in \mathbb{R}$, 
$\oplus$, $\otimes$, and $\operatorname{gyr}[.,.]$ denote respectively the binary operation, scalar multiplication, 
and gyroautomorphism of the manifold, 
$\ominus \mathbf{Y}$ denotes the left inverse of any point $\mathbf{Y}$ on the manifold 
such that $\ominus \mathbf{Y} \oplus \mathbf{Y} = e$, $e$ is the left identity of the corresponding gyrovector space.    

With a slight abuse of terminology, we will refer to manifolds $M$ and $N$ as gyrovector spaces. 
Finally, assuming that $\bar{\mathbf{I}}$ and $\phi(\bar{\mathbf{I}})$ are respectively the identity elements 
of manifolds $M$ and $N$ and that $\bar{\mathbf{I}}$ and $\phi(\bar{\mathbf{I}})$ are respectively the left identities 
of gyrovector spaces $M$ and $N$.  
We will study the connections between the basic operations and gyroautomorphisms of the two gyrovector spaces.  
Lemma~\ref{theorem:diffeo_binary_operation} gives such a connection for the binary operations. 
\begin{lemma}\label{theorem:diffeo_binary_operation}
Let $\mathbf{P},\mathbf{Q} \in M$. Denote by $\oplus_m$ and $\oplus_n$ the binary operations of gyrovector spaces $M$ and $N$, respectively.  
Then
\begin{equation}\label{eq:diffeo_binary_operation}
\mathbf{P} \oplus_m \mathbf{Q} = \phi^{-1}(\phi(\mathbf{P}) \oplus_n \phi(\mathbf{Q})).
\end{equation}
\end{lemma}

\paragraph{Proof} See Appendix~\ref{sec:appendix_diffeo_binary_operation}. 

Lemma~\ref{theorem:diffeo_binary_operation} states that the binary operation $\oplus_m$ can be performed
by first mapping its operands to gyrovector space $N$ via mapping $\phi(.)$, then computing the result of the binary operation $\oplus_n$ 
with the two resulting points in gyrovector space $N$, and finally returning back to the original gyrovector space $M$ 
via inverse mapping $\phi^{-1}(.)$ of $\phi(.)$. 
Similarly, Lemmas~\ref{theorem:diffeo_scalar_multiplication} and~\ref{theorem:diffeo_gyroautomorphism} 
give the connections for the scalar multiplications and gyroautomorphisms. 
\begin{lemma}\label{theorem:diffeo_scalar_multiplication}
Let $\mathbf{P} \in M$ and $t \in \mathbb{R}$. 
Denote by $\otimes_m$ and $\otimes_n$ the scalar multiplications of gyrovector spaces $M$ and $N$, respectively.  
Then
\begin{equation}\label{eq:diffeo_scalar_multiplication}
t \otimes_m \mathbf{P} = \phi^{-1}(t \otimes_n \phi(\mathbf{P})).
\end{equation}
\end{lemma}

\paragraph{Proof} See Appendix~\ref{sec:appendix_diffeo_scalar_multiplication}. 

\begin{lemma}\label{theorem:diffeo_gyroautomorphism}
Let $\mathbf{P},\mathbf{Q},\mathbf{R} \in M$. 
Denote by $\operatorname{gyr}_m[.,.]$ and $\operatorname{gyr}_n[.,.]$ the gyroautomorphisms of gyrovector spaces $M$ and $N$, respectively.  
Then
\begin{equation}\label{eq:diffeo_gyroautomorphism}
\operatorname{gyr}_m[\mathbf{P},\mathbf{Q}]\mathbf{R} = \phi^{-1}(\operatorname{gyr}_n[\phi(\mathbf{P}),\phi(\mathbf{Q})]\phi(\mathbf{R})). 
\end{equation}
\end{lemma}

\paragraph{Proof} See Appendix~\ref{sec:appendix_diffeo_gyroautomorphism}. 


The results from Lemmas~\ref{theorem:diffeo_binary_operation},~\ref{theorem:diffeo_scalar_multiplication},
~\ref{theorem:diffeo_gyroautomorphism} 
suggest an effective method for deriving closed-form expressions
of the basic operations and gyroautomorphisms for certain matrix manifolds 
(see Section~\ref{gr_gyrovector_spaces_from_diffeomorphism} 
and Appendix~\ref{sec:appendix_spd_gyrovector_spaces_from_diffeomorphism}). 
This method is supported by Theorems~\ref{theorem:gyrovector_spaces_from_diffeomorphism} 
and~\ref{theorem:gyrovector_spaces_from_diffeomorphism_grassmann}.  
\begin{theorem}\label{theorem:gyrovector_spaces_from_diffeomorphism}
Let $(G_n,\oplus_n,\otimes_n)$ be a gyrovector space.  
Let $\oplus_m$, $\otimes_m$, and $\operatorname{gyr}_m[.,.]$ be respectively the binary operation, scalar multiplication, 
and gyroautomorphism defined by Eqs.~(\ref{eq:diffeo_binary_operation}),~(\ref{eq:diffeo_scalar_multiplication}), 
and~(\ref{eq:diffeo_gyroautomorphism}) where $\phi(.)$ is a bijective isometry. 
Then $(G_m,\oplus_m,\otimes_m)$ forms a gyrovector space. 
\end{theorem}

\paragraph{Proof} See Appendix~\ref{sec:appendix_gyrovector_spaces_from_diffeomorphism}. 

A direct consequence of Theorem~\ref{theorem:gyrovector_spaces_from_diffeomorphism} follows. 
\begin{theorem}\label{theorem:gyrovector_spaces_from_diffeomorphism_grassmann}
Let $(G,\oplus_n)$ be a gyrocommutative and gyrononreductive gyrogroup.  
Let $\oplus_m$ and $\operatorname{gyr}_m[.,.]$ be respectively the binary operation and gyroautomorphism 
defined by Eqs.~(\ref{eq:diffeo_binary_operation}) and~(\ref{eq:diffeo_gyroautomorphism}) where $\phi(.)$ is a bijective isometry. 
Then $(G,\oplus_m)$ forms a gyrocommutative and gyrononreductive gyrogroup.  
\end{theorem}


\subsection{Grassmann Manifolds}
\label{subsec:grassmann_gorocommutative_gyrononreductive_gyrogroups}

We show how to construct the basic operations and
gyroautomorphism for Grassmann manifolds from the ONB perspective in Section~\ref{gr_gyrovector_spaces_from_diffeomorphism}. 
In Section~\ref{subsec:gr_isometries}, we study some isometries of Grassmann manifolds 
with respect to the canonical metric~\cite{Edelman98}.     

\subsubsection{Grassmann Gyrocommutative and Gyrononreductive Gyrogroups: The ONB Perspective}
\label{gr_gyrovector_spaces_from_diffeomorphism}

In~\citet{NguyenNeurIPS22}, closed-form expressions of the basic operations and gyroautomorphism 
for $\operatorname{Gr}_{n,p}$ have been derived. 
These can be obtained from Eqs.~(\ref{eq:matrix_matrix_addition}),~(\ref{eq:scalar_matrix_multiplication}), and~(\ref{eq:basic_gyroautomorphism}) as the exponential and logarithmic maps, and the parallel transport 
appear in closed-forms. However, the same method cannot be applied to Grassmann manifolds 
from the ONB perspective.  
This is because the exponential map and parallel transport in this case are all based on
SVD operations.         

We tackle the above problem using the following diffeomorphism~\cite{HelmkeOptiDynSystem} between $\widetilde{\operatorname{Gr}}_{n,p}$ 
and $\operatorname{Gr}_{n,p}$: 
\begin{equation*}
\tau: \widetilde{\operatorname{Gr}}_{n,p} \rightarrow \operatorname{Gr}_{n,p}, \hspace{1mm} \mathbf{U} \mapsto \mathbf{U}\mathbf{U}^T,    
\end{equation*}
where $\mathbf{U} \in \widetilde{\operatorname{Gr}}_{n,p}$. This leads to 
the following definitions. 

\begin{definition}\label{lem:matrix_matrix_addition_gr}
For $\mathbf{U},\mathbf{V} \in \widetilde{\operatorname{Gr}}_{n,p}$, 
assuming that $\mathbf{I}_{n,p}$ and $\mathbf{U}\mathbf{U}^T$ are not in each other's cut locus, 
then the binary operation $\mathbf{U} \widetilde{\oplus}_{gr} \mathbf{V}$ can be defined as
\begin{equation}\label{eq:matrix_matrix_addition_gr}
\mathbf{U} \widetilde{\oplus}_{gr} \mathbf{V} = \exp([\olsi{\mathbf{P}}, \mathbf{I}_{n,p}]) \mathbf{V},            
\end{equation}
where $\mathbf{I}_{n,p} = \begin{bmatrix} \mathbf{I}_p & 0 \\ 0 & 0 \end{bmatrix} \in M_{n,n}$ 
is the identity element of $\operatorname{Gr}_{n,p}$, 
$[.,.]$ denotes the matrix commutator, 
and $\olsi{\mathbf{P}} = \operatorname{Log}^{gr}_{\mathbf{I}_{n,p}}(\mathbf{U} \mathbf{U}^T)$ 
is the logarithmic map of $\mathbf{U} \mathbf{U}^T \in \operatorname{Gr}_{n,p}$ at $\mathbf{I}_{n,p}$.  
\end{definition}


\begin{definition}\label{lem:scalar_multiplication_gr}
For $\mathbf{U} \in \widetilde{\operatorname{Gr}}_{n,p}$ and $t \in \mathbb{R}$, 
assuming that $\mathbf{I}_{n,p}$ and $\mathbf{U}\mathbf{U}^T$ are not in each other's cut locus, 
then the scalar multiplication $t \widetilde{\otimes}_{gr} \mathbf{U}$ can be defined as
\begin{equation}\label{eq:scalar_multiplication_gr}
t \widetilde{\otimes}_{gr} \mathbf{U} = \exp([t\olsi{\mathbf{P}}, \mathbf{I}_{n,p}]) \widetilde{\mathbf{I}}_{n,p},
\end{equation}
where $\widetilde{\mathbf{I}}_{n,p} = \begin{bmatrix} \mathbf{I}_p \\ 0 \end{bmatrix} \in M_{n,p}$, and 
$\olsi{\mathbf{P}} = \operatorname{Log}^{gr}_{\mathbf{I}_{n,p}}(\mathbf{U} \mathbf{U}^T)$.          
\end{definition}


\begin{definition}\label{def:gyrovector_spaces_gr}
Define the binary operation $\widetilde{\oplus}_{gr}$ and the scalar multiplication $\widetilde{\otimes}_{gr}$ 
by Eqs.~(\ref{eq:matrix_matrix_addition_gr}) and~(\ref{eq:scalar_multiplication_gr}), respectively. 
For $\mathbf{U},\mathbf{V},\mathbf{W} \in \widetilde{\operatorname{Gr}}_{n,p}$,
assuming that 
$\mathbf{I}_{n,p}$ and $\mathbf{U}\mathbf{U}^T$ are not in each other's cut locus, 
$\mathbf{I}_{n,p}$ and $\mathbf{V}\mathbf{V}^T$ are not in each other's cut locus,
$\mathbf{I}_{n,p}$ and $\mathbf{U}\mathbf{U}^T \oplus_{gr} \mathbf{V}\mathbf{V}^T$ are not in each other's cut locus 
where $\oplus_{gr}$ is the binary operation~\cite{NguyenNeurIPS22} on $\operatorname{Gr}_{n,p}$,                 
then the gyroautomorphism generated by $\mathbf{U}$ and $\mathbf{V}$ can be defined as
\begin{equation*}
\widetilde{\operatorname{gyr}}_{gr}[\mathbf{U},\mathbf{V}]\mathbf{W} = \widetilde{F}_{gr}(\mathbf{U},\mathbf{V}) \mathbf{W},
\end{equation*}    
where $\widetilde{F}_{gr}(\mathbf{U},\mathbf{V})$ is given by 
\begin{align*}
\begin{split}
\widetilde{F}_{gr}(\mathbf{U},\mathbf{V}) = \exp(-[\olsi{\mathbf{P} \oplus_{gr} \mathbf{Q}}, \mathbf{I}_{n,p}]) \exp([\olsi{\mathbf{P}}, \mathbf{I}_{n,p}]) & \\ \exp([\olsi{\mathbf{Q}}, \mathbf{I}_{n,p}]),
\end{split}
\end{align*}
where 
$\olsi{\mathbf{P}} = \operatorname{Log}^{gr}_{\mathbf{I}_{n,p}}(\mathbf{U} \mathbf{U}^T)$,  
$\olsi{\mathbf{Q}} = \operatorname{Log}^{gr}_{\mathbf{I}_{n,p}}(\mathbf{V} \mathbf{V}^T)$, 
and $\olsi{\mathbf{P} \oplus_{gr} \mathbf{Q}} = \operatorname{Log}^{gr}_{\mathbf{I}_{n,p}}(\mathbf{U}\mathbf{U}^T \oplus_{gr} \mathbf{V}\mathbf{V}^T)$.        
\end{definition}



\subsubsection{Grassmann Gyroisometries - The Isometries of Grassmann Manifolds}
\label{subsec:gr_isometries}

Let $\ominus_{gr}$ and $\operatorname{gyr}_{gr}[.,.]$ be the inverse operation and
gyroautomorphism of $\operatorname{Gr}_{n,p}$. 
Guided by analogies with the Euclidean and hyperbolic geometries, 
we investigate in this section some isometries of Grassmann manifolds. 
First, we need to define the inner product on these manifolds. 
             

\begin{definition}[{\bf The Grassmann Inner Product}]\label{def:inner_product_grassmann_manifolds}
Let $\mathbf{P},\mathbf{Q} \in \operatorname{Gr}_{n,p}$. Then the Grassmann inner product of $\mathbf{P}$ and $\mathbf{Q}$ is defined as
\begin{align*}
\begin{split}
\langle \mathbf{P},\mathbf{Q} \rangle = \langle \operatorname{Log}^{gr}_{\mathbf{I}_{n,p}}(\mathbf{P}), \operatorname{Log}^{gr}_{\mathbf{I}_{n,p}}(\mathbf{Q}) \rangle_{\mathbf{I}_{n,p}},    
\end{split}
\end{align*}
where $\langle.,.\rangle_{\mathbf{I}_{n,p}}$ denotes the inner product at $\mathbf{I}_{n,p}$ given by the canonical metric of $\operatorname{Gr}_{n,p}$. Note that we use the notation $\langle .,. \rangle$ without subscript to denote the inner product
that is defined directly on Grassmann manifolds, and the notation $\langle .,. \rangle$ with subscript to denote the inner product
on tangent spaces of Grassmann manifolds.    
\end{definition}

The counterpart of the Euclidean distance function on $\operatorname{Gr}_{n,p}$ 
is defined below. 
\begin{definition}[{\bf The Grassmann Gyrodistance Function}]\label{def:gyrodistance_grassmann_manifolds}
Let $\mathbf{P},\mathbf{Q} \in \operatorname{Gr}_{n,p}$. Then the Grassmann gyrodistance function $d(\mathbf{P},\mathbf{Q})$ is defined as
\begin{equation*}
d(\mathbf{P},\mathbf{Q}) = \| \ominus_{gr} \mathbf{P} \oplus_{gr} \mathbf{Q} \|, 
\end{equation*}
where 
$\|.\|$ denotes the Grassmann norm induced by the Grassmann inner product given in Definition~\ref{def:inner_product_grassmann_manifolds}.  
\end{definition}

Grassmann gyroisometries now can be defined as follows. 

\begin{definition}[{\bf Grassmann Gyroisometries}]\label{def:gyroisometries_grassmann_gyrovector_spaces}
Let $\mathbf{P},\mathbf{Q} \in \operatorname{Gr}_{n,p}$.  
Then a map $\omega: \operatorname{Gr}_{n,p} \rightarrow \operatorname{Gr}_{n,p}$ is a Grassmann gyroisometry 
if it preserves the Grassmann gyrodistance between $\mathbf{P}$ and $\mathbf{Q}$, i.e., 
\begin{equation*}
d(\omega(\mathbf{P}),\omega(\mathbf{Q})) = d(\mathbf{P},\mathbf{Q}).
\end{equation*}
\end{definition}

The definitions of the Grassmann gyrodistance function and Grassmann gyroisometries
agree with those of the hyperbolic gyrodistance function and hyperbolic isometries~\cite{UngarHyperbolicNDim}.  
Theorems~\ref{theorem:left_gyrotranslational_isometry_gr},~\ref{theorem:gyroautomorphisms_gyroisometries_gr}, 
and~\ref{theorem:inverse_isometry_gr} characterize some Grassmann gyroisometries.  

\begin{theorem}\label{theorem:left_gyrotranslational_isometry_gr}
For any $\mathbf{P} \in \operatorname{Gr}_{n,p}$, a left Grassmann gyrotranslation by $\mathbf{P}$ 
is the map $\psi_{\mathbf{P}} : \operatorname{Gr}_{n,p} \rightarrow \operatorname{Gr}_{n,p}$ given by
\begin{equation*}
\psi_{\mathbf{P}}(\mathbf{Q}) = \mathbf{P} \oplus_{gr} \mathbf{Q},
\end{equation*}
where $\mathbf{Q} \in \operatorname{Gr}_{n,p}$.  
Then left Grassmann gyrotranslations are Grassmann gyroisometries.
\end{theorem}

\paragraph{Proof} See Appendix~\ref{sec:appendix_left_gyrotranslational_isometry_gr}.

\begin{theorem}\label{theorem:gyroautomorphisms_gyroisometries_gr}
Gyroautomorphisms $\operatorname{gyr}_{gr}[.,.]$ are Grassmann gyroisometries. 
\end{theorem}

\paragraph{Proof} See Appendix~\ref{sec:appendix_gyroautomorphisms_gyroisometries_gr}.

\begin{theorem}\label{theorem:inverse_isometry_gr}
A Grassmann inverse map is the map $\lambda: \operatorname{Gr}_{n,p} \rightarrow \operatorname{Gr}_{n,p}$ given by
\begin{equation*}
\lambda(\mathbf{P}) = \ominus_{gr} \mathbf{P},
\end{equation*}
where $\mathbf{P} \in \operatorname{Gr}_{n,p}$. 
Then Grassmann inverse maps are Grassmann gyroisometries.
\end{theorem}

\paragraph{Proof} See Appendix~\ref{sec:appendix_inverse_isometry_gr}.  

We note that the isometries of Grassmann manifolds with respect to different metrics have been
investigated in~\citet{BOTELHO20132226,GEHER20161585,GEHER2018287,IsoGrassQian21}.  
However, these works only show the general forms of these isometries, 
while our work gives specific expressions of some Grassmann gyroisometries with respect to 
the Grassmann gyrodistance function, thank to the closed-form expressions 
of left Grassmann gyrotranslations, 
gyroautomorphisms, 
and Grassmann inverse maps. 
To the best of our knowledge, these expressions of Grassmann gyroisometries have not appeared in previous works.   

\subsection{SPD Manifolds}
\label{subsec:gyrovector_spaces_spd_matrices}

In this section, we examine the similar concepts in Section~\ref{subsec:gr_isometries} for SPD manifolds.  
Section~\ref{subsec:spd_isometries} presents some isometries of SPD manifolds 
with Log-Euclidean~\cite{arsigny:inria-00070423}, Log-Cholesky~\cite{Lin_2019},  
and Affine-Invariant~\cite{pennec:inria-00070743} metrics. 
In Section~\ref{subsec:spd_multiclass_logistic_regression}, 
we define hyperplanes on SPD manifolds, and introduce the notion of 
SPD pseudo-gyrodistance from a SPD matrix or a set of SPD matrices to a hyperplane on SPD manifolds.  
These notations allow us to generalize MLR on SPD manifolds.  

\subsubsection{SPD Gyroisometries - The Isometries of SPD Manifolds}
\label{subsec:spd_isometries}

In~\citet{NguyenECCV22,NguyenNeurIPS22}, the author has shown that SPD manifolds with Log-Euclidean, Log-Cholesky, 
and Affine-Invariant metrics form gyrovector spaces referred to as LE, LC, and AI gyrovector spaces, respectively. 
We adopt the notations in these works and consider the case where $r=1$ (see~\citet{NguyenNeurIPS22}, Definition 3.1).  
Let $\oplus_{le},\oplus_{lc}$, and $\oplus_{ai}$ 
be the binary operations in LE, LC, and AI gyrovector spaces, respectively. 
Let $\otimes_{le},\otimes_{lc}$, and $\otimes_{ai}$ 
be the scalar multiplications in LE, LC, and AI gyrovector spaces, respectively. 
Let $\operatorname{gyr}_{le}[.,.],\operatorname{gyr}_{lc}[.,.]$, and $\operatorname{gyr}_{ai}[.,.]$ 
be the gyroautomorphisms in LE, LC, and AI gyrovector spaces, respectively. 
For convenience of presentation, we use the letter $g$ in the subscripts and superscripts 
of notations to indicate the Riemannian metric of the considered SPD manifold where $g \in \{ le,lc,ai \}$, 
unless otherwise stated.   
Denote by $\mathbf{I}_n$ the $n \times n$ identity matrix. 
We repeat the approach used in Section~\ref{subsec:gr_isometries} for SPD manifolds. 
The inner product on these manifolds is given below.  

\begin{definition}[{\bf The SPD Inner Product}]\label{def:inner_product_spd_matrices}
Let $\mathbf{P},\mathbf{Q} \in \operatorname{Sym}_n^+$. Then the SPD inner product of $\mathbf{P}$ and $\mathbf{Q}$ is defined as
\begin{align*}
\begin{split}
\langle \mathbf{P},\mathbf{Q} \rangle = \langle \operatorname{Log}^g_{\mathbf{I}_n}(\mathbf{P}), \operatorname{Log}^g_{\mathbf{I}_n}(\mathbf{Q}) \rangle_{\mathbf{I}_n},    
\end{split}
\end{align*}
where $\langle.,.\rangle_{\mathbf{I}_n}$ denotes the inner product at $\mathbf{I}_n$ given by the Riemannian metric of the considered manifold.  
\end{definition}

The SPD norm, SPD gyrodistance function, SPD gyroisometries, left SPD gyrotranslations, and SPD inverse maps 
are defined in the same way\footnote{For simplicity, we use the same notations for the SPD inner product, SPD norm, and SPD gyrodistance function as those on Grassmann manifolds since they should be clear from the context.} as those on Grassmann manifolds. 
Theorems~\ref{theorem:left_gyrotranslational_gyroisometries_ai} and~\ref{theorem:gyroautomorphisms_isometries_spd} 
characterize some SPD gyroisometries of LE, LC, and AI gyrovector spaces that are fully analogous with Grassmann gyroisometries. 

\begin{theorem}\label{theorem:left_gyrotranslational_gyroisometries_ai}
Left SPD gyrotranslations 
are SPD gyroisometries.
\end{theorem}

\paragraph{Proof} See Appendix~\ref{sec:appendix_left_gyrotranslational_gyroisometries_ai}.

\begin{theorem}\label{theorem:gyroautomorphisms_isometries_spd}
Gyroautomorphisms $\operatorname{gyr}_g[.,.]$ 
are SPD gyroisometries. 
\end{theorem}

\paragraph{Proof} See Appendix~\ref{sec:appendix_gyroautomorphisms_isometries_spd}.

\begin{theorem}\label{theorem:inverse_isometries_spd}
SPD inverse maps are SPD gyroisometries. 
\end{theorem}

\paragraph{Proof} See Appendix~\ref{sec:appendix_inverse_isometries_spd}.

The SPD gyroisometries given in 
Theorems~\ref{theorem:left_gyrotranslational_gyroisometries_ai},~\ref{theorem:gyroautomorphisms_isometries_spd}, 
and~\ref{theorem:inverse_isometries_spd} 
belong to a family of isometries of SPD manifolds discussed in~\citet{MolnarJordanTripleSPD15,MOLNAR2015141}. 
The difference between these works and ours is that our SPD gyroisometries are obtained from the gyrovector space perspective.    
Furthermore, our method can be applied to any metric on SPD manifolds as long as the basic operations and gyroautomorphism 
associated with that metric verify the axioms of gyrovector spaces considered in~\citet{NguyenNeurIPS22}.    

\subsubsection{Multiclass Logistic Regression on SPD Manifolds}
\label{subsec:spd_multiclass_logistic_regression}

Inspired by the works in~\citet{LebanonMarginClassifierICML04,NEURIPS2018_dbab2adc} that 
generalize MLR to multinomial and hyperbolic geometries, 
here we aim to generalize MLR to SPD manifolds. 

Given $K$ classes, MLR computes the probability of each of the output classes as
\begin{align}\label{eq:mlr_reexpression}
\begin{split}
p(y=k|x) = \frac{\exp( w_k^Tx  + b_k)}{\sum_{i=1}^K \exp( w_i^Tx  + b_i)} & \propto \exp( w_k^Tx + b_k), 
\end{split}
\end{align}
where $x$ is an input sample,  
$b_k \in \mathbb{R}$, $x,w_k \in \mathbb{R}^n, k=1,\ldots,K$. 

As shown in~\citet{LebanonMarginClassifierICML04,NEURIPS2018_dbab2adc}, Eq.~(\ref{eq:mlr_reexpression}) can be rewritten as
\begin{equation*}
p(y=k|x) \propto \exp(\operatorname{sign}( w_k^Tx + b_k) \| w_k \| d(x,\mathcal{H}_{w_k,b_k})),  
\end{equation*}
where $d(x,\mathcal{H}_{w_k,b_k})$ is the margin distance from point $x$ to a hyperplane $\mathcal{H}_{w_k,b_k}$.  

The generalization of MLR to SPD manifolds thus requires the definitions of
hyperplanes and margin distances in such manifolds. 
Guided by analogies with hyperbolic geometry~\cite{UngarHyperbolicNDim,NEURIPS2018_dbab2adc}, 
hyperplanes on SPD manifolds 
can be defined as follows. 

\begin{figure}[t]
  \begin{center}
    \begin{tabular}{c}      
      \includegraphics[width=1.0\linewidth, trim = 20 160 30 40, clip=true]{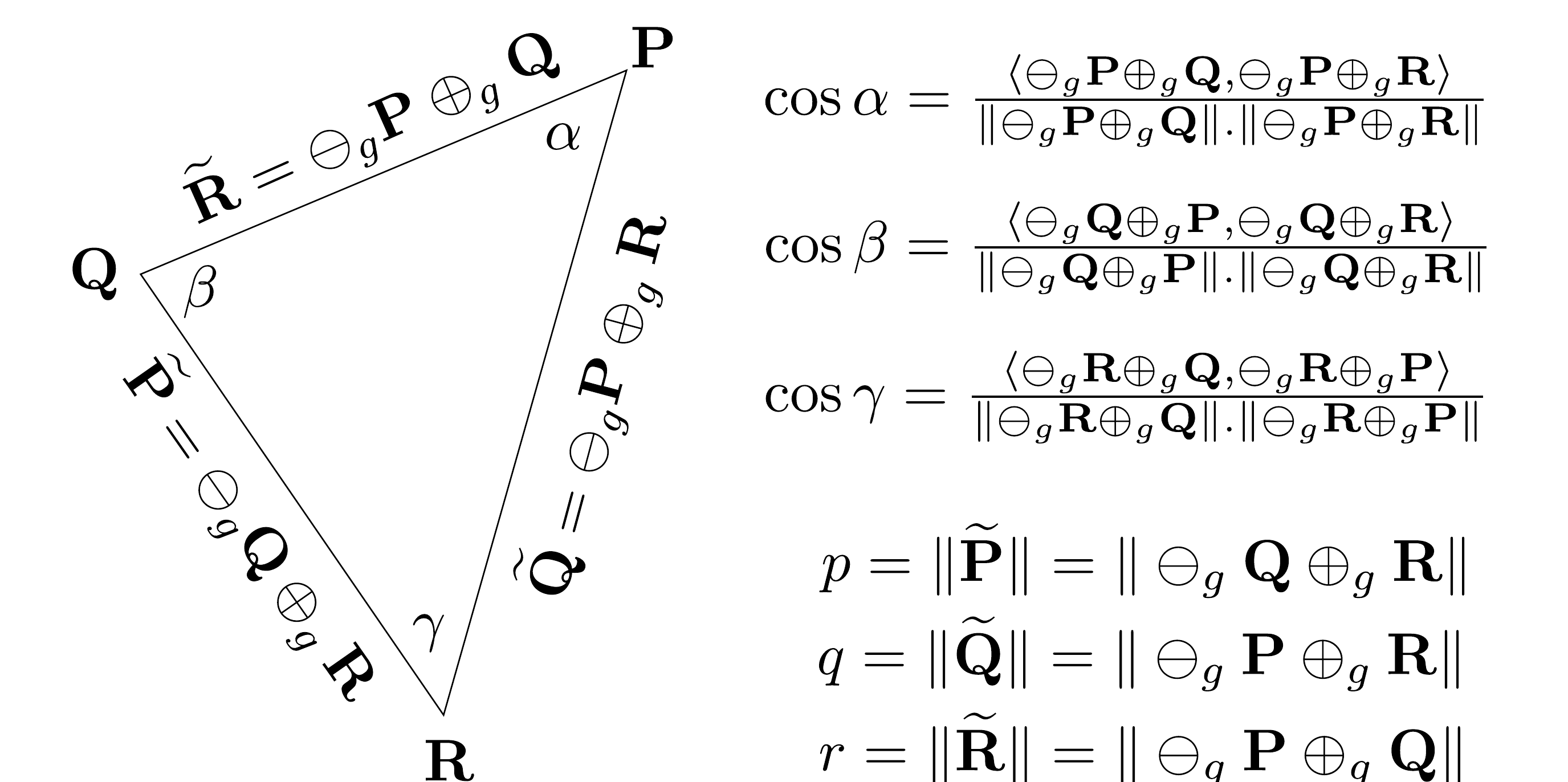}       
    \end{tabular}
  \end{center} 
  \caption{\label{fig:law_gyrocosines} Illustration of a SPD gyrotriangle, SPD gyroangles, and SPD gyrosides in a gyrovector space 
$(\operatorname{Sym}_n^+,\oplus_g,\otimes_g)$. 
}    
\end{figure}

\begin{definition}[{\bf SPD Hypergyroplanes}]\label{def:spd_hyperplanes}
For $\mathbf{P} \in \operatorname{Sym}_n^+$, $\mathbf{W} \in \mathcal{T}_{\mathbf{P}} \operatorname{Sym}_n^+$, 
SPD hypergyroplanes are defined as
\begin{equation}\label{eq:spd_hyperplanes}
\mathcal{H}_{\mathbf{W},\mathbf{P}} = \{ \mathbf{Q} \in \operatorname{Sym}_n^+: \langle \operatorname{Log}^g_{\mathbf{P}}(\mathbf{Q}),\mathbf{W}\rangle_{\mathbf{P}} = 0 \}.        
\end{equation}
\end{definition}

In order to define the margin distance from a SPD matrix to a SPD hypergyroplane, we need to generalize the notion 
of gyroangles on SPD manifolds, given below. 

\begin{definition}[{\bf The SPD Gyrocosine Function and SPD Gyroangles}]\label{def:gyrocosines_gyroangles_spd_matrices}
Let $\mathbf{P},\mathbf{Q}$, and $\mathbf{R}$ be three distinct SPD gyropoints (SPD matrices) in a gyrovector space
$(\operatorname{Sym}_n^+,\oplus_g,\otimes_g)$. 
The SPD gyrocosine of the measure of the SPD gyroangle $\alpha$, $0 \le \alpha \le \pi$, between 
$\ominus_g \mathbf{P} \oplus_g \mathbf{Q}$ and $\ominus_g \mathbf{P} \oplus_g \mathbf{R}$ 
is given by the equation
\begin{equation*}
\cos \alpha = \frac{\langle \ominus_g \mathbf{P} \oplus_g \mathbf{Q},\ominus_g \mathbf{P} \oplus_g \mathbf{R} \rangle}{\| \ominus_g \mathbf{P} \oplus_g \mathbf{Q} \|. \| \ominus_g \mathbf{P} \oplus_g \mathbf{R} \|}.  
\end{equation*}
The SPD gyroangle $\alpha$ is denoted by $\alpha = \angle \mathbf{Q} \mathbf{P} \mathbf{R}$.        
\end{definition}

Notice that our definition of the SPD gyrocosine of a SPD gyroangle is not based on unit gyrovectors~\cite{UngarHyperbolicNDim} 
and thus is not the same as that of the gyrocosine of a gyroangle in hyperbolic spaces.  
Similarly to Euclidean and hyperbolic spaces, one can state the Law of SPD gyrocosines.  
It will be useful later on when we introduce the concept of SPD pseudo-gyrodistance from a SPD matrix to a SPD hypergyroplane.  

\begin{theorem}[{\bf The Law of SPD Gyrocosines}]\label{theorem:gyrocosines_law}
Let $\mathbf{P},\mathbf{Q}$, and $\mathbf{R}$ be three distinct SPD gyropoints in a gyrovector space 
$(\operatorname{Sym}_n^+,\oplus_g,\otimes_g)$ where $g \in \{ le,lc \}$.        
Let $\widetilde{\mathbf{P}}=\ominus_g \mathbf{Q} \oplus_g \mathbf{R}$,  
$\widetilde{\mathbf{Q}}=\ominus_g \mathbf{P} \oplus_g \mathbf{R}$, and
$\widetilde{\mathbf{R}}=\ominus_g \mathbf{P} \oplus_g \mathbf{Q}$ be the SPD gyrosides of the SPD gyrotriangle 
formed by the three SPD gyropoints.  
Let $p = \| \widetilde{\mathbf{P}} \|$, $q = \| \widetilde{\mathbf{Q}} \|$, and $r = \| \widetilde{\mathbf{R}} \|$.  
Let $\alpha = \angle \mathbf{Q} \mathbf{P} \mathbf{R}$, 
$\beta = \angle \mathbf{P} \mathbf{Q} \mathbf{R}$, 
and $\gamma = \angle \mathbf{P} \mathbf{R} \mathbf{Q}$ be the SPD gyroangles of the SPD gyrotriangle. Then
\begin{equation*}
p^2 = q^2 + r^2 - 2qr \cos \alpha. 
\end{equation*}
\begin{equation*}
q^2 = p^2 + r^2 - 2pr \cos \beta. 
\end{equation*}
\begin{equation*}
r^2 = p^2 + q^2 - 2pq \cos \gamma.   
\end{equation*}

\end{theorem}

\paragraph{Proof} See Appendix~\ref{sec:appendix_gyrocosines_law}.

\begin{figure}[t]
  \begin{center}
    \begin{tabular}{c}      
      \includegraphics[width=1.0\linewidth, trim = 180 220 100 120, clip=true]{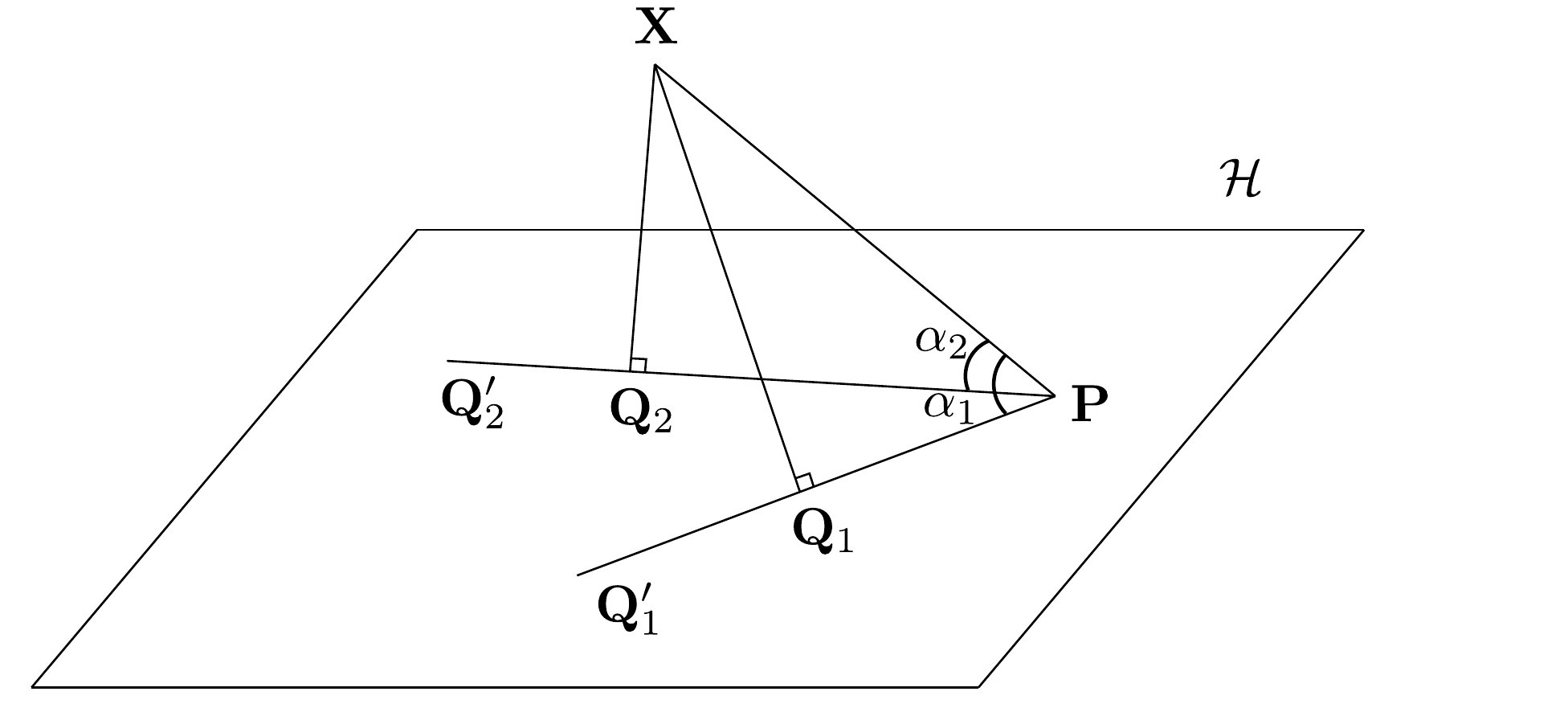}       
    \end{tabular}
  \end{center} 
  \caption{\label{fig:hyperplane_generalization} Illustration of the distance from a point $\mathbf{X}$ to a hyperplane $\mathcal{H}$ in $\mathbb{R}^n$. 
Here $\mathbf{P} \in \mathcal{H}$, 
$\mathbf{Q}'_1$ and $\mathbf{Q}'_2$ are two distinct points such that 
$\mathbf{Q}'_1,\mathbf{Q}'_2 \in \mathcal{H} \setminus \{ \mathbf{P} \}$,  
$\mathbf{Q}_1$ and $\mathbf{Q}_2$ are the projections of $\mathbf{X}$ on lines $\mathbf{P}\mathbf{Q}'_1$ and $\mathbf{P}\mathbf{Q}'_2$ 
that are supposed to belong to these lines, 
respectively, $\alpha_1$ and $\alpha_2$ are the angles that lines $\mathbf{P}\mathbf{Q}'_1$ and $\mathbf{P}\mathbf{Q}'_2$ 
make with line $\mathbf{P}\mathbf{X}$, respectively. 
If $\cos(\alpha_2) \ge \cos(\alpha_1)$, then $\| \mathbf{X}\mathbf{Q}_2 \|_F \le \| \mathbf{X}\mathbf{Q}_1 \|_F$. 
The distance from $\mathbf{X}$ to hyperplane $\mathcal{H}$ 
is obtained when $\cos(\alpha)$, $\alpha$ is the angle between 
lines $\mathbf{P}\mathbf{X}$ and $\mathbf{P}\mathbf{Q}$, $\mathbf{Q} \in \mathcal{H} \setminus \{ \mathbf{P} \}$, 
gets the maximum value.   
}    
\end{figure}

Fig.~\ref{fig:law_gyrocosines} illustrates the notions given in Theorem~\ref{theorem:gyrocosines_law}. 
It states that one can calculate a SPD gyroside of a SPD gyrotriangle
when the SPD gyroangle opposite to the SPD gyroside and the other two SPD gyrosides are known. 
This result is fully analogous with those in Euclidean and hyperbolic spaces. 
The Law of SPD gyrosines is given in Appendix~\ref{appendix:law_of_spd_gyrosines}. 

We now introduce the concept of SPD pseudo-gyrodistance from a SPD matrix to a SPD hypergyroplane 
that is inspired from a property of the distance from a point in $\mathbb{R}^n$ to a hyperplane in $\mathbb{R}^n$. 
The key idea is illustrated in Fig.~\ref{fig:hyperplane_generalization}.  

\begin{definition}[{\bf The SPD Pseudo-gyrodistance from a SPD Matrix to a SPD Hypergyroplane}]\label{def:pseudo_distances_spd_hyperplanes}
Let $\mathcal{H}_{\mathbf{W},\mathbf{P}}$ be a SPD hypergyroplane, 
and $\mathbf{X} \in \operatorname{Sym}_n^+$. The SPD pseudo-gyrodistance from $\mathbf{X}$ to $\mathcal{H}_{\mathbf{W},\mathbf{P}}$ is defined as
\begin{equation*}
\bar{d}(\mathbf{X},\mathcal{H}_{\mathbf{W},\mathbf{P}}) = \sin(\angle \mathbf{X} \mathbf{P} \bar{\mathbf{Q}}) d(\mathbf{X},\mathbf{P}),
\end{equation*}
where $\bar{\mathbf{Q}}$ is given by
\begin{align*}
\begin{split}
\bar{\mathbf{Q}} = \argmax_{\mathbf{Q} \in \mathcal{H}_{\mathbf{W},\mathbf{P}} \setminus \{ \mathbf{P} \}} \Big( \frac{\langle \ominus_g \mathbf{P} \oplus_g \mathbf{Q}, \ominus_g \mathbf{P} \oplus_g \mathbf{X} \rangle}{\| \ominus_g \mathbf{P} \oplus_g \mathbf{Q} \|.\| \ominus_g \mathbf{P} \oplus_g \mathbf{X} \|} \Big).
\end{split}
\end{align*}

By convention, $\sin(\angle \mathbf{X} \mathbf{P} \mathbf{Q}) = 0$ 
for any $\mathbf{X},\mathbf{Q} \in \mathcal{H}_{\mathbf{W},\mathbf{P}}$.  
\end{definition}

The SPD gyrodistance from $\mathbf{X}$ to $\mathcal{H}_{\mathbf{W},\mathbf{P}}$ is defined as
\begin{equation*}
d(\mathbf{X},\mathcal{H}_{\mathbf{W},\mathbf{P}}) = \min_{\mathbf{Q} \in \mathcal{H}_{\mathbf{W},\mathbf{P}}} d(\mathbf{X},\mathbf{Q}).
\end{equation*}


From Theorem~\ref{theorem:gyrocosines_law}, it turns out that the SPD pseudo-gyrodistance agrees with 
the SPD gyrodistance in certain cases. In particular, we have the following results.  


\begin{theorem}[{\bf The SPD Gyrodistance from a SPD Matrix to a SPD Hypergyroplane in a LE Gyrovector Space}]\label{theorem:distance_to_SPD_hyperplanes_log_euclidean}
Let $\mathcal{H}_{\mathbf{W},\mathbf{P}}$ be a SPD hypergyroplane in a gyrovector space $(\operatorname{Sym}_n^+,\oplus_{le},\otimes_{le})$, 
and $\mathbf{X} \in \operatorname{Sym}_n^+$. Then the SPD pseudo-gyrodistance from $\mathbf{X}$ to  
$\mathcal{H}_{\mathbf{W},\mathbf{P}}$ is equal to the SPD gyrodistance from $\mathbf{X}$ to  
$\mathcal{H}_{\mathbf{W},\mathbf{P}}$ and is given by
\begin{equation*}
d(\mathbf{X},\mathcal{H}_{\mathbf{W},\mathbf{P}}) = \frac{| \langle \log(\mathbf{X}) - \log(\mathbf{P}),D\log_{\mathbf{P}}(\mathbf{W}) \rangle_F |}{\| D\log_{\mathbf{P}}(\mathbf{W}) \|_F}.
\end{equation*} 
\end{theorem}

\paragraph{Proof} See Appendix~\ref{sec:appendix_distance_to_SPD_hyperplanes_log_euclidean}.

\begin{theorem}[{\bf The SPD Gyrodistance from a SPD Matrix to a SPD Hypergyroplane in a LC Gyrovector Space}]\label{theorem:distance_to_SPD_hyperplanes_log_cholesky}
Let $\mathcal{H}_{\mathbf{W},\mathbf{P}}$ be a SPD hypergyroplane in a gyrovector space $(\operatorname{Sym}_n^+,\oplus_{lc},\otimes_{lc})$,  
and $\mathbf{X} \in \operatorname{Sym}_n^+$. Then the SPD pseudo-gyrodistance from $\mathbf{X}$ to  
$\mathcal{H}_{\mathbf{W},\mathbf{P}}$ is equal to the SPD gyrodistance from $\mathbf{X}$ to  
$\mathcal{H}_{\mathbf{W},\mathbf{P}}$ and is given by
\begin{equation*}
d(\mathbf{X},\mathcal{H}_{\mathbf{W},\mathbf{P}}) = \frac{| \langle \mathbf{A},\mathbf{B} \rangle_F |}{\| \mathbf{B} \|_F},
\end{equation*} 
where
\begin{equation*}
\mathbf{A} = -\lfloor \varphi(\mathbf{P}) \rfloor + \lfloor \varphi(\mathbf{X}) \rfloor + \log( \mathbb{D}(\varphi(\mathbf{P}))^{-1} \mathbb{D}(\varphi(\mathbf{X})) ), 
\end{equation*}
\begin{equation*}
\mathbf{B} = \lfloor \widetilde{\mathbf{W}} \rfloor + \mathbb{D}(\varphi(\mathbf{P}))^{-1} \mathbb{D}(\widetilde{\mathbf{W}}),
\end{equation*}
\begin{equation*} 
\widetilde{\mathbf{W}} = \varphi(\mathbf{P}) \Big( \varphi(\mathbf{P})^{-1} \mathbf{W} (\varphi(\mathbf{P})^{-1})^T \Big)_{\frac{1}{2}},
\end{equation*}  
where $\lfloor \mathbf{Y} \rfloor$ is a 
matrix of the same size as matrix $\mathbf{Y} \in \operatorname{M}_{n,n}$ whose $(i,j)$ element is $\mathbf{Y}_{ij}$ 
if $i > j$ and is zero otherwise, 
$\mathbb{D}(\mathbf{Y})$ is a diagonal matrix of the same
size as matrix $\mathbf{Y}$ whose $(i,i)$ element is $\mathbf{Y}_{ii}$, 
$\mathbf{Y}_{\frac{1}{2}}$ is the lower triangular part of $\mathbf{Y}$ with the diagonal entries halved,  
and $\varphi(\mathbf{Q})$ denotes the Cholesky factor of $\mathbf{Q} \in \operatorname{Sym}_n^+$, i.e., 
$\varphi(\mathbf{Q})$ is a lower triangular matrix with positive diagonal entries 
such that $\mathbf{Q} = \varphi(\mathbf{Q}) \varphi(\mathbf{Q})^T$.     
\end{theorem}

\paragraph{Proof} See Appendix~\ref{sec:appendix_distance_to_SPD_hyperplanes_log_cholesky}.


We cannot establish an equivalent result in the case of AI gyrovector spaces. 
Nevertheless, a closed-form expression for the SPD pseudo-gyrodistance can still be obtained in this case. 

\begin{theorem}[{\bf The SPD Pseudo-gyrodistance from a SPD Matrix to a SPD Hypergyroplane in an AI Gyrovector Space}]\label{theorem:distance_to_SPD_hyperplanes_affine_invariant}
Let $\mathcal{H}_{\mathbf{W},\mathbf{P}}$ be a SPD hypergyroplane in a gyrovector space $(\operatorname{Sym}_n^+,\oplus_{ai},\otimes_{ai})$, and $\mathbf{X} \in \operatorname{Sym}_n^+$.  
Then the SPD pseudo-gyrodistance from $\mathbf{X}$ to $\mathcal{H}_{\mathbf{W},\mathbf{P}}$ is given by
\begin{equation*}
\bar{d}(\mathbf{X},\mathcal{H}_{\mathbf{W},\mathbf{P}}) = \frac{| \langle \log(\mathbf{P}^{-\frac{1}{2}} \mathbf{X} \mathbf{P}^{-\frac{1}{2}}), \mathbf{P}^{-\frac{1}{2}} \mathbf{W} \mathbf{P}^{-\frac{1}{2}} \rangle_F |}{ \| \mathbf{P}^{-\frac{1}{2}} \mathbf{W} \mathbf{P}^{-\frac{1}{2}} \|_F }.
\end{equation*} 
\end{theorem}

\paragraph{Proof} See Appendix~\ref{sec:appendix_distance_to_SPD_hyperplanes_affine_invariant}.        


The results in Theorems~\ref{theorem:distance_to_SPD_hyperplanes_log_euclidean},
~\ref{theorem:distance_to_SPD_hyperplanes_log_cholesky}, 
and~\ref{theorem:distance_to_SPD_hyperplanes_affine_invariant} lead to Corollary~\ref{corollary:distance_set_log_euclidean} 
that concerns with the SPD gyrodistance and pseudo-gyrodistance from a set of SPD matrices to
a SPD hypergyroplane.  

\begin{table}[t]
\begin{center}
  \resizebox{1.0\linewidth}{!}{
  \def\arraystretch{1.2}
  \begin{tabular}{| l | c | c | c | c | c |}    
    \hline
    Dataset & SPDNet & SPDNetBN & GyroLE & GyroLC & GyroAI  \\          
    \hline             
    HDM05  & 72.83 & {\bf 76.42} & 72.64 & 63.78 & 73.34 \\ 
    \#HDM05 & 6.58 & 6.68 & 6.53 & 6.53 & 6.53 \\ 
    \hline    
    FPHA & 89.25 & 91.34 & {\bf 94.61} & 82.43 & 93.39 \\      
    \#FPHA & 0.99 & 1.03 & 0.95 & 0.95 & 0.95 \\
    \hline
    NTU60 & 77.82 & 79.61 & 81.68 & 72.26 & {\bf 82.75}   \\
    \#NTU60 & 1.80 & 2.06 & 1.49 & 1.49 & 1.49 \\      
    \hline	
  \end{tabular}
  } 
\end{center}
\caption{\label{tab:exp_compare_with_spd_neural_networks} Accuracy comparison (\%) of our SPD models against SPDNet and SPDNetBN 
with comparable model sizes (MB). 
}
\end{table}

\begin{corollary}\label{corollary:distance_set_log_euclidean}
Let $\mathbf{P}_1,\ldots,\mathbf{P}_N,\mathbf{Q}_1,\ldots,\mathbf{Q}_N$, 
and $\mathbf{X}_1,\ldots,\mathbf{X}_N \in \operatorname{Sym}_n^+$. 
Let $\mathbf{W}_1,\ldots,\mathbf{W}_N \in \operatorname{Sym}_n$. 
Denote by $\diag(\mathbf{P}_{1},\ldots,\mathbf{P}_{N})$ the following matrix:
\begin{equation*}
\diag(\mathbf{P}_{1},\ldots,\mathbf{P}_{N}) = \begin{bmatrix} \mathbf{P}_{1} \cdots \cdots \\ \cdots \mathbf{P}_{2} \cdots \\ \cdots \cdots \mathbf{P}_{N}  \end{bmatrix},
\end{equation*}
where the diagonal entries of $\mathbf{P}_i,i=1,\ldots,N$ belong to the diagonal entries of $\diag(\mathbf{P}_{1},\ldots,\mathbf{P}_{N})$.  
Let $\mathbf{P} = \diag(\mathbf{P}_{1},\ldots,\mathbf{P}_{N})$,
$\mathbf{Q} = \diag(\mathbf{Q}_{1},\ldots,\mathbf{Q}_{N})$, 
$\mathbf{X} = \diag(\mathbf{X}_{1},\ldots,\mathbf{X}_{N})$, 
and $\mathbf{W} = \diag(\mathbf{W}_{1},\ldots,\mathbf{W}_{N})$. 

(1) Denote by $\mathcal{H}_{\mathbf{W},\mathbf{P}}$ a SPD hypergyroplane 
in a gyrovector space $(\operatorname{Sym}_n^+,\oplus_{le},\otimes_{le})$. 
Then the SPD gyrodistance from $\mathbf{X}$ to $\mathcal{H}_{\mathbf{W},\mathbf{P}}$ is given by
\begin{equation}\label{eq:distance_set_to_hyperplane_le}
d(\mathbf{X},\mathcal{H}_{\mathbf{W},\mathbf{P}}) = \frac{| \sum_{i=1}^N \langle \log(\mathbf{X}_i) - \log(\mathbf{P}_i),D\log_{\mathbf{P}_i}(\mathbf{W}_i) \rangle_F |}{ \sqrt{ \sum_{i=1}^N \| D\log_{\mathbf{P}_i}(\mathbf{W}_i) \|_F^2} }.
\end{equation} 

(2) Denote by $\mathcal{H}_{\mathbf{W},\mathbf{P}}$ a SPD hypergyroplane 
in a gyrovector space $(\operatorname{Sym}_n^+,\oplus_{lc},\otimes_{lc})$. 
Then the SPD gyrodistance from $\mathbf{X}$ to $\mathcal{H}_{\mathbf{W},\mathbf{P}}$ is given by
\begin{equation}\label{eq:distance_set_to_hyperplane_lc}
d(\mathbf{X},\mathcal{H}_{\mathbf{W},\mathbf{P}}) = \frac{| \sum_{i=1}^N \langle \mathbf{A}_i,\mathbf{B}_i \rangle_F |}{\sqrt{ \sum_{i=1}^N \| \mathbf{B}_i \|_F^2} },
\end{equation} 
where
\begin{equation*}
\mathbf{A}_i = -\lfloor \varphi(\mathbf{P}_i) \rfloor + \lfloor \varphi(\mathbf{X}_i) \rfloor + \log( \mathbb{D}(\varphi(\mathbf{P}_i))^{-1} \mathbb{D}(\varphi(\mathbf{X}_i)) ), 
\end{equation*}
\begin{equation*}
\mathbf{B}_i = \lfloor \widetilde{\mathbf{W}}_i \rfloor + \mathbb{D}(\varphi(\mathbf{P}_i))^{-1} \mathbb{D}(\widetilde{\mathbf{W}}_i),
\end{equation*}
\begin{equation*} 
\widetilde{\mathbf{W}}_i = \varphi(\mathbf{P}_i) \Big( \varphi(\mathbf{P}_i)^{-1} \mathbf{W}_i (\varphi(\mathbf{P}_i)^{-1})^T \Big)_{\frac{1}{2}}.
\end{equation*}  

(3) Denote by $\mathcal{H}_{\mathbf{W},\mathbf{P}}$ a SPD hypergyroplane 
in a gyrovector space $(\operatorname{Sym}_n^+,\oplus_{ai},\otimes_{ai})$. Then the SPD pseudo-gyrodistance from $\mathbf{X}$ to $\mathcal{H}_{\mathbf{W},\mathbf{P}}$ is given by
\begin{equation}\label{eq:distance_set_to_hyperplane_ai}
\bar{d}(\mathbf{X},\mathcal{H}_{\mathbf{W},\mathbf{P}}) = \frac{| \sum_{i=1}^N \langle \log(\mathbf{P}_i^{-\frac{1}{2}} \mathbf{X}_i \mathbf{P}_i^{-\frac{1}{2}}), \mathbf{P}_i^{-\frac{1}{2}} \mathbf{W}_i \mathbf{P}_i^{-\frac{1}{2}} \rangle_F |}{\sqrt{ \sum_{i=1}^N \| \mathbf{P}_i^{-\frac{1}{2}} \mathbf{W}_i \mathbf{P}_i^{-\frac{1}{2}} \|_F^2} }. 
\end{equation} 
\end{corollary}

\paragraph{Proof} See Appendix~\ref{sec:appendix_distance_set_log_euclidean}.

\begin{table}[t]
\begin{center}
  \resizebox{1.0\linewidth}{!}{
  \def\arraystretch{1.2}
  \begin{tabular}{| l | c | c | c | c |}    
    \hline
    Dataset & GyroAI-HAUNet & MLR-LE & MLR-LC & MLR-AI  \\          
    \hline   
    HDM05  & 78.14 & 77.62 & 71.35 & {\bf 79.84} \\  
    \#HDM05 & 0.31 & 0.60 & 0.60 & 0.60 \\ 
    \hline
    FPHA & 96.00 & {\bf 96.44} & 88.62 & 96.26 \\      
    \#FPHA & 0.11 & 0.21 & 0.21 & 0.21 \\
    \hline
    NTU60 & 94.72 & 95.87 & 88.24 & {\bf 96.48} \\ 
    \#NTU60 & 0.02 & 0.05 & 0.05 & 0.05 \\     
    \hline	
  \end{tabular}
  } 
\end{center}
\caption{\label{tab:exp_compare_with_gyroai_haunet} Accuracy comparison (\%) of our SPD models against GyroAI-HAUNet. 
}
\end{table}

\section{Experiments}

In this section, we report results of our experiments for two applications, 
i.e.,  human action recognition and knowledge graph completion. 
Details on the datasets and our experimental settings 
are given in Appendix~\ref{sec:appendix_har_exp_settings}.   

\subsection{Human Action Recognition}  
\label{app:hau}


We use three datasets, 
i.e., HDM05~\cite{hdm05}, FPHA~\cite{Garcia-HernandoCVPR18}, and NTU60~\cite{Shahroudy16NTU}. 

\begin{table*}[t]
\begin{center}
  \resizebox{0.75\linewidth}{!}{
  \def\arraystretch{1.2}
  \begin{tabular}{| l | c | c | c | c | c | c | c | c |}    
    \hline
    Dataset & LSTM & ST-TR & HypGRU & ST-GCN & Shift-GCN & MLR-LE & MLR-LC & MLR-AI  \\          
    \hline   
    HDM05  & 72.82 & 76.12 & 58.50 & 76.58 & {\bf 80.28} & 77.62 & 71.35 & 79.84 \\  
    \#HDM05 & 0.54 & 27.73 & 0.61 & 17.73 & 4.20 & 0.60 & 0.60 & 0.60 \\ 
    \hline
    FPHA & 81.22 & 91.34 & 61.42 & 78.78 & 91.08 & 96.44 & 88.62 & 96.26 \\      
    \#FPHA & 0.41 & 27.55 & 0.47 & 17.60 & 3.84 & 0.21 & 0.21 &	0.21 \\
    \hline
    NTU60 & 87.27 & 93.78 & 88.03 & 91.75 & 95.01 & 95.87 & 88.24 & 96.48 \\ 
    \#NTU60 & 0.035 & 27.50 & 0.039 & 17.66 & 3.90 & 0.05 & 0.05 & 0.05 \\     
    \hline	
  \end{tabular}
  } 
\end{center}
\caption{\label{tab:exp_compare_with_sota} Accuracy comparison (\%) of our SPD models against state-of-the-art models. 
}
\end{table*}

\subsubsection{SPD Neural Networks}
\label{subsec:activity_understanding_results_spd_neural_networks}

We design three networks, each of them is composed of a layer based on the Affine-Invariant translation model 
and of a MLR (Log-Euclidean, Log-Cholesky, and Affine-Invariant, see Section~\ref{subsec:spd_multiclass_logistic_regression}). 
These networks are compared against SPDNet~\cite{HuangGool17}\footnote{\url{https://github.com/zhiwu-huang/SPDNet}} 
and SPDNetBN~\cite{BrooksRieBatNorm19}\footnote{\url{https://papers.nips.cc/paper/2019/hash/6e69ebbfad976d4637bb4b39de261bf7-Abstract.html}}. 
Temporal pyramid representation is used as in~\citet{NguyenNeurIPS22}.  
Each sequence is then represented by a set of SPD matrices. 
We use Eqs.~(\ref{eq:distance_set_to_hyperplane_le}),~(\ref{eq:distance_set_to_hyperplane_lc}), 
and~(\ref{eq:distance_set_to_hyperplane_ai}) to compute the SPD gyrodistances and pseudo-gyrodistances for MLR. 
Results of the five networks 
are given in Tab.~\ref{tab:exp_compare_with_spd_neural_networks}.          
On HDM05 dataset, GyroLE is on par with SPDNet while GyroAI outperforms SPDNet. 
On FPHA and NTU60 datasets, GyroLE and GyroAI outperform both SPDNet and SPDNetBN. 
            
We also compare GyroAI-HAUNet in~\citet{NguyenNeurIPS22} against three other networks in which we replace
the classification layer of GyroAI-HAUNet with a MLR based on Log-Euclidean, Log-Cholesky, and Affine-Invariant metrics, respectively.  
Results of the four networks are shown in Tab.~\ref{tab:exp_compare_with_gyroai_haunet}.  
The best results are obtained by our models MLR-AI or MLR-LE. 
However, these models have 2x more parameters than GyroAI-HAUNet. 

Tab.~\ref{tab:exp_compare_with_sota} reports results of our SPD models and those of some state-of-the-art models
from four categories of neural networks: recurrent neural networks (i.e., LSTM), hyperbolic neural networks (i.e., HypGRU~\cite{NEURIPS2018_dbab2adc}), 
graph neural networks (i.e., ST-GCN~\cite{YanAAAI18} and Shift-GCN~\cite{ChenShiftGCN20}), 
and transformers (i.e., ST-TR~\cite{Plizzari2021skeleton}).
MLR-LE and MLR-AI outperform the other networks on FPHA and NTU60 datasets. 
Also, our networks use far fewer parameters than the transformer and graph neural networks. 

\subsubsection{Grassmann Neural Networks}
\label{subsec:activity_understanding_results_grassmann_neural_networks}

Huang et al.~\cite{HuangAAAI18} proposed a discriminative Grassmann neural network called GrNet. 
The network applies the FRmap layer to reduce the dimension of input matrices. 
Orthogonal matrices are then obtained from the outputs of the FRmap layer via QR-decomposition performed by the ReOrth layer.  
This creates an issue in the backward pass of the ReOrth layer, where the inverse of upper-triangular matrices must be computed. 
In practice, these matrices are often ill-conditioned and cannot be returned by 
popular deep learning frameworks like 
Tensorflow\footnote{\url{https://github.com/master/tensorflow-riemopt/tree/master/examples/grnet}.} and Pytorch. 
We address this issue by replacing the FRmap and ReOrth layers with a layer based on the 
Grassmann translation model (see Section~\ref{subsec:gr_isometries}).  
The resulting network GyroGr is compared against GrNet based on 
its official Matlab code\footnote{\url{https://github.com/zhiwu-huang/GrNet}.}. 
Results of the two networks are given in Tab.~\ref{tab:exp_grassmann_trans_vs_frmap_orthomap}.  
GyroGr outperforms GrNet by 3.60\%, 2.79\%, and 2.14\% on HDM05, FPHA, and NTU60 datasets, respectively.  
These results clearly demonstrate the effectiveness of the Grassmann translation model in a 
discriminative Grassmann neural network like GrNet.  

\begin{table}[t]
\begin{center}
  \resizebox{0.62\linewidth}{!}{
  \def\arraystretch{1.3}
  \begin{tabular}{| l | c | c | c |}    
    \hline
    Method & HDM05 & FPHA & NTU60  \\          
    \hline   
    GrNet  & 52.71 & 81.91 & 65.45 \\  
    GyroGr & {\bf 56.32} & {\bf 84.70} & {\bf 67.60} \\ 
    \hline	
  \end{tabular}
  } 
\end{center}
\caption{\label{tab:exp_grassmann_trans_vs_frmap_orthomap} Accuracy comparison (\%) of GyroGr against GrNet. 
}
\end{table}

We also conduct another experiment in order to compare the Grassmann translation model and Grassmann scaling model
in~\citet{NguyenNeurIPS22} within the framework of GrNet. To this end, we design a new network from GyroGr 
by replacing the Grassmann translation layer with a Grassmann scaling layer. 
Results of the two networks are presented in Tab.~\ref{tab:exp_grassmann_trans_vs_scale}.  
GyroGr significantly outperforms GyroGr-Scaling on all the datasets, 
showing that the Grassmann translation operation is much more effective than the matrix scaling 
within the framework of GrNet. 

\begin{table}[t]
\begin{center}
  \resizebox{0.72\linewidth}{!}{
  \def\arraystretch{1.3}
  \begin{tabular}{| l | c | c | c |}    
    \hline
    Method & HDM05 & FPHA & NTU60  \\          
    \hline       
    GyroGr-Scaling  & 45.69  & 65.74 & 55.26 \\      
    GyroGr & {\bf 56.32} & {\bf 84.70} & {\bf 67.60} \\      
    \hline	
  \end{tabular}
  } 
\end{center}
\caption{\label{tab:exp_grassmann_trans_vs_scale} Accuracy comparison (\%) of GyroGr against the Grassmann scaling model in~\citet{NguyenNeurIPS22}. 
}
\end{table}

\subsection{Knowledge Graph Completion}
\label{app:kgc}

 
The goal of this experiment is to compare the Grassmann model based on the projector perspective 
in~\citet{NguyenNeurIPS22} against the one based on the ONB perspective. 
We use two datasets, i.e., WN18RR~\cite{MillerWordNet95} and FB15k-237~\cite{Toutanova2015RepresentingTF}.

Following~\citet{balazevic2019multi,NguyenNeurIPS22}, we design a model that learns a scoring function
\begin{equation*}\label{eq:kgc_score_function}
\phi_{kgc}(e_s,r,e_o) = -d((\mathbf{A} \widetilde{\otimes} \mathbf{S}) \widetilde{\oplus}_{gr} \mathbf{R}, \mathbf{O})^2 + b_s + b_o,
\end{equation*}
where $\mathbf{S}$ and $\mathbf{O}$ are embeddings of the subject and object entities, respectively,  
$\mathbf{R}$ and $\mathbf{A}$ are matrices associated with relation $r$, 
$b_s,b_o \in \mathbb{R}$ are scalar biases for the subject and object entities, respectively. 
The operation $\widetilde{\otimes}$ is defined as
\begin{equation*} 
\mathbf{A} \widetilde{\oplus}_{gr} \mathbf{P} = \exp\bigg(\begin{bmatrix} 0 & \mathbf{A} * \mathbf{B} \\ -(\mathbf{A}*\mathbf{B})^T & 0 \end{bmatrix}\bigg) \widetilde{\mathbf{I}}_{n,p}, 
\end{equation*}  
where $\mathbf{A} \in \operatorname{M}_{p,n-p}$, and $\mathbf{P}$ is given by
\begin{equation*}
\mathbf{P} = \exp\bigg( \begin{bmatrix} 0 & \mathbf{B} \\ -\mathbf{B}^T & 0 \end{bmatrix} \bigg) \widetilde{\mathbf{I}}_{n,p}.                 
\end{equation*}

The binary operation $\widetilde{\oplus}_{gr}$ is defined in Eq.~(\ref{eq:matrix_matrix_addition_gr}). 
We use the distance function~\cite{Edelman98}
\begin{equation}\label{eq:kgc_distance_function}
d(\mathbf{P},\mathbf{Q}) = \| \theta \|_2,
\end{equation}
where $\theta_i,i=1,\ldots,p$ are the principle angles between two subspaces spanned by the columns of $\mathbf{P}$ 
and $\mathbf{Q}$, i.e., $\mathbf{U} \diag(\cos(\theta_1),\ldots,\cos(\theta_p)) \mathbf{V}^T$ 
is the SVD of $\mathbf{P}^T \mathbf{Q}$. 

\begin{table}[t]
\begin{center}
  \resizebox{1.0\linewidth}{!}{
  \def\arraystretch{1.3}
  \begin{tabular}{| c | l | c | c | c | c |}    
    \hline
    \multirow{2}{*}{DOF} & \multirow{2}{*}{Model} & \multirow{2}{*}{MRR} & \multirow{2}{*}{H@1} & \multirow{2}{*}{H@3} & \multirow{2}{*}{H@10}  \\ 
    & & & & & \\
    \hline        
    \multirow{2}{*}{144} & GyroGr-KGC$_{\operatorname{proj+sca}}$ & 44.2 & 38.5 & 46.8 & 54.6  \\ 
    & GyroGr-KGC$_{\operatorname{onb+sca}}$ & {\bf 44.9} & {\bf 39.5} & {\bf 47.2} & 54.6  \\ 
    \hline
  \end{tabular}
  } 
\end{center}
\caption{\label{tab:exp_kgc_onb_projector_wn18rr} 
Comparison of our Grassmann model against the Grassmann model in~\citet{NguyenNeurIPS22} 
on the validation set of WN18RR dataset.  
GyroGr-KGC$_{\operatorname{onb+sca}}$ learns embeddings in $\widetilde{\operatorname{Gr}}_{24,12}$.
GyroGr-KGC$_{\operatorname{proj+sca}}$ learns embeddings in $\operatorname{Gr}_{24,12}$  
(DOF stands for degrees of freedom).  
}
\end{table}

\begin{table}[t]
\begin{center}
  \resizebox{1.0\linewidth}{!}{
  \def\arraystretch{1.3}
  \begin{tabular}{| c | l | c | c | c | c |}    
    \hline
    \multirow{2}{*}{DOF} & \multirow{2}{*}{Model} & \multirow{2}{*}{MRR} & \multirow{2}{*}{H@1} & \multirow{2}{*}{H@3} & \multirow{2}{*}{H@10}  \\ 
    & & & & & \\
    \hline        
    \multirow{2}{*}{144} & GyroGr-KGC$_{\operatorname{proj+sca}}$ & 29.3 & 20.5 & 32.4 & 46.8  \\ 
    & GyroGr-KGC$_{\operatorname{onb+sca}}$ & {\bf 29.9} & {\bf 20.8} & {\bf 33.2} & {\bf 48.2}  \\ 
    \hline
  \end{tabular}
  } 
\end{center}
\caption{\label{tab:exp_kgc_onb_projector_fb237} 
Comparison of our Grassmann model against the Grassmann model in~\citet{NguyenNeurIPS22} 
on the validation set of FB15k-237 dataset.  
GyroGr-KGC$_{\operatorname{onb+sca}}$ learns embeddings in $\widetilde{\operatorname{Gr}}_{24,12}$.
GyroGr-KGC$_{\operatorname{proj+sca}}$ learns embeddings in $\operatorname{Gr}_{24,12}$.  
}
\end{table}


Results of our model 
and the Grassmann model in~\citet{NguyenNeurIPS22} on the validation sets of WN18RR and FB15k-237 datasets 
are shown in Tabs.~\ref{tab:exp_kgc_onb_projector_wn18rr} and~\ref{tab:exp_kgc_onb_projector_fb237}, respectively. 
Our model GyroGr-KGC$_{\operatorname{onb+sca}}$ gives the same or better performance 
than GyroGr-KGC$_{\operatorname{proj+sca}}$ in all cases. 
In particular, on WN18RR dataset, GyroGr-KGC$_{\operatorname{onb+sca}}$ outperforms 
GyroGr-KGC$_{\operatorname{proj+sca}}$ by 1\% in terms of H@1.
On FB15k-237 dataset, GyroGr-KGC$_{\operatorname{onb+sca}}$ outperforms 
GyroGr-KGC$_{\operatorname{proj+sca}}$ by 1.4\% in terms of H@10.  

\subsection{Complexity Analysis}
\label{sec:complexity_analysis}

Let $n$ be the size of input matrices (SPD or projection matrices), 
$n_c$ be the number of action classes, $n_s$ be the number of SPD matrices used by GyroAI and MLR-AI for representing an action sequence, 
$n_t$ and $n_p$ be the number of transformation matrices and the number of projection matrices for the W-ProjPooling layer in GyroGr, respectively. For the sake of simplicity, we analyze the complexity of the models for one training sample and one iteration.  
    
\begin{itemize}
\item GyroAI: 
The binary operation has time complexity $O(n^3)$ and memory complexity $O(n^2)$. The MLR has time complexity $O(n_cn_sn^3)$ and memory complexity $O(n_cn_sn^2)$. 
\item MLR-AI: 
The RNN cell has time complexity $O(n^3)$ and memory complexity $O(n^2)$. The MLR has time complexity $O(n_cn_sn^3)$ and memory complexity $O(n_cn_sn^2)$. 
\item GyroGr: 
The Grassmann translation layer and OrthMap layer of GrNet have time complexity $O(n_tn^3)$ and memory complexity $O(n_tn^2)$. The W-ProjPooling layer (pooling within one projection matrix) has time complexity $O(n_pn^2)$ and memory complexity $O(n_pn^2)$. The classification layer has time complexity $O(n_tn_cn^2)$ and memory complexity $O(n_tn_cn^2)$. 
\item GyroGr-KGC: The computation of the scoring function has time complexity $O(n^3)$ and memory complexity $O(n^2)$.
\end{itemize}

\section{Limitation}
\label{sec:discussion}

As pointed out in~\citet{shimizu2021hyperbolic}, the hyperbolic MLR~\cite{NEURIPS2018_dbab2adc} is over-parameterized 
because of the reparameterization of the scalar term $b_k$ in Eq.~(\ref{eq:mlr_reexpression}) as a vector $\mathbf{p}_k \in \mathbf{R}^n$. 
Since our definition of SPD hypergyroplanes follows that of Poincar\'e hyperplane~\cite{NEURIPS2018_dbab2adc}, 
our MLR suffers from the same problem. More precisely, 
in order to parameterize a SPD hypergyroplane, we use two symmetric matrices, 
i.e., $\mathbf{P}$ and $\mathbf{W}$ for each class (see Eq.~(\ref{eq:spd_hyperplanes})).  
Thus our MLR requires $n(n+1)K$ parameters, while a linear layer with input SPD matrices of the same size 
requires only $(n(n+1)/2+1)K$ parameters. 
This problem should be addressed in future work.



\section{Conclusion}
\label{sec:conclusion}

We have generalized the notions of inner product and gyroangles in gyrovector spaces for SPD and Grassmann manifolds. 
We have studied some isometric models on SPD and Grassmann manifolds, 
and reformulated MLR on SPD manifolds. 
We have compared our models against state-of-the-art models for the tasks of 
human action recognition and knowledge graph completion. 

\bibliography{references}
\bibliographystyle{icml2023}

\appendix
\onecolumn

\section{Experimental Details}
\label{sec:appendix_har_exp_settings}

\subsection{Human Action Recognition}

\paragraph{HDM05~\cite{hdm05}} 
It has 2337 sequences of 3D skeleton data classified into $130$ classes. 
Each frame contains the 3D coordinates of $31$ body joints.          
We use all the action classes and follow the experimental protocol of~\citet{Harandi2018DimensionalityRO} 
in which 2 subjects are used for training 
and the remaining 3 subjects are used for testing. 

\paragraph{FPHA~\cite{Garcia-HernandoCVPR18}} 
It has 1175 sequences of 3D skeleton data classified into $45$ classes. 
Each frame contains the 3D coordinates of $21$ hand joints.
We follow the experimental protocol of~\citet{Garcia-HernandoCVPR18} in which $600$ sequences are used for training 
and $575$ sequences are used for testing.

\paragraph{NTU60~\cite{Shahroudy16NTU}} 
It has 56880 sequences of 3D skeleton data classified into $60$ classes.
Each frame contains the 3D coordinates of $25$ or $50$ body joints.          
We use the mutual actions and follow the cross-subject experimental protocol of~\citet{Shahroudy16NTU} in which 
data from 20 subjects are used for training, and those from the other 20 subjects are used for testing.

\paragraph{SPD Neural Networks}
As in~\citet{HuangGool17,BrooksRieBatNorm19,NguyenNeurIPS22}, each sequence is represented by a covariance matrix.
The sizes of the covariance matrices are $93 \times 93$, $60 \times 60$, and $150 \times 150$ for 
HDM05, FPHA, and NTU60 datasets, respectively.   
For SPDNet, the same architecture as the one in~\citet{HuangGool17} is used with three Bimap layers. 
For SPDNetBN, the same architecture as the one in~\citet{BrooksRieBatNorm19} is used with three Bimap layers. 
For all the networks, the sizes of the transformation matrices 
for the experiments on HDM05, FPHA, and NTU60 datasets are set to 
$93 \times 93$, $60 \times 60$, and $150 \times 150$, respectively. 

\paragraph{Grassmann Neural Networks}
Following~\citet{HuangAAAI18}, action sequences are represented by linear subspaces of order 10 
which belong to $\operatorname{Gr}_{93,10}$, $\operatorname{Gr}_{60,10}$, and $\operatorname{Gr}_{150,10}$ 
for the experiments on HDM05, FPHA, and NTU datasets, respectively. 
For GrNet, the sizes of the connection weights are set respectively to $93 \times 93$, $60 \times 60$, and $150 \times 150$ 
for the experiments on HDM05, FPHA, and NTU datasets. 

Our networks are implemented with Pytorch framework. 
They are trained using cross-entropy loss and
Adadelta optimizer for 2000 epochs. 
The learning rate is set to $10^{-3}$. 
We use a batch size of 32 for HDM05 and FPHA datasets, and a batch size of 256 for NTU60 dataset.
We run each model three times and report the best accuracy from these three runs~\cite{NEURIPS2018_dbab2adc,NguyenNeurIPS22}. 

\subsection{Knowledge Graph Completion}

\paragraph{WN18RR~\cite{MillerWordNet95}}
It is a subset of WordNet~\cite{MillerWordNet95}, a hierarchical collection of relations between words, 
created from WN18~\cite{BordesMultiRel13} by removing the inverse of many relations from validation and 
test sets. 
It contains 40,943 entities and 11 relations.

\paragraph{FB15k-237~\cite{Toutanova2015RepresentingTF}}
It is a subset of Freebase~\cite{BollackerFreebase08}, a collection of real world facts, 
created in the same way as WN18RR from FB15k~\cite{BordesMultiRel13}. 
It contains 14,541 entities and 237 relations.

The networks are implemented with Pytorch framework. 
They are trained using binary cross-entropy loss and SGD optimizer for 2000 epochs.  
The learning rate is set to $10^{-3}$ with weight decay of $10^{-5}$.  
The batch size is set to $4096$. 
The number of negative samples is set to $10$. 
These settings are taken from~\citet{FedericoGyrocalculusSPD21}.   
We test with embeddings in $\operatorname{Gr}_{n,p}$ and $\widetilde{\operatorname{Gr}}_{n,p}$
where $(n,p) \in \{ (2k,k) \},k=5,6,\ldots,14$. The models give the best results with $(n,p)=(24,12)$.   
The MRR and hits at $K$ (H@K, $K=1,3,10$) are used as evaluation metrics~\cite{balazevic2019multi}. 
Early stopping is used when the MRR score of the model on the validation set does not improve after 500 epochs. 
In all experiments, the models that obtain the best
MRR scores on the validation set are used for testing. 

\section{Gyrogroups and Gyrovector Spaces}
\label{sec:gyrogroups_gyrovector_spaces}

Gyrovector spaces form the setting for hyperbolic geometry in the same way that vector spaces form the setting for Euclidean geometry~\cite{UngarBeyonEinstein02,UngarAnalyticHyp05,UngarHyperbolicNDim}. 
We recap the definitions of gyrogroups and gyrocommutative gyrogroups 
proposed in~\cite{UngarBeyonEinstein02,UngarAnalyticHyp05,UngarHyperbolicNDim}. 
For greater mathematical detail and in-depth discussion, we refer the interested reader to these papers.  

\begin{definition}[{\bf Gyrogroups~\cite{UngarHyperbolicNDim}}]\label{def:gyrovector_spaces}
A pair $(G,\oplus)$ is a groupoid in the sense that it is a nonempty set, $G$, with a binary operation, $\oplus$.  
A groupoid $(G,\oplus)$ is a gyrogroup if its binary operation satisfies the following axioms for $a,b,c \in G$:

(G1) There is at least one element $e \in G$ called a left identity such that $e \oplus a = a$. 

(G2) There is an element $\ominus a \in G$ called a left inverse of $a$ such that $\ominus a \oplus a = e$. 

(G3) There is an automorphism $\operatorname{gyr}[a,b]: G \rightarrow G$ for each $a,b \in G$ such that
\begin{equation*}
a \oplus (b \oplus c) = (a \oplus b) \oplus \operatorname{gyr}[a,b]c \hspace{3mm} \text{(Left Gyroassociative Law)}. 
\end{equation*}

The automorphism $\operatorname{gyr}[a,b]$ is called the gyroautomorphism, or the gyration of $G$ generated by $a,b$. 

(G4) $\operatorname{gyr}[a,b] = \operatorname{gyr}[a \oplus b,b]$ (Left Reduction Property). 
\end{definition}

\begin{definition}[{\bf Gyrocommutative Gyrogroups~\cite{UngarHyperbolicNDim}}]\label{def:gyrocommutative_gyrogroups}
A gyrogroup $(G,\oplus)$ is gyrocommutative if it satisfies
\begin{equation*}
a \oplus b = \operatorname{gyr}[a,b](b \oplus a) \hspace{3mm} \text{(Gyrocommutative Law).}
\end{equation*}
\end{definition}   

The following definition of gyrovector spaces is slightly different from Definition 3.2 in~\cite{UngarHyperbolicNDim}.        

\begin{definition}[{\bf Gyrovector Spaces}]\label{def:abstract_gyrovector_spaces}
A gyrocommutative gyrogroup $(G,\oplus)$ equipped with a scalar multiplication
\begin{equation*}
(t,x) \rightarrow t \odot x: \mathbb{R} \times G \rightarrow G
\end{equation*}
is called a gyrovector space if it satisfies the following axioms for $s,t \in \mathbb{R}$ and $a,b,c \in G$:

(V1) $1 \odot a = a, 0 \odot a = t \odot e = e,$ and $(-1) \odot a = \ominus a$.

(V2) $(s+t) \odot a = s \odot a \oplus t \odot a$.

(V3) $(st) \odot a = s \odot (t \odot a)$.

(V4) $\operatorname{gyr}[a,b](t \odot c) = t \odot \operatorname{gyr}[a,b]c$.  

(V5) $\operatorname{gyr}[s \odot a, t \odot a] = \operatorname{Id}$, where $\operatorname{Id}$ is the identity map.         
\end{definition}

\section{Gyrovector Spaces of SPD Matrices with a Log-Cholesky Geometry}
\label{sec:appendix_spd_gyrovector_spaces_from_diffeomorphism}

The recent work~\cite{NguyenECCV22} has shown the gyro-structure of SPD manifolds with a Log-Cholesky geometry~\cite{Lin_2019}.        
Here we present another method based on Lemmas~\ref{theorem:diffeo_binary_operation},~\ref{theorem:diffeo_scalar_multiplication}, 
and~\ref{theorem:diffeo_gyroautomorphism} for deriving closed-form expressions of the basic operations and gyroautomorphism 
of these manifolds. 
 
Using Eqs.~(\ref{eq:matrix_matrix_addition}),~(\ref{eq:scalar_matrix_multiplication}), and~(\ref{eq:basic_gyroautomorphism}), 
we first derive closed-form expressions of the basic operations and gyroautomorphism for $\operatorname{L}^+_n$, 
the space of $n \times n$ lower triangular matrices with positive diagonal entries.   

Let $\mathbf{U},\mathbf{V},\mathbf{W} \in \operatorname{L}^+_n$ and $t \in \mathbb{R}$. Then
\begin{equation*}
\mathbf{U} \oplus_{lt} \mathbf{V} = \lfloor \mathbf{U} \rfloor + \lfloor \mathbf{V} \rfloor + \mathbb{D}(\mathbf{U})\mathbb{D}(\mathbf{V}), 
\end{equation*}
\begin{equation*}
t \otimes_{lt} \mathbf{U} = t\lfloor \mathbf{U} \rfloor + \mathbb{D}(\mathbf{U})^t, 
\end{equation*}
\begin{equation*}
\operatorname{gyr}_{lt}[\mathbf{U},\mathbf{V}]\mathbf{W} = Id, 
\end{equation*}
where $\oplus_{lt}$, $\otimes_{lt}$, and $\operatorname{gyr}_{lt}[.,.]$ denote the binary operation, scalar multiplication, 
and gyroautomorphism of $\operatorname{L}^+_n$, respectively.  

As shown in~\citet{Lin_2019}, 
there exists a diffeomorphism between $\operatorname{L}^+_n$ and $\operatorname{Sym}^+_n$ given by:
\begin{equation*}
\xi: \operatorname{Sym}^+_n \rightarrow \operatorname{L}^+_n, \hspace{3mm} \mathbf{P} \rightarrow \mathbf{U}, \mathbf{U}\mathbf{U}^T=\mathbf{P}.  
\end{equation*}

This diffeomorphism gives us a simple way to obtain closed-form expressions of the basic operations 
and gyroautomorphism for SPD manifolds with a Log-Cholesky geometry, that is,
\begin{equation*}
\mathbf{P} \oplus_{lc} \mathbf{Q} = (\lfloor \varphi(\mathbf{P}) \rfloor + \lfloor \varphi(\mathbf{Q}) \rfloor + \mathbb{D}(\varphi(\mathbf{P}))\mathbb{D}(\varphi(\mathbf{Q}))).(\lfloor \varphi(\mathbf{P}) \rfloor + \lfloor \varphi(\mathbf{Q}) \rfloor + \mathbb{D}(\varphi(\mathbf{P}))\mathbb{D}(\varphi(\mathbf{Q})))^T,   
\end{equation*}
\begin{equation*}
t \otimes_{lc} \mathbf{P} = (t\lfloor \varphi(\mathbf{P}) \rfloor + \mathbb{D}(\varphi(\mathbf{P}))^t).(t\lfloor \varphi(\mathbf{P}) \rfloor + \mathbb{D}(\varphi(\mathbf{P}))^t)^T, 
\end{equation*}
\begin{equation*}
\operatorname{gyr}_{lc}[\mathbf{P},\mathbf{Q}]\mathbf{R} = Id, 
\end{equation*}
where $\mathbf{P},\mathbf{Q},\mathbf{R} \in \operatorname{Sym}^+_n$ and $t \in \mathbb{R}$.  


\section{The Law of SPD Gyrosines}
\label{appendix:law_of_spd_gyrosines}

\begin{theorem}[{\bf The Law of SPD Gyrosines}]\label{theorem:gyrosines_law}
Let $\mathbf{P},\mathbf{Q}$, and $\mathbf{R}$ be three distinct SPD gyropoints in a gyrovector space 
$(\operatorname{Sym}_n^+,\oplus_g,\otimes_g)$ where $g \in \{ le,lc \}$.        
Let $\widetilde{\mathbf{P}}=\ominus_g \mathbf{Q} \oplus_g \mathbf{R}$,  
$\widetilde{\mathbf{Q}}=\ominus_g \mathbf{P} \oplus_g \mathbf{R}$, and
$\widetilde{\mathbf{R}}=\ominus_g \mathbf{P} \oplus_g \mathbf{Q}$ be the SPD gyrosides of the SPD gyrotriangle 
formed by the three SPD gyropoints.  
Let $p = \| \widetilde{\mathbf{P}} \|$, $q = \| \widetilde{\mathbf{Q}} \|$, and $r = \| \widetilde{\mathbf{R}} \|$.  
Let $\alpha = \angle \mathbf{Q} \mathbf{P} \mathbf{R}$, 
$\beta = \angle \mathbf{P} \mathbf{Q} \mathbf{R}$, 
and $\gamma = \angle \mathbf{P} \mathbf{R} \mathbf{Q}$ be the SPD gyroangles of the SPD gyrotriangle. Then
\begin{equation*}
\frac{\sin(\alpha)}{p} = \frac{\sin(\beta)}{q} = \frac{\sin(\gamma)}{r}.  
\end{equation*}

\end{theorem}

\begin{proof}
This is a direct consequence of the Law of SPD gyrocosines.  
\end{proof}

\section{Proof of Lemma~\ref{theorem:diffeo_binary_operation}}
\label{sec:appendix_diffeo_binary_operation}

\begin{proof}

We first recall some results from~\citet{GallierQuaintance20}. 

\begin{proposition}[\cite{GallierQuaintance20}]\label{propo:local_isometry_preserve}
Let $M$ and $N$ be two Riemannian manifolds. 
If $\phi: M \rightarrow N$ is a local isometry, then the following concepts are preserved:

(1) Parallel translation along a curve. If $\mathcal{T}_{\delta}$ denotes parallel transport along the curve $\delta$ 
and if $\mathcal{T}_{\phi \circ \delta}$ denotes parallel transport along the curve $\phi \circ \delta$, then
\begin{equation}\label{eq:isometry_preserve_parallel_transport}
D \phi_{\delta(1)} \circ \mathcal{T}_{\delta} = \mathcal{T}_{\phi \circ \delta} \circ D \phi_{\delta(0)}.        
\end{equation}

(2) Exponential maps. We have
\begin{equation}\label{eq:isometry_preserve_exponential_maps}
\phi \circ \operatorname{Exp}_{\mathbf{P}} = \operatorname{Exp}_{\phi(\mathbf{P})} \circ D \phi_{\mathbf{P}}.        
\end{equation}
\end{proposition}

We also need to prove the following result.  
\begin{proposition}\label{propo:isometry_preserve_logarithmic_maps}
Let $\phi: M \rightarrow N$ be an isometry. Then
\begin{equation}
\operatorname{Log}_{\mathbf{P}}(\mathbf{Q}) = (D \phi^{-1}_{\phi(\mathbf{P})})(\operatorname{Log}_{\phi(\mathbf{P})}(\phi(\mathbf{Q}))).        
\end{equation}
\end{proposition}

\begin{proof}
Since $\phi$ is an isometry, its inverse $\phi^{-1}$ is an isometry. Therefore, from Eq.~(\ref{eq:isometry_preserve_exponential_maps}) 
we have
\begin{align*}
\begin{split}
\phi^{-1} \circ \operatorname{Exp}_{\phi(\mathbf{P})}(\operatorname{Log}_{\phi(\mathbf{P})}(\phi(\mathbf{Q}))) &= \operatorname{Exp}_{\phi^{-1}(\phi(\mathbf{P}))} \circ (D \phi^{-1}_{\phi(\mathbf{P})}) (\operatorname{Log}_{\phi(\mathbf{P})}(\phi(\mathbf{Q}))) \\ &= \operatorname{Exp}_{\mathbf{P}} \circ (D \phi^{-1}_{\phi(\mathbf{P})}) (\operatorname{Log}_{\phi(\mathbf{P})}(\phi(\mathbf{Q}))).
\end{split}
\end{align*}

Hence
\begin{equation*}
\phi^{-1} \circ \phi(\mathbf{Q}) = \operatorname{Exp}_{\mathbf{P}} \circ (D \phi^{-1}_{\phi(\mathbf{P})}) (\operatorname{Log}_{\phi(\mathbf{P})}(\phi(\mathbf{Q}))),
\end{equation*}
which is equivalent to
\begin{equation*}
\mathbf{Q} = \operatorname{Exp}_{\mathbf{P}} \circ (D \phi^{-1}_{\phi(\mathbf{P})}) (\operatorname{Log}_{\phi(\mathbf{P})}(\phi(\mathbf{Q}))).
\end{equation*}

Therefore
\begin{equation*}
\operatorname{Log}_{\mathbf{P}}(\mathbf{Q}) = (D \phi^{-1}_{\phi(\mathbf{P})})(\operatorname{Log}_{\phi(\mathbf{P})}(\phi(\mathbf{Q}))).        
\end{equation*}

\end{proof}

According to the definition of the binary operation $\oplus_m$ in Eq.~(\ref{eq:matrix_matrix_addition}),
\begin{align}\label{eq:proof_diffeomorphism_binary_operation}
\begin{split}
\mathbf{P} \oplus_m \mathbf{Q} &= \operatorname{Exp}_{\mathbf{P}}(\mathcal{T}_{\bar{\mathbf{I}} \rightarrow \mathbf{P}}(\operatorname{Log}_{\bar{\mathbf{I}}}(\mathbf{Q}))) \\ &\overset{(1)}{=} \operatorname{Exp}_{\mathbf{P}}(\mathcal{T}_{\bar{\mathbf{I}} \rightarrow \mathbf{P}}((D \phi^{-1}_{\phi(\bar{\mathbf{I}})})(\operatorname{Log}_{\phi(\bar{\mathbf{I}})}(\phi(\mathbf{Q}))))) \\ &\overset{(2)}{=} \operatorname{Exp}_{\mathbf{P}} \Big( (D \phi^{-1}_{\phi(\mathbf{P})}) \big( \mathcal{T}_{\phi(\bar{\mathbf{I}}) \rightarrow \phi(\mathbf{P})}(\operatorname{Log}_{\phi(\bar{\mathbf{I}})}(\phi(\mathbf{Q}))) \big) \Big) \\ &\overset{(3)}{=} \phi^{-1} \Big( \operatorname{Exp}_{\phi(\mathbf{P})} \big( \mathcal{T}_{\phi(\bar{\mathbf{I}}) \rightarrow \phi(\mathbf{P})}(\operatorname{Log}_{\phi(\bar{\mathbf{I}})}(\phi(\mathbf{Q}))) \big) \Big) \\ &\overset{(4)}{=} \phi^{-1} (\phi(\mathbf{P}) \oplus_n \phi(\mathbf{Q})). 
\end{split}
\end{align}

The derivation of Eq.~(\ref{eq:proof_diffeomorphism_binary_operation}) follows. 

(1) follows from Proposition~\ref{propo:isometry_preserve_logarithmic_maps}. 

(2) follows from Eq.~(\ref{eq:isometry_preserve_parallel_transport}). 

(3) follows from Eq.~(\ref{eq:isometry_preserve_exponential_maps}). 

(4) follows from the definition of the binary operation $\oplus_n$. 

\end{proof}

\section{Proof of Lemma~\ref{theorem:diffeo_scalar_multiplication}}
\label{sec:appendix_diffeo_scalar_multiplication}

\begin{proof}
According to the definition of the scalar multiplication $\otimes_m$ in Eq.~(\ref{eq:scalar_matrix_multiplication}),
\begin{align}\label{eq:proof_diffeomorphism_scalar_multiplication}
\begin{split}
t \otimes_m \mathbf{P} &= \operatorname{Exp}_{\bar{\mathbf{I}}}(t\operatorname{Log}_{\bar{\mathbf{I}}} (\mathbf{P})) \\ &\overset{(1)}{=} \operatorname{Exp}_{\bar{\mathbf{I}}} \big( t (D \phi^{-1}_{\phi(\bar{\mathbf{I}})}) (\operatorname{Log}_{\phi(\bar{\mathbf{I}})}(\phi(\mathbf{P}))) \big) \\ &\overset{(2)}{=} \operatorname{Exp}_{\bar{\mathbf{I}}} \big( (D \phi^{-1}_{\phi(\bar{\mathbf{I}})}) (t\operatorname{Log}_{\phi(\bar{\mathbf{I}})}(\phi(\mathbf{P}))) \big) \\ &\overset{(3)}{=} \phi^{-1} \big( \operatorname{Exp}_{\phi(\bar{\mathbf{I}})}(t\operatorname{Log}_{\phi(\bar{\mathbf{I}})}(\phi(\mathbf{P}))) \big) \\ &\overset{(4)}{=} \phi^{-1}(t \otimes_n \phi(\mathbf{P})).        
\end{split}
\end{align}

The derivation of Eq.~(\ref{eq:proof_diffeomorphism_scalar_multiplication}) follows.   

(1) follows from Proposition~\ref{propo:isometry_preserve_logarithmic_maps}. 

(2) follows from the fact that $D \phi^{-1}$ is a linear operator. 

(3) follows from Eq.~(\ref{eq:isometry_preserve_exponential_maps}). 

(4) follows from the definition of the scalar multiplication $\otimes_n$. 

\end{proof}

\section{Proof of Lemma~\ref{theorem:diffeo_gyroautomorphism}}
\label{sec:appendix_diffeo_gyroautomorphism}

\begin{proof}

For any $\mathbf{Y} \in M$, we have
\begin{align}\label{eq:diffeo_gyroautomorphism_identity_element}
\begin{split}
\phi(\bar{\mathbf{I}}) &= \phi(\ominus_m \mathbf{Y} \oplus_m \mathbf{Y}) \\ &\overset{(1)}{=} \phi(\ominus_m \mathbf{Y}) \oplus_n \phi(\mathbf{Y}), 
\end{split}
\end{align}
where (1) follows from Eq.~(\ref{eq:diffeo_binary_operation}). 

Note that
\begin{align}\label{eq:diffeo_gyroautomorphism_chain_equation}
\begin{split}
\operatorname{gyr}_m[\mathbf{P},\mathbf{Q}]\mathbf{R} &\overset{(1)}{=} \big( \ominus_m (\mathbf{P} \oplus_m \mathbf{Q}) \big) \oplus_m \big( \mathbf{P} \oplus_m (\mathbf{Q} \oplus_m \mathbf{R}) \big) \\ &\overset{(2)}{=} \phi^{-1} \Big( \phi \big( \ominus_m (\mathbf{P} \oplus_m \mathbf{Q}) \big) \oplus_n \phi \big( \mathbf{P} \oplus_m (\mathbf{Q} \oplus_m \mathbf{R}) \big) \Big) \\ &\overset{(3)}{=} \phi^{-1} \Big( \ominus_n \big( \phi (\mathbf{P} \oplus_m \mathbf{Q}) \big) \oplus_n \phi \big( \mathbf{P} \oplus_m (\mathbf{Q} \oplus_m \mathbf{R}) \big) \Big) \\ &\overset{(4)}{=} \phi^{-1} \Big( \ominus_n \big( \phi (\mathbf{P}) \oplus_n \phi (\mathbf{Q}) \big) \oplus_n \phi \big( \mathbf{P} \oplus_m (\mathbf{Q} \oplus_m \mathbf{R}) \big) \Big) \\ &\overset{(5)}{=} \phi^{-1} \Big( \ominus_n \big( \phi (\mathbf{P}) \oplus_n \phi (\mathbf{Q}) \big) \oplus_n  \big( \phi (\mathbf{P}) \oplus_n \phi (\mathbf{Q} \oplus_m \mathbf{R}) \big) \Big) \\ &\overset{(6)}{=} \phi^{-1} \Big( \ominus_n \big( \phi (\mathbf{P}) \oplus_n \phi (\mathbf{Q}) \big) \oplus_n  \big( \phi (\mathbf{P}) \oplus_n ( \phi (\mathbf{Q}) \oplus_n \phi (\mathbf{R})) \big) \Big) \\ &\overset{(7)}{=} \phi^{-1} ( \operatorname{gyr}_n[\phi (\mathbf{P}),\phi (\mathbf{Q})]\phi (\mathbf{R}) ).   
\end{split}
\end{align}

The derivation of Eq.~(\ref{eq:diffeo_gyroautomorphism_chain_equation}) follows. 

(1) follows from Eq.~(\ref{eq:basic_gyroautomorphism}).

(2) follows from Eq.~(\ref{eq:diffeo_binary_operation}).

(3) follows from Eq.~(\ref{eq:diffeo_gyroautomorphism_identity_element}).   

(4), (5), and (6) follow from Eq.~(\ref{eq:diffeo_binary_operation}). 



(7) follows from Eq.~(\ref{eq:basic_gyroautomorphism}).

\end{proof}

\section{Proof of Theorem~\ref{theorem:gyrovector_spaces_from_diffeomorphism}}
\label{sec:appendix_gyrovector_spaces_from_diffeomorphism}


{\noindent \bf Axiom (G1)}

\begin{proof}

Let $\phi(\bar{\mathbf{I}})$ be a left identity in $G_n$ where $\bar{\mathbf{I}} \in G_m$. Then for $\mathbf{P} \in G_m$, we have
\begin{equation*}
\phi(\mathbf{P}) = \phi(\bar{\mathbf{I}}) \oplus_n \phi(\mathbf{P}) = \phi(\bar{\mathbf{I}} \oplus_m \mathbf{P}),
\end{equation*}
which shows that $\mathbf{P} = \bar{\mathbf{I}} \oplus_m \mathbf{P}$ and therefore $\bar{\mathbf{I}}$ is a left identity in $G_m$.  

\end{proof}

{\noindent \bf Axiom (G2)}

\begin{proof}

For $\mathbf{P} \in G_m$, by the assumption that $(G_n,\oplus_n,\otimes_n)$ is a gyrovector space, 
there exists a left inverse $\ominus_n \phi(\mathbf{P})$ of $\phi(\mathbf{P})$ such that
\begin{equation*}
\ominus_n \phi(\mathbf{P}) \oplus_n \phi(\mathbf{P}) = \phi(\bar{\mathbf{I}}).
\end{equation*}

Hence
\begin{equation*}
\bar{\mathbf{I}} = \phi^{-1}( \ominus_n \phi(\mathbf{P}) \oplus_n \phi(\mathbf{P}) ) = \phi^{-1}(\ominus_n \phi(\mathbf{P})) \oplus_m \mathbf{P}, 
\end{equation*}
which shows that $\phi^{-1}(\ominus_n \phi(\mathbf{P})) \in G_m$ is a left inverse of $\mathbf{P}$.  

\end{proof}

{\noindent \bf Axiom (G3)}

\begin{proof}

For $\mathbf{P},\mathbf{Q},\mathbf{R} \in G_m$, we have
\begin{align}\label{eq:axiom_g3_proof}
\begin{split}
\mathbf{P} \oplus_m (\mathbf{Q} \oplus_m \mathbf{R}) &\overset{(1)}{=} \mathbf{P} \oplus_m \phi^{-1}(\phi(\mathbf{Q}) \oplus_n \phi(\mathbf{R})) \\ &\overset{(2)}{=} \phi^{-1} \big( \phi(\mathbf{P}) \oplus_n (\phi(\mathbf{Q}) \oplus_n \phi(\mathbf{R})) \big) \\ &\overset{(3)}{=} \phi^{-1} \big( (\phi(\mathbf{P}) \oplus_n \phi(\mathbf{Q})) \oplus_n \operatorname{gyr}_n[\phi(\mathbf{P}),\phi(\mathbf{Q})]\phi(\mathbf{R}) \big) \\ &\overset{(4)}{=} \phi^{-1} \big( (\phi(\mathbf{P}) \oplus_n \phi(\mathbf{Q})) \oplus_n \phi(\operatorname{gyr}_m[\mathbf{P},\mathbf{Q}]\mathbf{R}) \big) \\ &\overset{(5)}{=} \phi^{-1}(\phi(\mathbf{P}) \oplus_n \phi(\mathbf{Q})) \oplus_m \operatorname{gyr}_m[\mathbf{P},\mathbf{Q}]\mathbf{R} \\ &\overset{(6)}{=} (\mathbf{P} \oplus_m \mathbf{Q}) \oplus_m \operatorname{gyr}_m[\mathbf{P},\mathbf{Q}]\mathbf{R}.
\end{split}
\end{align}

The derivation of Eq.~(\ref{eq:axiom_g3_proof}) follows.

(1) follows from the definition of the binary operation $\oplus_m$.

(2) follows from the definition of the binary operation $\oplus_m$.

(3) follows from Axiom (G3) verified by gyrovector space $(G_n,\oplus_n,\otimes_n)$.

(4) follows from the definition of the gyroautomorphism $\operatorname{gyr}_m[.,.]$.

(5) follows from the definition of the binary operation $\oplus_m$.

(6) follows from the definition of the binary operation $\oplus_m$.  

\end{proof}

{\noindent \bf Axiom (G4)}

\begin{proof}

For $\mathbf{P},\mathbf{Q},\mathbf{R} \in G_m$, we have
\begin{align}\label{eq:axiom_g4_proof}
\begin{split}
\operatorname{gyr}_m[\mathbf{P},\mathbf{Q}]\mathbf{R} &\overset{(1)}{=} \phi^{-1}\big( \operatorname{gyr}_n[\phi(\mathbf{P}),\phi(\mathbf{Q})]\phi(\mathbf{R}) \big) \\ &\overset{(2)}{=} \phi^{-1}\big( \operatorname{gyr}_n[\phi(\mathbf{P}) \oplus_n \phi(\mathbf{Q}),\phi(\mathbf{Q})]\phi(\mathbf{R}) \big) \\ &\overset{(3)}{=} \phi^{-1}\big( \operatorname{gyr}_n[\phi(\mathbf{P} \oplus_m \mathbf{Q}),\phi(\mathbf{Q})]\phi(\mathbf{R}) \big) \\ &\overset{(4)}{=} \phi^{-1}\big( \phi( \operatorname{gyr}_m[\mathbf{P} \oplus_m \mathbf{Q},\mathbf{Q}]\mathbf{R} ) \big) \\ &\overset{(5)}{=} \operatorname{gyr}_m[\mathbf{P} \oplus_m \mathbf{Q},\mathbf{Q}]\mathbf{R}.
\end{split}
\end{align}

The derivation of Eq.~(\ref{eq:axiom_g4_proof}) follows.

(1) follows from the definition of the gyroautomorphism $\operatorname{gyr}_m[.,.]$.

(2) follows from Axiom (G4) verified by gyrovector space $(G_n,\oplus_n,\otimes_n)$.

(3) follows from the definition of the binary operation $\oplus_m$.

(4) follows from the definition of the gyroautomorphism $\operatorname{gyr}_m[.,.]$.

\end{proof}

{\noindent \bf Gyrocommutative Law}

\begin{proof}

For $\mathbf{P},\mathbf{Q} \in G_m$, we have
\begin{align}\label{eq:axiom_gyrocommutative_proof}
\begin{split}
\mathbf{P} \oplus_m \mathbf{Q} &\overset{(1)}{=} \phi^{-1}\big( \phi(\mathbf{P}) \oplus_n \phi(\mathbf{Q}) \big) \\ &\overset{(2)}{=} \phi^{-1}\big( \operatorname{gyr}_n[\phi(\mathbf{P}),\phi(\mathbf{Q})](\phi(\mathbf{Q}) \oplus_n \phi(\mathbf{P})) \big) \\ &\overset{(3)}{=} \phi^{-1}\big( \operatorname{gyr}_n[\phi(\mathbf{P}),\phi(\mathbf{Q})]\phi(\mathbf{Q} \oplus_m \mathbf{P}) \big) \\ &\overset{(4)}{=} \operatorname{gyr}_m[\mathbf{P},\mathbf{Q}](\mathbf{Q} \oplus_m \mathbf{P}).
\end{split}
\end{align}

The derivation of Eq.~(\ref{eq:axiom_gyrocommutative_proof}) follows.

(1) follows from the definition of the binary operation $\oplus_m$.

(2) follows from the Gyrocommutative Law verified by gyrovector space $(G_n,\oplus_n,\otimes_n)$. 

(3) follows from the definition of the binary operation $\oplus_m$.

(4) follows from the definition of the gyroautomorphism $\operatorname{gyr}_m[.,.]$.  

\end{proof}

{\noindent \bf Axiom (V1)}

\begin{proof}

For $t \in \mathbb{R}$ and $\mathbf{P} \in G_m$, 
by the assumption that $(G_n,\oplus_n,\otimes_n)$ is a gyrovector space and 
from Eqs.~(\ref{eq:diffeo_scalar_multiplication}) and~(\ref{eq:diffeo_gyroautomorphism_identity_element}), we have
\begin{equation*}
1 \otimes_m \mathbf{P} = \phi^{-1}(1 \otimes_n \phi(\mathbf{P})) = \phi^{-1}(\phi(\mathbf{P})) = \mathbf{P}.
\end{equation*}

\begin{equation*}
0 \otimes_m \mathbf{P} = \phi^{-1}(0 \otimes_n \phi(\mathbf{P})) = \phi^{-1}(\phi(\bar{\mathbf{I}})) = \bar{\mathbf{I}}. 
\end{equation*}

\begin{equation*}
t \otimes_m \bar{\mathbf{I}} = \phi^{-1}(t \otimes_n \phi(\bar{\mathbf{I}})) = \phi^{-1}(\phi(\bar{\mathbf{I}})) = \bar{\mathbf{I}}.         
\end{equation*}

\begin{equation*}
(-1) \otimes_m \mathbf{P} = \phi^{-1}((-1) \otimes_n \phi(\mathbf{P})) = \phi^{-1}(\ominus_n \phi(\mathbf{P})) = \phi^{-1}(\phi(\ominus_m \mathbf{P})) = \ominus_m \mathbf{P}.        
\end{equation*}


\end{proof}

{\noindent \bf Axiom (V2)}

\begin{proof}

For $s,t \in \mathbb{R}$ and $\mathbf{P} \in G_m$, we have
\begin{align}\label{eq:diffeo_gyrovector_spaces_v2}
\begin{split}
(s+t) \otimes_m \mathbf{P} &\overset{(1)}{=} \phi^{-1}((s+t) \otimes_n \phi(\mathbf{P})) \\ &\overset{(2)}{=} \phi^{-1} (s \otimes_n \phi(\mathbf{P}) \oplus_n t \otimes_n \phi(\mathbf{P})) \\ &\overset{(3)}{=} \phi^{-1} ( \phi(\phi^{-1}( s \otimes_n \phi(\mathbf{P}) )) \oplus_n \phi(\phi^{-1}( t \otimes_n \phi(\mathbf{P}) )) ) \\ &\overset{(4)}{=} \phi^{-1}(s \otimes_n \phi(\mathbf{P})) \oplus_m \phi^{-1}(t \otimes_n \phi(\mathbf{P})) \\ &\overset{(5)}{=} s \otimes_m \mathbf{P} \oplus_m t \otimes_m \mathbf{P}. 
\end{split}
\end{align}

The derivation of Eq.~(\ref{eq:diffeo_gyrovector_spaces_v2}) follows.   

(1) follows from the definition of the scalar multiplication $\otimes_m$. 

(2) follows from Axiom (V2) verified by gyrovector space $(G_n,\oplus_n,\otimes_n)$. 

(3) follows from the fact that $\phi$ is an isometry. 

(4) follows from the definition of the binary operation $\oplus_m$. 

(5) follows from the definition of the scalar multiplication $\otimes_m$. 

\end{proof}

{\noindent \bf Axiom (V3)}

\begin{proof}

For $s,t \in \mathbb{R}$ and $\mathbf{P} \in G_m$, we have
\begin{align}\label{eq:diffeo_gyrovector_spaces_v3}
\begin{split}
(st) \otimes_m \mathbf{P} &\overset{(1)}{=} \phi^{-1}( (st) \otimes_n \phi(\mathbf{P}) ) \\ &\overset{(2)}{=} \phi^{-1}( s \otimes_n (t \otimes_n \phi(\mathbf{P})) ) \\ &\overset{(3)}{=} \phi^{-1}( s \otimes_n \phi(\phi^{-1} (t \otimes_n \phi(\mathbf{P})) ) ) \\ &\overset{(4)}{=} s \otimes_m \phi^{-1}(t \otimes_n \phi(\mathbf{P})) \\ &\overset{(5)}{=} s \otimes_m (t \otimes_m \mathbf{P}). 
\end{split}
\end{align}

The derivation of Eq.~(\ref{eq:diffeo_gyrovector_spaces_v3}) follows.   

(1) follows from the definition of the scalar multiplication $\otimes_m$. 

(2) follows from Axiom (V3) verified by gyrovector space $(G_n,\oplus_n,\otimes_n)$. 

(3) follows from the fact that $\phi$ is an isometry. 

(4) follows from the definition of the scalar multiplication $\otimes_m$. 

(5) follows from the definition of the scalar multiplication $\otimes_m$. 

\end{proof}

{\noindent \bf Axiom (V4)}

\begin{proof}

For $t \in \mathbb{R}$ and $\mathbf{P},\mathbf{Q},\mathbf{R} \in G_m$, we have
\begin{align}\label{eq:diffeo_gyrovector_spaces_v4}
\begin{split}
\operatorname{gyr}_m[\mathbf{P},\mathbf{Q}](t \otimes_m \mathbf{R}) &\overset{(1)}{=} \phi^{-1}( \operatorname{gyr}_n[\phi(\mathbf{P}),\phi(\mathbf{Q})] \phi(t \otimes_m \mathbf{R}) ) \\ &\overset{(2)}{=} \phi^{-1}( \operatorname{gyr}_n[\phi(\mathbf{P}),\phi(\mathbf{Q})] \phi(\phi^{-1}(t \otimes_n \phi(\mathbf{R}))) ) \\ &\overset{(3)}{=} \phi^{-1}( \operatorname{gyr}_n[\phi(\mathbf{P}),\phi(\mathbf{Q})] (t \otimes_n \phi(\mathbf{R})) ) \\ &\overset{(4)}{=} \phi^{-1} (t \otimes_n \operatorname{gyr}_n[\phi(\mathbf{P}),\phi(\mathbf{Q})]\phi(\mathbf{R})) \\ &\overset{(5)}{=} \phi^{-1} (t \otimes_n \phi(\phi^{-1}( \operatorname{gyr}_n[\phi(\mathbf{P}),\phi(\mathbf{Q})]\phi(\mathbf{R})))) \\ &\overset{(6)}{=} t \otimes_m \phi^{-1}(\operatorname{gyr}_n[\phi(\mathbf{P}),\phi(\mathbf{Q})]\phi(\mathbf{R})) \\ &\overset{(7)}{=} t \otimes_m \operatorname{gyr}_m[\mathbf{P},\mathbf{Q}]\mathbf{R}.        
\end{split}
\end{align}

The derivation of Eq.~(\ref{eq:diffeo_gyrovector_spaces_v4}) follows.   

(1) follows from the definition of the gyroautomorphism $\operatorname{gyr}_m[.,.]$. 

(2) follows from the definition of the scalar multiplication $\otimes_m$. 

(3) follows from the fact that $\phi$ is an isometry. 

(4) follows from Axiom (V4) verified by gyrovector space $(G_n,\oplus_n,\otimes_n)$. 

(5) follows from the fact that $\phi$ is an isometry. 

(6) follows from the definition of the scalar multiplication $\otimes_m$. 

(7) follows from the definition of the gyroautomorphism $\operatorname{gyr}_m[.,.]$. 

\end{proof}

{\noindent \bf Axiom (V5)}

\begin{proof}

For $s,t \in \mathbb{R}$ and $\mathbf{P},\mathbf{Q} \in G_m$, we have
\begin{align}\label{eq:diffeo_gyrovector_spaces_v5}
\begin{split}
\operatorname{gyr}_m[s \otimes_m \mathbf{P},t \otimes_m \mathbf{P}]\mathbf{Q} &\overset{(1)}{=} \phi^{-1}( \operatorname{gyr}_n[\phi(s \otimes_m \mathbf{P}),\phi(t \otimes_m \mathbf{P})] \phi(\mathbf{Q}) ) \\ &\overset{(2)}{=} \phi^{-1}( \operatorname{gyr}_n[\phi( \phi^{-1}( s \otimes_n \phi(\mathbf{P})) ),\phi( \phi^{-1}( t \otimes_n \phi(\mathbf{P})) )] \phi(\mathbf{Q}) ) \\ &\overset{(3)}{=} \phi^{-1}( \operatorname{gyr}_n[s \otimes_n \phi(\mathbf{P}), t \otimes_n \phi(\mathbf{P})] \phi(\mathbf{Q}) ) \\ &\overset{(4)}{=} \phi^{-1}( \phi(\mathbf{Q}) ) \\ &\overset{(5)}{=} \mathbf{Q}.         
\end{split}
\end{align}

The derivation of Eq.~(\ref{eq:diffeo_gyrovector_spaces_v5}) follows.   

(1) follows from the definition of the gyroautomorphism $\operatorname{gyr}_m[.,.]$. 

(2) follows from the definition of the scalar multiplication $\otimes_m$. 

(3) follows from the fact that $\phi$ is an isometry. 

(4) follows from Axiom (V5) verified by gyrovector space $(G_n,\oplus_n,\otimes_n)$. 

(5) follows from the fact that $\phi$ is an isometry. 

\end{proof}







\section{Proof of Theorem~\ref{theorem:left_gyrotranslational_isometry_gr}}
\label{sec:appendix_left_gyrotranslational_isometry_gr}

\begin{proof}

We first prove the following lemma: 

\begin{lemma}\label{lemma:left_gyrotranslation_gr}
Gyrogroups $(\operatorname{Gr}_{n,p},\oplus_{gr})$ verify the Left Gyrotranslation Law~\cite{UngarHyperbolicNDim}, that is,
\begin{equation*}
\ominus_{gr} (\mathbf{P} \oplus_{gr} \mathbf{Q}) \oplus_{gr} (\mathbf{P} \oplus_{gr} \mathbf{R}) = \operatorname{gyr}[\mathbf{P},\mathbf{Q}](\ominus_{gr} \mathbf{Q} \oplus_{gr} \mathbf{R}),
\end{equation*} 
where $\mathbf{P},\mathbf{Q},\mathbf{R} \in \operatorname{Gr}_{n,p}$.  
\end{lemma}

\begin{proof}
First, note that gyrogroups $(\operatorname{Gr}_{n,p},\oplus_{gr})$ verify the Left Cancellation Law~\cite{UngarHyperbolicNDim}, i.e., 
\begin{equation*}
\ominus_{gr} \mathbf{P} \oplus_{gr} (\mathbf{P} \oplus_{gr} \mathbf{Q}) = \mathbf{Q},
\end{equation*}
where $\mathbf{P},\mathbf{Q} \in \operatorname{Gr}_{n,p}$. 

We have
\begin{align*}
\begin{split}
(\mathbf{P} \oplus_{gr} \mathbf{Q}) \oplus_{gr} \operatorname{gyr}[\mathbf{P},\mathbf{Q}](\ominus_{gr} \mathbf{Q} \oplus_{gr} \mathbf{R}) &\overset{(1)}{=} \mathbf{P} \oplus_{gr} (\mathbf{Q} \oplus_{gr} (\ominus_{gr} \mathbf{Q} \oplus_{gr} \mathbf{R})) \\&\overset{(2)}{=} \mathbf{P} \oplus_{gr} \mathbf{R},
\end{split}
\end{align*}
where (1) follows from the Left Gyroassociative Law, and (2) follows from the Left Cancellation Law. 
Hence
\begin{align*}
\begin{split}
\ominus_{gr} (\mathbf{P} \oplus_{gr} \mathbf{Q}) \oplus_{gr} (\mathbf{P} \oplus_{gr} \mathbf{R}) &= \ominus_{gr} (\mathbf{P} \oplus_{gr} \mathbf{Q}) \oplus_{gr} ( (\mathbf{P} \oplus_{gr} \mathbf{Q}) \oplus_{gr} \operatorname{gyr}[\mathbf{P},\mathbf{Q}](\ominus_{gr} \mathbf{Q} \oplus_{gr} \mathbf{R}) ) \\&\overset{(1)}{=} \operatorname{gyr}[\mathbf{P},\mathbf{Q}](\ominus_{gr} \mathbf{Q} \oplus_{gr} \mathbf{R}),
\end{split}
\end{align*}
where (1) follows from the Left Cancellation Law.
\end{proof}

We also need to prove the following lemma:

\begin{lemma}\label{lemma:gyroautomorphisms_preserve_norm_gr}
Gyroautomorphisms $\operatorname{gyr}_{gr}[.,.]$ preserve the norm. 
\end{lemma}

\begin{proof}

Denote by $\operatorname{O}_n$ the space of $n \times n$ orthogonal matrices, 
$\langle . \rangle_F$ the Frobenius inner product, 
$\| . \|_F$ the Frobenius norm.  
Let $\mathbf{P},\mathbf{Q},\mathbf{R} \in \operatorname{Gr}_{n,p}$. Then notice that
\begin{equation*}
\operatorname{gyr}_{gr}[\mathbf{P},\mathbf{Q}]\mathbf{R} = \begin{bmatrix} \mathbf{O}_1 & 0 \\ 0 & \mathbf{O}_2 \end{bmatrix} \mathbf{R} \begin{bmatrix} \mathbf{O}_1 & 0 \\ 0 & \mathbf{O}_2 \end{bmatrix}^T,
\end{equation*}
where $\mathbf{O}_1 \in \operatorname{O}_p$, $\mathbf{O}_2 \in \operatorname{O}_{n-p}$. 
Let $\mathbf{O} = \begin{bmatrix} \mathbf{O}_1 & 0 \\ 0 & \mathbf{O}_2 \end{bmatrix}$. Then


\begin{align}\label{eq:gyroautomorphisms_preserve_norm_gr_final}
\begin{split}
\| \operatorname{gyr}_{gr}[\mathbf{P},\mathbf{Q}]\mathbf{R} \| &= \| \operatorname{Log}^{gr}_{\mathbf{I}_{n,p}}(\operatorname{gyr}_{gr}[\mathbf{P},\mathbf{Q}]\mathbf{R}) \|_F \\ &\overset{(1)}{=} \| \operatorname{Log}^{gr}_{\mathbf{O} \mathbf{I}_{n,p} \mathbf{O}^T}(\mathbf{O} \mathbf{R} \mathbf{O}^T) \|_F \\ &\overset{(2)}{=} \| \mathbf{O} \operatorname{Log}^{gr}_{\mathbf{I}_{n,p}}(\mathbf{R}) \mathbf{O}^T \|_F \\ &= \| \operatorname{Log}^{gr}_{\mathbf{I}_{n,p}}(\mathbf{R}) \|_F \\ &= \| \mathbf{R} \|.  
\end{split}
\end{align}

The derivation of Eq.~(\ref{eq:gyroautomorphisms_preserve_norm_gr_final}) follows. 

(1) follows from the fact that $\mathbf{I}_{n,p} = \mathbf{O} \mathbf{I}_{n,p} \mathbf{O}^T$.

(2) follows from~\citet{NguyenNeurIPS22} (see Lemma 3.19).  


\end{proof}

We now have the following chain of equations:
\begin{align*}
\begin{split}
\| \ominus_{gr} (\mathbf{A} \oplus_{gr} \mathbf{P}) \oplus_{gr} (\mathbf{A} \oplus_{gr} \mathbf{Q}) \| &\overset{(1)}{=} \| \operatorname{gyr}[\mathbf{A},\mathbf{P}](\ominus_{gr} \mathbf{P} \oplus_{gr} \mathbf{Q}) \| \\&\overset{(2)}{=} \| \ominus_{gr} \mathbf{P} \oplus_{gr} \mathbf{Q} \|, 
\end{split}
\end{align*}
where (1) follows from the Left Gyrotranslation Law, 
and (2) follows from the invariance of the norm under gyroautomorphisms (Lemma~\ref{lemma:gyroautomorphisms_preserve_norm_gr}). 

\end{proof}

\section{Proof of Theorem~\ref{theorem:gyroautomorphisms_gyroisometries_gr}}
\label{sec:appendix_gyroautomorphisms_gyroisometries_gr}

\begin{proof}
Let $\mathbf{P},\mathbf{Q},\mathbf{R},\mathbf{S} \in \operatorname{Gr}_{n,p}$. 
Then by the Left Gyroassociative Law and Left Cancellation Law,
\begin{equation*}
\operatorname{gyr}[\mathbf{P},\mathbf{Q}]\mathbf{R} = \ominus_{gr} (\mathbf{P} \oplus_{gr} \mathbf{Q}) \oplus_{gr} (\mathbf{P} \oplus_{gr} (\mathbf{Q} \oplus_{gr} \mathbf{R})), 
\end{equation*}
\begin{equation*}
\operatorname{gyr}[\mathbf{P},\mathbf{Q}]\mathbf{S} = \ominus_{gr} (\mathbf{P} \oplus_{gr} \mathbf{Q}) \oplus_{gr} (\mathbf{P} \oplus_{gr} (\mathbf{Q} \oplus_{gr} \mathbf{S})).
\end{equation*}

Let $\mathbf{X} = \ominus_{gr} (\mathbf{P} \oplus_{gr} \mathbf{Q})$, $\mathbf{Y} = \mathbf{P} \oplus_{gr} (\mathbf{Q} \oplus_{gr} \mathbf{R})$, and $\mathbf{Z} = \mathbf{P} \oplus_{gr} (\mathbf{Q} \oplus_{gr} \mathbf{S})$. 
Then we have the following chain of equations:
\begin{align}\label{eq:gyroautomorphisms_preserve_gyrodistance}
\begin{split}
d(\operatorname{gyr}[\mathbf{P},\mathbf{Q}]\mathbf{R},\operatorname{gyr}[\mathbf{P},\mathbf{Q}]\mathbf{S}) &= \| \ominus_{gr} \operatorname{gyr}[\mathbf{P},\mathbf{Q}]\mathbf{R} \oplus_{gr} \operatorname{gyr}[\mathbf{P},\mathbf{Q}]\mathbf{S} \| \\&= \| \ominus_{gr} (\mathbf{X} \oplus_{gr} \mathbf{Y}) \oplus_{gr} (\mathbf{X} \oplus_{gr} \mathbf{Z}) \| \\&\overset{(1)}{=} \| \operatorname{gyr}[\mathbf{X},\mathbf{Y}](\ominus_{gr} \mathbf{Y} \oplus_{gr} \mathbf{Z}) \| \\&\overset{(2)}{=} \| \ominus_{gr} \mathbf{Y} \oplus_{gr} \mathbf{Z} \| \\&= \| \ominus_{gr} (\mathbf{P} \oplus_{gr} (\mathbf{Q} \oplus_{gr} \mathbf{R})) \oplus_{gr} (\mathbf{P} \oplus_{gr} (\mathbf{Q} \oplus_{gr} \mathbf{S})) \| \\&\overset{(3)}{=} \| \operatorname{gyr}[\mathbf{P},\mathbf{Q} \oplus_{gr} \mathbf{R}]( \ominus_{gr} (\mathbf{Q} \oplus_{gr} \mathbf{R}) \oplus_{gr} (\mathbf{Q} \oplus_{gr} \mathbf{S}) ) \| \\&\overset{(4)}{=} \| \ominus_{gr} (\mathbf{Q} \oplus_{gr} \mathbf{R}) \oplus_{gr} (\mathbf{Q} \oplus_{gr} \mathbf{S}) \| \\&\overset{(5)}{=} \| \operatorname{gyr}[\mathbf{Q},\mathbf{R}](\ominus_{gr} \mathbf{R} \oplus_{gr} \mathbf{S}) \| \\&\overset{(6)}{=} \| \ominus_{gr} \mathbf{R} \oplus_{gr} \mathbf{S} \| \\&= d(\mathbf{R},\mathbf{S}).
\end{split}
\end{align}

The derivation of Eq.~(\ref{eq:gyroautomorphisms_preserve_gyrodistance}) follows.

(1) follows from the Left Gyrotranslation Law.

(2) follows from Lemma~\ref{lemma:gyroautomorphisms_preserve_norm_gr}.

(3) follows from the Left Gyrotranslation Law.

(4) follows from Lemma~\ref{lemma:gyroautomorphisms_preserve_norm_gr}.

(5) follows from the Left Gyrotranslation Law.

(6) follows from Lemma~\ref{lemma:gyroautomorphisms_preserve_norm_gr}.

\end{proof}


\section{Proof of Theorem~\ref{theorem:inverse_isometry_gr}}
\label{sec:appendix_inverse_isometry_gr}

\begin{proof}

We first prove the following Lemma: 

\begin{lemma}\label{lemma:inverse_preserve_norm_gr}
Grassmann inverse maps preserve the norm.  
\end{lemma}

\begin{proof}
For $\mathbf{P} \in \operatorname{Gr}_{n,p}$, we have
\begin{align*}
\begin{split}
\| \ominus_{gr} \mathbf{P} \| &= \| \operatorname{Log}^{gr}_{\mathbf{I}_{n,p}}(\ominus_{gr} \mathbf{P}) \|_F \\&= \| -\operatorname{Log}^{gr}_{\mathbf{I}_{n,p}}(\mathbf{P}) \|_F \\&= \| \operatorname{Log}^{gr}_{\mathbf{I}_{n,p}}(\mathbf{P}) \|_F \\&= \| \mathbf{P} \|.
\end{split}
\end{align*}

\end{proof}

For $\mathbf{P}, \mathbf{Q} \in \operatorname{Gr}_{n,p}$, we have
\begin{align}\label{eq:inverse_isometry_gr_1}
\begin{split}
\| \ominus_{gr} \mathbf{P} \oplus_{gr} \mathbf{Q} \| &\overset{(1)}{=} \| \ominus_{gr} (\ominus \mathbf{P} \oplus_{gr} \mathbf{Q}) \| \\&= \| \ominus_{gr} (\ominus_{gr} \mathbf{P} \oplus_{gr} \mathbf{Q}) \oplus_{gr} (\ominus_{gr} \mathbf{P} \oplus_{gr} \mathbf{P}) \| \\&\overset{(2)}{=} \| \operatorname{gyr}[\ominus_{gr} \mathbf{P},\mathbf{Q}](\ominus_{gr} \mathbf{Q} \oplus_{gr} \mathbf{P}) \| \\&\overset{(3)}{=} \| \ominus_{gr} \mathbf{Q} \oplus_{gr} \mathbf{P} \|.        
\end{split}
\end{align}

The derivation of Eq.~(\ref{eq:inverse_isometry_gr_1}) follows.

(1) follows from Lemma~\ref{lemma:inverse_preserve_norm_gr}.

(2) follows from the Left Gyrotranslation Law.

(3) follows from Lemma~\ref{lemma:gyroautomorphisms_preserve_norm_gr}. 

Notice that
\begin{align}\label{eq:inverse_isometry_gr_2}
\begin{split}
\| \mathbf{P} \oplus_{gr} (\ominus_{gr} \mathbf{Q}) \| &\overset{(1)}{=} \| \operatorname{gyr}[\mathbf{P},\ominus_{gr} \mathbf{Q}](\ominus_{gr} \mathbf{Q} \oplus_{gr} \mathbf{P}) \| \\&\overset{(2)}{=} \| \ominus_{gr} \mathbf{Q} \oplus_{gr} \mathbf{P} \|,
\end{split}
\end{align}
where (1) follows from the Gyrocommutative Law, and (2) follows from Lemma~\ref{lemma:gyroautomorphisms_preserve_norm_gr}.

Combining Eqs.~(\ref{eq:inverse_isometry_gr_1}) and~(\ref{eq:inverse_isometry_gr_2}) results in
\begin{align*}
\begin{split}
\| \ominus_{gr} \mathbf{P} \oplus_{gr} \mathbf{Q} \| &= \| \mathbf{P} \oplus_{gr} (\ominus_{gr} \mathbf{Q}) \| \\&= \| \ominus_{gr} (\ominus_{gr} \mathbf{P}) \oplus_{gr} (\ominus \mathbf{Q}) \|,
\end{split}
\end{align*}
which leads to the conclusion of the theorem. 

\end{proof}

\section{Proof of Theorem~\ref{theorem:left_gyrotranslational_gyroisometries_ai}}
\label{sec:appendix_left_gyrotranslational_gyroisometries_ai}

\begin{proof}

We need to prove the following lemmas:

\begin{lemma}\label{lemma:left_gyrotranslation_spd}
Gyrovector spaces $(\operatorname{Sym}^+_n,\oplus_g,\otimes_g)$ verify the Left Gyrotranslation Law, that is,
\begin{equation*}
\ominus_g (\mathbf{P} \oplus_g \mathbf{Q}) \oplus_g (\mathbf{P} \oplus_g \mathbf{R}) = \operatorname{gyr}[\mathbf{P},\mathbf{Q}](\ominus_g \mathbf{Q} \oplus_g \mathbf{R}),
\end{equation*} 
where $\mathbf{P},\mathbf{Q},\mathbf{R} \in \operatorname{Gr}_{n,p}$.  
\end{lemma}

\begin{proof}

Note that gyrovector spaces $(\operatorname{Sym}^+_n,\oplus_g,\otimes_g)$ 
verify the Left Cancellation Law and Left Gyroassociative Law. 
Then the lemma can be proved by using the same arguments as those in the proof of Lemma~\ref{lemma:left_gyrotranslation_gr}.  

\end{proof}

\begin{lemma}\label{lemma:gyroautomorphisms_preserve_norm_spd}
Gyroautomorphisms $\operatorname{gyr}_g[.,.]$ preserve the norm. 
\end{lemma}

\begin{proof}

The lemma can be easily proved by using the expressions of gyroautomorphisms $\operatorname{gyr}_g[.,.]$.  

\end{proof}

We now have the following chain of equations:
\begin{align*}
\begin{split}
\| \ominus_g (\mathbf{A} \oplus_g \mathbf{P}) \oplus_g (\mathbf{A} \oplus_g \mathbf{Q}) \| &\overset{(1)}{=} \| \operatorname{gyr}[\mathbf{A},\mathbf{P}](\ominus_g \mathbf{P} \oplus_g \mathbf{Q}) \| \\&\overset{(2)}{=} \| \ominus_g \mathbf{P} \oplus_g \mathbf{Q} \|, 
\end{split}
\end{align*}
where (1) follows from the Left Gyrotranslation Law (Lemma~\ref{lemma:left_gyrotranslation_spd}), 
and (2) follows from the invariance of the norm under gyroautomorphisms (Lemma~\ref{lemma:gyroautomorphisms_preserve_norm_spd}). 

\end{proof}

\section{Proof of Theorem~\ref{theorem:gyroautomorphisms_isometries_spd}}
\label{sec:appendix_gyroautomorphisms_isometries_spd}

\begin{proof}

Note that gyrovector spaces $(\operatorname{Sym}^+_n,\oplus_g,\otimes_g)$ 
verify the Left Gyroassociative Law, Left Cancellation Law, and Left Gyrotranslation Law (Lemma~\ref{lemma:left_gyrotranslation_spd}) 
and that gyroautomorphisms $\operatorname{gyr}_g[.,.]$ preserve the norm (Lemma~\ref{lemma:gyroautomorphisms_preserve_norm_spd}). 
Then the lemma can be proved by using the same arguments as those 
in the proof of Theorem~\ref{theorem:gyroautomorphisms_gyroisometries_gr}.  
\end{proof}

\section{Proof of Theorem~\ref{theorem:inverse_isometries_spd}}
\label{sec:appendix_inverse_isometries_spd}

\begin{proof}

The following lemma can be proved by using the same arguments as those in the proof of Lemma~\ref{lemma:inverse_preserve_norm_gr}.

\begin{lemma}\label{lemma:inverse_preserve_norm_spd}
SPD inverse maps preserve the norm.  
\end{lemma}

Note that gyrovector spaces $(\operatorname{Sym}_n^+,\oplus_g,\otimes_g)$ verify 
the Gyrocommutative Law and Left Gyrotranslation Law (Lemma~\ref{lemma:left_gyrotranslation_spd}). 
Note also that SPD inverse maps preserve the norm (Lemma~\ref{lemma:inverse_preserve_norm_spd}) 
and that gyroautomorphisms $\operatorname{gyr}_g[.,.]$ preserve the norm (Lemma~\ref{lemma:gyroautomorphisms_preserve_norm_spd}). 
Then the lemma can be proved by using the same arguments as those 
in the proof of Theorem~\ref{theorem:inverse_isometry_gr}.

\end{proof}

\section{Proof of Theorem~\ref{theorem:gyrocosines_law}}
\label{sec:appendix_gyrocosines_law}

{\noindent \bf LC Gyrovector Spaces}

\begin{proof}

We only need to prove the fist identity. For $\mathbf{P},\mathbf{Q},\mathbf{R} \in \operatorname{Sym}^+_n$, 
by the Left Gyrotranslation Law,
\begin{equation*}
\ominus_g(\ominus_g \mathbf{P} \oplus_g \mathbf{Q}) \oplus_g (\ominus_g \mathbf{P} \oplus_g \mathbf{R}) = \operatorname{gyr}[\ominus_g \mathbf{P}, \mathbf{Q}](\ominus_g \mathbf{Q} \oplus_g \mathbf{R}).
\end{equation*}

Since gyroautomorphisms preserve the norm, we have
\begin{equation*}
\| \ominus_g(\ominus_g \mathbf{P} \oplus_g \mathbf{Q}) \oplus_g (\ominus_g \mathbf{P} \oplus_g \mathbf{R}) \| = \| \operatorname{gyr}[\ominus_g \mathbf{P}, \mathbf{Q}](\ominus_g \mathbf{Q} \oplus_g \mathbf{R}) \| = \| \ominus_g \mathbf{Q} \oplus_g \mathbf{R} \|,
\end{equation*}
which results in
\begin{equation}\label{eq:gyrovector_spaces_gyrocosine_proof_main}
\| \ominus_g \widetilde{\mathbf{R}} \oplus_g \widetilde{\mathbf{Q}} \| = \| \widetilde{\mathbf{P}} \|. 
\end{equation}

Hence
\begin{align}\label{eq:lc_gyrovector_spaces_gyrocosine_proof_main_1}
\begin{split}
p^2 = \| \widetilde{\mathbf{P}} \|^2 &= \| \ominus_{lc} \widetilde{\mathbf{R}} \oplus_{lc} \widetilde{\mathbf{Q}} \|^2 \\ &= \langle \lfloor \varphi(\widetilde{\mathbf{Q}}) \rfloor - \lfloor \varphi(\widetilde{\mathbf{R}}) \rfloor + \log( \mathbb{D}(\varphi(\widetilde{\mathbf{R}}))^{-1} \mathbb{D}(\varphi(\widetilde{\mathbf{Q}})) ), \lfloor \varphi(\widetilde{\mathbf{Q}}) \rfloor - \lfloor \varphi(\widetilde{\mathbf{R}}) \rfloor + \log( \mathbb{D}(\varphi(\widetilde{\mathbf{R}}))^{-1} \mathbb{D}(\varphi(\widetilde{\mathbf{Q}})) ) \rangle_F \\ &= \| \lfloor \varphi(\widetilde{\mathbf{Q}}) \rfloor \|^2_F + \| \lfloor \varphi(\widetilde{\mathbf{R}}) \rfloor \|^2_F - 2\langle \lfloor \varphi(\widetilde{\mathbf{Q}}) \rfloor, \lfloor \varphi(\widetilde{\mathbf{R}}) \rfloor \rangle_F + \\& \| \log( \mathbb{D}(\varphi(\widetilde{\mathbf{Q}})) ) \|^2_F + \| \log( \mathbb{D}(\varphi(\widetilde{\mathbf{R}})) ) \|^2_F - 2\langle \log( \mathbb{D}(\varphi(\widetilde{\mathbf{Q}})) ), \log( \mathbb{D}(\varphi(\widetilde{\mathbf{R}})) ) \rangle_F.
\end{split}
\end{align}

Notice that
\begin{align}\label{eq:lc_gyrovector_spaces_gyrocosine_proof_main_2}
\begin{split}
q^2 + r^2 = \| \widetilde{\mathbf{Q}} \|^2 + \| \widetilde{\mathbf{R}} \|^2 &= \| \lfloor \varphi(\widetilde{\mathbf{Q}}) \rfloor \|^2_F + \| \lfloor \varphi(\widetilde{\mathbf{R}}) \rfloor \|^2_F + \| \log( \mathbb{D}(\varphi(\widetilde{\mathbf{Q}})) ) \|^2_F + \| \log( \mathbb{D}(\varphi(\widetilde{\mathbf{R}})) ) \|^2_F, 
\end{split}
\end{align}

and
\begin{align}\label{eq:lc_gyrovector_spaces_gyrocosine_proof_main_3}
\begin{split}
2\langle \widetilde{\mathbf{Q}},\widetilde{\mathbf{R}} \rangle = 2\langle \lfloor \varphi(\widetilde{\mathbf{Q}}) \rfloor, \lfloor \varphi(\widetilde{\mathbf{R}}) \rfloor \rangle_F + 2\langle \log( \mathbb{D}(\varphi(\widetilde{\mathbf{Q}})) ), \log( \mathbb{D}(\varphi(\widetilde{\mathbf{R}})) ) \rangle_F. 
\end{split}
\end{align}

Combining Eqs.~(\ref{eq:lc_gyrovector_spaces_gyrocosine_proof_main_1}),~(\ref{eq:lc_gyrovector_spaces_gyrocosine_proof_main_2}), 
and~(\ref{eq:lc_gyrovector_spaces_gyrocosine_proof_main_3}), we get
\begin{equation*}
p^2 = q^2 + r^2 - 2\langle \widetilde{\mathbf{Q}},\widetilde{\mathbf{R}} \rangle = q^2 + r^2 -2qr \cos \alpha.    
\end{equation*}

\end{proof}

{\noindent \bf LE Gyrovector Spaces}

\begin{proof}

From Eq.~(\ref{eq:gyrovector_spaces_gyrocosine_proof_main}),
\begin{align*}
\begin{split}
p^2 = \| \widetilde{\mathbf{P}} \|^2 &= \| \ominus_{le} \widetilde{\mathbf{R}} \oplus_{le} \widetilde{\mathbf{Q}} \|^2 \\ &= \| \log(\widetilde{\mathbf{Q}}) - \log(\widetilde{\mathbf{R}}) \|^2_F \\ &= \| \log(\widetilde{\mathbf{Q}}) \|^2_F + \| \log(\widetilde{\mathbf{R}}) \|^2_F - 2\langle \log(\widetilde{\mathbf{Q}}), \log(\widetilde{\mathbf{R}}) \rangle_F \\ &= \| \widetilde{\mathbf{Q}} \|^2 + \| \widetilde{\mathbf{R}} \|^2 -2\langle \widetilde{\mathbf{Q}},\widetilde{\mathbf{R}} \rangle \\ &= q^2 + r^2 - 2qr\cos \alpha.    
\end{split}
\end{align*}

\end{proof}

\section{Proof of Theorem~\ref{theorem:distance_to_SPD_hyperplanes_log_euclidean}}
\label{sec:appendix_distance_to_SPD_hyperplanes_log_euclidean}

\begin{proof}

We need to prove the following lemmas: 

\begin{lemma}\label{lem:distance_to_SPD_hyperplanes_lemma1}
Let $\mathbf{P}$ and $\mathbf{Q}$ be two distinct points in a gyrovector space $(\operatorname{Sym}_n^+,\oplus_g,\otimes_g)$. 
Then the geodesic $\delta_{\mathbf{P} \rightarrow \mathbf{Q}}(t), 0 \le t \le 1$ joining $\mathbf{P}$ 
and $\mathbf{Q}$ that passes through $\mathbf{P}$ when $t = 0$ and passes through $\mathbf{Q}$ when $t = 1$ is given by
\begin{equation*}
\delta_{\mathbf{P} \rightarrow \mathbf{Q}}(t) = \mathbf{P} \oplus_g t \otimes_g (\ominus_g \mathbf{P} \oplus_g \mathbf{Q}).
\end{equation*}
\end{lemma}

\begin{proof}
The lemma can be proved using the expressions of the binary operation, inverse operation, and scalar multiplication in 
LE, LC, and AI gyrovector spaces given in~\citet{NguyenECCV22,NguyenNeurIPS22}.  
\end{proof}

\begin{lemma}\label{lem:distance_to_SPD_hyperplanes_lemma2}
Let $\mathbf{P}$ and $\mathbf{Q}$ be two distinct points in a gyrovector space $(\operatorname{Sym}_n^+,\oplus_g,\otimes_g)$. 
Let $\mathbf{S}$ be a point on the geodesic $\delta_{\mathbf{P} \rightarrow \mathbf{Q}}(t), 0 \le t \le 1$  
joining $\mathbf{P}$ and $\mathbf{Q}$, and $\mathbf{R} \notin \delta_{\mathbf{P} \rightarrow \mathbf{Q}}(t)$. 
Then
\begin{equation*}
\angle \mathbf{R} \mathbf{P} \mathbf{S} = \angle \mathbf{R} \mathbf{P} \mathbf{Q}.  
\end{equation*}
\end{lemma}

\begin{proof}
By Lemma~\ref{lem:distance_to_SPD_hyperplanes_lemma1}, $\delta_{\mathbf{P} \rightarrow \mathbf{Q}}(t)$ can be given as
\begin{equation}\label{eq:geodesic_equation}
\delta_{\mathbf{P} \rightarrow \mathbf{Q}}(t) = \mathbf{P} \oplus_g t \otimes_g (\ominus_g \mathbf{P} \oplus_g \mathbf{Q}).
\end{equation}

By the definition of the scalar multiplication,
\begin{equation*}
t \otimes_g \mathbf{P} = \operatorname{Exp}^g_{\mathbf{I}_n}( t \operatorname{Log}^g_{\mathbf{I}_n}(\mathbf{P}) ),
\end{equation*}
where $t \in \mathbb{R}$, which results in
\begin{equation}\label{eq:identity_scalar_multiplication}
\operatorname{Log}_{\mathbf{I}_n}(t \otimes_g \mathbf{P}) = t \operatorname{Log}^g_{\mathbf{I}_n}(\mathbf{P}).  
\end{equation}

From Eq.~(\ref{eq:identity_scalar_multiplication}), we get
\begin{equation*}
\| \operatorname{Log}_{\mathbf{I}_n}(t \otimes_g \mathbf{P}) \|_F = \| t \operatorname{Log}^g_{\mathbf{I}_n}(\mathbf{P}) \|_F = t \| \operatorname{Log}^g_{\mathbf{I}_n}(\mathbf{P}) \|_F,  
\end{equation*}
which leads to
\begin{equation}\label{eq:identity_scalar_multiplication_norm}
\| t \otimes_g \mathbf{P} \| = t \| \mathbf{P} \|.    
\end{equation}

By the definition of the SPD inner product and Eq.~(\ref{eq:identity_scalar_multiplication}), we also have
\begin{align}\label{eq:identity_inner_product}
\begin{split}
\langle \mathbf{P},t \otimes_g \mathbf{Q} \rangle &= \langle \operatorname{Log}_{\mathbf{I}_n}(\mathbf{P}), \operatorname{Log}_{\mathbf{I}_n}(t \otimes_g \mathbf{Q}) \rangle_F \\ &= \langle \operatorname{Log}_{\mathbf{I}_n}(\mathbf{P}), t \operatorname{Log}_{\mathbf{I}_n}(\mathbf{Q}) \rangle_F \\ &= t \langle \operatorname{Log}_{\mathbf{I}_n}(\mathbf{P}), \operatorname{Log}_{\mathbf{I}_n}(\mathbf{Q})  \rangle_F \\ &= t \langle \mathbf{P},\mathbf{Q} \rangle.  
\end{split}
\end{align}

Therefore
\begin{align}\label{eq:identity_angle_geodesic}
\begin{split}
\cos(\angle \mathbf{R} \mathbf{P} \mathbf{S}) &\overset{(1)}{=} \frac{\langle \ominus_g \mathbf{P} \oplus_g \mathbf{R},\ominus_g \mathbf{P} \oplus_g \mathbf{S} \rangle}{\| \ominus_g \mathbf{P} \oplus_g \mathbf{R} \|. \| \ominus_g \mathbf{P} \oplus_g \mathbf{S} \|} \\ &\overset{(2)}{=} \frac{\langle \ominus_g \mathbf{P} \oplus_g \mathbf{R},\ominus_g \mathbf{P} \oplus_g (\mathbf{P} \oplus_g t \otimes_g (\ominus_g \mathbf{P} \oplus_g \mathbf{Q})) \rangle}{\| \ominus_g \mathbf{P} \oplus_g \mathbf{R} \|. \| \ominus_g \mathbf{P} \oplus_g (\mathbf{P} \oplus_g t \otimes_g (\ominus_g \mathbf{P} \oplus_g \mathbf{Q})) \|} \\ &\overset{(3)}{=} \frac{\langle \ominus_g \mathbf{P} \oplus_g \mathbf{R},t \otimes_g (\ominus_g \mathbf{P} \oplus_g \mathbf{Q}) \rangle}{\| \ominus_g \mathbf{P} \oplus_g \mathbf{R} \|. \| t \otimes_g (\ominus_g \mathbf{P} \oplus_g \mathbf{Q}) \|} \\ &\overset{(4)}{=} \frac{t \langle \ominus_g \mathbf{P} \oplus_g \mathbf{R},\ominus_g \mathbf{P} \oplus_g \mathbf{Q} \rangle}{t \| \ominus_g \mathbf{P} \oplus_g \mathbf{R} \|. \| \ominus_g \mathbf{P} \oplus_g \mathbf{Q} \|} \\ &\overset{(5)}{=} \cos(\angle \mathbf{R} \mathbf{P} \mathbf{Q}).  
\end{split}
\end{align}

The derivation of Eq.~(\ref{eq:identity_angle_geodesic}) follows.

(1) follows from the definition of SPD gyroangles. 

(2) follows from Eq.~(\ref{eq:geodesic_equation}).

(3) follows from the Left Cancellation Law.  

(4) follows from Eqs.~(\ref{eq:identity_scalar_multiplication_norm}) and~(\ref{eq:identity_inner_product}).  

(5) follows from the definition of SPD gyroangles. 

This leads to the conclusion of the lemma.  

\end{proof}

\begin{lemma}\label{lem:distance_to_SPD_hyperplanes_lemma3}
Let $\mathbf{P}$ and $\mathbf{Q}$ be two distinct points in a gyrovector space $(\operatorname{Sym}_n^+,\oplus_g,\otimes_g)$, 
$g \in \{ le,lc \}$. 
Denote by $\delta_{\mathbf{P} \rightarrow \mathbf{Q}}(t), 0 \le t \le 1$ the geodesic joining $\mathbf{P}$ and $\mathbf{Q}$, 
$\mathbf{P}'= \delta_{\mathbf{P} \rightarrow \mathbf{Q}}(-\infty)$, $\mathbf{Q}' = \delta_{\mathbf{P} \rightarrow \mathbf{Q}}(\infty)$,   
and $\mathbf{R} \in (\operatorname{Sym}_n^+,\oplus_g,\otimes_g)$ 
such that $\mathbf{R} \notin \delta_{\mathbf{P} \rightarrow \mathbf{Q}}(t),t \in \mathbb{R}$.   
Then there exists a unique $\mathbf{S} \in \delta_{\mathbf{P} \rightarrow \mathbf{Q}}(t),t \in \mathbb{R}$ 
such that $\angle \mathbf{R} \mathbf{S} \mathbf{Q}' = \frac{\pi}{2}$.  
\end{lemma}

{\noindent \bf LE Gyrovector Spaces}

\begin{proof}


First, it is easy to see that any points $\mathbf{P}',\mathbf{Q}'$ such that 
$\mathbf{Q} \in \delta_{\mathbf{P} \rightarrow \mathbf{Q}'}(t)$ and
$\mathbf{P} \in \delta_{\mathbf{P}' \rightarrow \mathbf{Q}}(t),0 \le t \le 1$ can be written as 
$\delta_{\mathbf{P} \rightarrow \mathbf{Q}}(t)$ in Eq.~(\ref{eq:geodesic_equation}) where $t \in \mathbb{R}$.   
We have
\begin{align*}
\begin{split}
\cos( \angle \mathbf{R}\mathbf{S}\mathbf{Q}' ) &\overset{(1)}{=} \frac{\langle \ominus_{le} \mathbf{S} \oplus_{le} \mathbf{R},\ominus_{le} \mathbf{S} \oplus_{le} \mathbf{Q}' \rangle}{\| \ominus_{le} \mathbf{S} \oplus_{le} \mathbf{R} \|. \| \ominus_{le} \mathbf{S} \oplus_{le} \mathbf{Q}' \|} \\ &= \frac{\langle \log(\mathbf{R}) - \log(\mathbf{S}),\log(\mathbf{Q}') - \log(\mathbf{S}) \rangle_F}{\| \log(\mathbf{R}) - \log(\mathbf{S}) \|_F.\| \log(\mathbf{Q}') - \log(\mathbf{S}) \|_F} \\ &\overset{(2)}{=} \frac{\langle \log(\mathbf{R}) - (1-t)\log(\mathbf{P}') - t\log(\mathbf{Q}'),\log(\mathbf{Q}') - (1-t)\log(\mathbf{P}') - t\log(\mathbf{Q}') \rangle_F}{\| \log(\mathbf{R}) - (1-t)\log(\mathbf{P}') - t\log(\mathbf{Q}') \|_F.\| \log(\mathbf{Q}') - (1-t)\log(\mathbf{P}') - t\log(\mathbf{Q}') \|_F}  \\ &= \frac{\langle \log(\mathbf{R}) - \log(\mathbf{P}') - t(\log(\mathbf{Q}')-\log(\mathbf{P}')),\log(\mathbf{Q}') - \log(\mathbf{P}') \rangle_F}{\| \log(\mathbf{R}) - \log(\mathbf{P}') - t(\log(\mathbf{Q}')-\log(\mathbf{P}')) \|_F.\| \log(\mathbf{Q}') - \log(\mathbf{P}') \|_F}, 
\end{split}
\end{align*}
where (1) follows from the definition of SPD gyroangles, and (2) follows from the equation of geodesics in LE gyrovector spaces.  



It can be seen that there exists $t \in \mathbb{R}$ 
such that $\langle \log(\mathbf{R}) - \log(\mathbf{P}') - t(\log(\mathbf{Q}')-\log(\mathbf{P}')),\log(\mathbf{Q}') - \log(\mathbf{P}') \rangle_F = 0$ and thus $\cos( \angle \mathbf{R}\mathbf{S}\mathbf{Q}' ) = 0$, 
or equivalently, $\angle \mathbf{R}\mathbf{S}\mathbf{Q}' = \frac{\pi}{2}$.  
Now, assuming that there exists two distinct points 
$\mathbf{S},\mathbf{S}' \in \delta_{\mathbf{P} \rightarrow \mathbf{Q}}(t),t \in \mathbb{R}$ 
such that $\angle \mathbf{R} \mathbf{S} \mathbf{Q}' = \angle \mathbf{R} \mathbf{S}' \mathbf{Q}' = \frac{\pi}{2}$.
Let $p = \| \ominus_{le} \mathbf{S}' \oplus_{le} \mathbf{R} \|$, 
$q = \| \ominus_{le} \mathbf{S} \oplus_{le} \mathbf{R} \|$, 
and $r = \| \ominus_{le} \mathbf{S} \oplus_{le} \mathbf{S}' \|$.  
By the Law of SPD gyrocosines (see Theorem~\ref{theorem:gyrocosines_law}), 
\begin{equation*}
p^2 = q^2 + r^2,
\end{equation*}
and
\begin{equation*}
q^2 = p^2 + r^2,
\end{equation*}
which leads to contradiction as $r > 0$. 
We conclude that there exists a unique $\mathbf{S}$ that verifies the property in Lemma~\ref{lem:distance_to_SPD_hyperplanes_lemma3}.  


\end{proof}

{\noindent \bf LC Gyrovector Spaces}

Note that the geodesic $\delta_{\mathbf{P} \rightarrow \mathbf{Q}}(t), 0 \le t \le 1$ joining $\mathbf{P}$ and $\mathbf{Q}$ 
can be written as
\begin{equation*}
\delta_{\mathbf{P} \rightarrow \mathbf{Q}}(t) = \big( \varphi(\mathbf{P}) \oplus_{lc} t \otimes_{lc} (\ominus_{lc} \varphi(\mathbf{P}) \oplus_{lc} \varphi(\mathbf{Q})) \big) \big( \varphi(\mathbf{P}) \oplus_{lc} t \otimes_{lc} (\ominus_{lc} \varphi(\mathbf{P}) \oplus_{lc} \varphi(\mathbf{Q})) \big)^T.  
\end{equation*}

Some manipulations lead to
\begin{equation*}
\cos( \angle \mathbf{R}\mathbf{S}\mathbf{Q}' ) = \frac{ \langle \mathbf{A},\mathbf{B} \rangle_F }{ \| \mathbf{A} \|_F . \| \mathbf{B} \|_F },
\end{equation*}
where
\begin{equation*}
\mathbf{A} = (\lfloor \varphi(\mathbf{R}) \rfloor - \lfloor \varphi(\mathbf{P}) \rfloor) + \log(\mathbb{D}(\varphi(\mathbf{R}))) - \log(\mathbb{D}(\varphi(\mathbf{P}))) +  t(\lfloor \varphi(\mathbf{P}) \rfloor - \lfloor \varphi(\mathbf{Q}') \rfloor) +t\big( \log(\mathbb{D}(\varphi(\mathbf{P}))) - \log(\mathbb{D}(\varphi(\mathbf{Q}'))) \big),
\end{equation*}
\begin{equation*}
\mathbf{B} = \lfloor \varphi(\mathbf{P}) \rfloor - \lfloor \varphi(\mathbf{Q}') \rfloor + \log(\mathbb{D}(\varphi(\mathbf{P}))) - \log(\mathbb{D}(\varphi(\mathbf{Q}'))).  
\end{equation*}

It can be seen that there exists $t \in \mathbb{R}$ such that $\langle \mathbf{A},\mathbf{B} \rangle_F = 0$ 
and thus $\cos( \angle \mathbf{R}\mathbf{S}\mathbf{Q}' ) = 0$, or equivalently, 
$\angle \mathbf{R}\mathbf{S}\mathbf{Q}' = \frac{\pi}{2}$.  
The uniqueness of $\mathbf{S}$ can be proved by using the same arguments as for LE gyrovector spaces.  

\begin{lemma}\label{lem:distance_to_SPD_hyperplanes_lemma4}
Let $\mathcal{H}_{\mathbf{W},\mathbf{P}}$ be a SPD hypergyroplane 
in a gyrovector space $(\operatorname{Sym}_n^+,\oplus_g,\oplus_g)$, 
and $\mathbf{Q} \in \mathcal{H}_{\mathbf{W},\mathbf{P}} \setminus \{ \mathbf{P} \}$.   
Then all points on the geodesic $\delta_{\mathbf{P} \rightarrow \mathbf{Q}}(t)$ belong to $\mathcal{H}_{\mathbf{W},\mathbf{P}}$. 
\end{lemma}

\begin{proof}

We have
\begin{align}\label{eq:all_geodesic_on_hyperplane}
\begin{split}
\operatorname{Log}_{\mathbf{P}}(\delta_{\mathbf{P} \rightarrow \mathbf{Q}}(t)) &\overset{(1)}{=} \operatorname{Log}_{\mathbf{P}}(\mathbf{P} \oplus_g t \otimes_g (\ominus_g \mathbf{P} \oplus_g \mathbf{Q})) \\ &\overset{(2)}{=} \operatorname{Log}_{\mathbf{P}} \big( \operatorname{Exp}_{\mathbf{P}}(\mathcal{T}_{\mathbf{I}_n \rightarrow \mathbf{P}}(\operatorname{Log}_{\mathbf{I}_n}(t \otimes_g (\ominus_g \mathbf{P} \oplus_g \mathbf{Q})))) \big) \\ &= \mathcal{T}_{\mathbf{I}_n \rightarrow \mathbf{P}}(\operatorname{Log}_{\mathbf{I}_n}(t \otimes_g (\ominus_g \mathbf{P} \oplus_g \mathbf{Q}))) \\ &\overset{(3)}{=} \mathcal{T}_{\mathbf{I}_n \rightarrow \mathbf{P}}(t\operatorname{Log}_{\mathbf{I}_n}(\ominus_g \mathbf{P} \oplus_g \mathbf{Q})) \\ &= t \mathcal{T}_{\mathbf{I}_n \rightarrow \mathbf{P}}(\operatorname{Log}_{\mathbf{I}_n}(\ominus_g \mathbf{P} \oplus_g \mathbf{Q})) \\ &= t \operatorname{Log}_{\mathbf{P}} \Big( \operatorname{Exp}_{\mathbf{P}} \big( \mathcal{T}_{\mathbf{I}_n \rightarrow \mathbf{P}}(\operatorname{Log}_{\mathbf{I}_n}(\ominus_g \mathbf{P} \oplus_g \mathbf{Q})) \big) \Big) \\ &\overset{(4)}{=} t \operatorname{Log}_{\mathbf{P}}( \mathbf{P} \oplus_g (\ominus_g \mathbf{P} \oplus_g \mathbf{Q}) ) \\ &\overset{(5)}{=} t \operatorname{Log}_{\mathbf{P}}(\mathbf{Q}). 
\end{split}
\end{align}

The derivation of Eq.~(\ref{eq:all_geodesic_on_hyperplane}) follows.

(1) follows from Eq.~(\ref{eq:geodesic_equation}).

(2) follows from the definition of the binary operation in Eq.~(\ref{eq:matrix_matrix_addition}).

(3) follows from Eq.~(\ref{eq:identity_scalar_multiplication}).

(4) follows from the definition of the binary operation in Eq.~(\ref{eq:matrix_matrix_addition}).

(5) follows from the Left Cancellation Law.  

Therefore
\begin{equation*}
\langle \operatorname{Log}_{\mathbf{P}}(\delta_{\mathbf{P} \rightarrow \mathbf{Q}}(t)), \mathbf{W} \rangle_{\mathbf{P}} = \langle t \operatorname{Log}_{\mathbf{P}}(\mathbf{Q}), \mathbf{W} \rangle_{\mathbf{P}},
\end{equation*}
which results in 
$\langle \operatorname{Log}_{\mathbf{P}}(\delta_{\mathbf{P} \rightarrow \mathbf{Q}}(t)), \mathbf{W} \rangle_{\mathbf{P}} = 0$.  
This shows that all points on the geodesic $\delta_{\mathbf{P} \rightarrow \mathbf{Q}}(t)$ belong to SPD hypergyroplane $\mathcal{H}_{\mathbf{W},\mathbf{P}}$.  
\end{proof}


Let $\mathcal{H}_{\mathbf{W},\mathbf{P}}$ be a SPD hypergyroplane 
in a gyrovector space $(\operatorname{Sym}_n^+,\oplus_{le},\otimes_{le})$,
$\mathbf{X} \notin \mathcal{H}_{\mathbf{W},\mathbf{P}}$,  
and $\mathbf{Q}^* \in \mathcal{H}_{\mathbf{W},\mathbf{P}}$ such that
\begin{equation}\label{eq:pseudo_distance_coincides_distance_1}
d(\mathbf{X},\mathbf{Q}^*) = \min_{\mathbf{Q} \in \mathcal{H}_{\mathbf{W},\mathbf{P}}} d(\mathbf{X},\mathbf{Q}) = d(\mathbf{X},\mathcal{H}_{\mathbf{W},\mathbf{P}}).   
\end{equation}

We prove the first part of the theorem, i.e., 
\begin{equation*}
\bar{d}(\mathbf{X},\mathcal{H}_{\mathbf{W},\mathbf{P}}) = d(\mathbf{X},\mathcal{H}_{\mathbf{W},\mathbf{P}}).  
\end{equation*}

We consider two cases:

{\noindent \em Case 1: $\mathbf{Q}^* \neq \mathbf{P}$}.

If $\angle \mathbf{X}\mathbf{Q}^*\mathbf{P} \neq \frac{\pi}{2}$, then by Lemma~\ref{lem:distance_to_SPD_hyperplanes_lemma3},  
there exists a unique $\mathbf{Q}^{**} \in \delta_{\mathbf{P} \rightarrow \mathbf{Q}^*}(t) ,t \in \mathbf{R},\mathbf{Q}^{**} \neq \mathbf{Q}^*$ such that $\angle \mathbf{X}\mathbf{Q}^{**}\mathbf{Q}' = \frac{\pi}{2}$ where $\mathbf{Q}' = \delta_{\mathbf{P} \rightarrow \mathbf{Q}^*}(\infty)$. 
By Lemma~\ref{lem:distance_to_SPD_hyperplanes_lemma2}, $\angle \mathbf{X}\mathbf{Q}^{**}\mathbf{Q}^* = \frac{\pi}{2}$.  
By the Law of SPD gyrosines,
\begin{equation*}
d(\mathbf{X},\mathbf{Q}^{**}) = \sin(\angle \mathbf{X} \mathbf{Q}^* \mathbf{Q}^{**}) d(\mathbf{X},\mathbf{Q}^*),
\end{equation*}
which means that $d(\mathbf{X},\mathbf{Q}^{**}) < d(\mathbf{X},\mathbf{Q}^*)$. 
By Lemma~\ref{lem:distance_to_SPD_hyperplanes_lemma4}, $\mathbf{Q}^{**} \in \mathcal{H}_{\mathbf{W},\mathbf{P}}$. 
This leads to a contradiction because of the definition of $\mathbf{Q}^*$.  
Therefore, we must have $\angle \mathbf{X}\mathbf{Q}^*\mathbf{P} = \frac{\pi}{2}$.    
Now, by the Law of SPD gyrosines,
\begin{equation*}
d(\mathbf{X},\mathbf{Q}^*) = \sin(\angle \mathbf{X} \mathbf{P} \mathbf{Q}^*) d(\mathbf{X},\mathbf{P}).
\end{equation*}

We thus deduce that
\begin{equation*}
\sin(\angle \mathbf{X} \mathbf{P} \mathbf{Q}^*) = \min_{\mathbf{Q} \in \mathcal{H}_{\mathbf{W},\mathbf{P}} \setminus \{ \mathbf{P} \}} \sin(\angle \mathbf{X} \mathbf{P} \mathbf{Q}),
\end{equation*}
or equivalently,
\begin{equation*}
\cos(\angle \mathbf{X} \mathbf{P} \mathbf{Q}^*) = \max_{\mathbf{Q} \in \mathcal{H}_{\mathbf{W},\mathbf{P}} \setminus \{ \mathbf{P} \}} \cos(\angle \mathbf{X} \mathbf{P} \mathbf{Q}),
\end{equation*}

Therefore
\begin{equation}\label{eq:pseudo_distance_coincides_distance_2}
d(\mathbf{X},\mathbf{Q}^*) = \bar{d}(\mathbf{X},\mathcal{H}_{\mathbf{W},\mathbf{P}}).  
\end{equation}

Combining Eqs.~(\ref{eq:pseudo_distance_coincides_distance_1}) and~(\ref{eq:pseudo_distance_coincides_distance_2}) 
leads to
\begin{equation*}
\bar{d}(\mathbf{X},\mathcal{H}_{\mathbf{W},\mathbf{P}}) = d(\mathbf{X},\mathcal{H}_{\mathbf{W},\mathbf{P}}).  
\end{equation*}



{\noindent \em Case 2: $\mathbf{Q}^* = \mathbf{P}$}.  

For any $\mathbf{Q} \in \mathcal{H}_{\mathbf{W},\mathbf{P}} \setminus \{ \mathbf{P} \}$, 
by the same arguments as above, we must have $\angle \mathbf{X} \mathbf{P} \mathbf{Q} = \frac{\pi}{2}$ and therefore  
\begin{equation*}
\bar{d}(\mathbf{X},\mathcal{H}_{\mathbf{W},\mathbf{P}}) = \sin(\angle \mathbf{X} \mathbf{P} \mathbf{Q}) d(\mathbf{X},\mathbf{P}) = d(\mathbf{X},\mathbf{P}) = d(\mathbf{X},\mathbf{Q}^*) = d(\mathbf{X},\mathcal{H}_{\mathbf{W},\mathbf{P}}), 
\end{equation*}
which concludes the first part of the theorem. 

We now prove the second part of the theorem, i.e., 
\begin{equation*}
d(\mathbf{X},\mathcal{H}_{\mathbf{W},\mathbf{P}}) = \frac{| \langle \log(\mathbf{X}) - \log(\mathbf{P}),D\log_{\mathbf{P}}(\mathbf{W}) \rangle_F |}{\| D\log_{\mathbf{P}}(\mathbf{W}) \|_F}.
\end{equation*}

Again, we consider two cases:

{\noindent \em Case 1: $\mathbf{Q}^* \neq \mathbf{P}$}.

For $\mathbf{Q} \in \operatorname{Sym}_n^+$, note that
\begin{equation*}
\mathbf{Q} = \exp_{\mathbf{P}}(\operatorname{Log}^{le}_{\mathbf{P}}(\mathbf{Q})) \overset{(1)}{=} \exp(\log(\mathbf{P}) + D\log_{\mathbf{P}}(\operatorname{Log}^{le}_{\mathbf{P}}(\mathbf{Q}))),
\end{equation*}
where (1) follows from the expression of the exponential map associated with Log-Euclidean metrics.

Hence
\begin{equation}\label{eq:distance_spd_hyperplane_key_identity_le}
D\log_{\mathbf{P}}(\operatorname{Log}^{le}_{\mathbf{P}}(\mathbf{Q})) = \log(\mathbf{Q}) - \log(\mathbf{P}).  
\end{equation}

We then have
\begin{align*}
\begin{split}
\langle \operatorname{Log}^{le}_{\mathbf{P}}(\mathbf{Q}),\mathbf{W} \rangle_{\mathbf{P}} &\overset{(1)}{=} \langle D\log_{\mathbf{P}}(\operatorname{Log}^{le}_{\mathbf{P}}(\mathbf{Q})), D\log_{\mathbf{P}}(\mathbf{W}) \rangle_F \\ &\overset{(2)}{=} \langle \log(\mathbf{Q}) - \log(\mathbf{P}), D\log_{\mathbf{P}}(\mathbf{W}) \rangle_F, 
\end{split}
\end{align*}
where (1) follows from the fact that LE metrics are bi-invariant, 
and (2) follows from Eq.~(\ref{eq:distance_spd_hyperplane_key_identity_le}).  
Thus, for $\mathbf{Q} \in \mathcal{H}_{\mathbf{W},\mathbf{P}}$, we have
\begin{equation}\label{eq:distance_spd_hyperplane_condition_le}
\langle \log(\mathbf{Q}) - \log(\mathbf{P}), D\log_{\mathbf{P}}(\mathbf{W}) \rangle_F = 0.
\end{equation}

We need to find $\mathbf{Q}^* \in \mathcal{H}_{\mathbf{W},\mathbf{P}} \setminus \{ \mathbf{P} \}$ such that
\begin{align*}
\begin{split}
\mathbf{Q}^* &= \argmax_{\mathbf{Q} \in \mathcal{H}_{\mathbf{W},\mathbf{P}} \setminus \{ \mathbf{P} \}} \frac{ \langle \ominus_{le} \mathbf{P} \oplus_{le} \mathbf{Q}, \ominus_{le} \mathbf{P} \oplus_{le} \mathbf{X} \rangle }{ \| \ominus_{le} \mathbf{P} \oplus_{le} \mathbf{Q} \|. \| \ominus_{le} \mathbf{P} \oplus_{le} \mathbf{X} \| } \\ &\overset{(1)}{=} \argmax_{\mathbf{Q} \in \mathcal{H}_{\mathbf{W},\mathbf{P}} \setminus \{ \mathbf{P} \}} \frac{ \langle \operatorname{Log}^{le}_{\mathbf{I}_n}(\ominus_{le} \mathbf{P} \oplus_{le} \mathbf{Q}), \operatorname{Log}^{le}_{\mathbf{I}_n}(\ominus_{le} \mathbf{P} \oplus_{le} \mathbf{X}) \rangle_F }{ \| \operatorname{Log}^{le}_{\mathbf{I}_n}(\ominus_{le} \mathbf{P} \oplus_{le} \mathbf{Q}) \|_F . \| \operatorname{Log}^{le}_{\mathbf{I}_n}(\ominus_{le} \mathbf{P} \oplus_{le} \mathbf{X}) \|_F } \\ &\overset{(2)}{=} \argmax_{\mathbf{Q} \in \mathcal{H}_{\mathbf{W},\mathbf{P}} \setminus \{ \mathbf{P} \}} \frac{ \langle \log(\mathbf{Q}) - \log(\mathbf{P}), \log(\mathbf{X}) - \log(\mathbf{P}) \rangle_F }{ \| \log(\mathbf{Q}) - \log(\mathbf{P}) \|_F . \| \log(\mathbf{X}) - \log(\mathbf{P}) \|_F }, 
\end{split}
\end{align*}
where (1) follows from the definition of the SPD inner product, 
and (2) follows from the expressions of the binary operation $\oplus_{le}$ and inverse operation $\ominus_{le}$.  

Our problem returns to the one of finding the minimum angle between the vector $\log(\mathbf{X}) - \log(\mathbf{P})$ 
and the Euclidean hyperplane described by Eq.~(\ref{eq:distance_spd_hyperplane_condition_le}).  
The SPD gyrodistance $d(\mathbf{X},\mathcal{H}_{\mathbf{W},\mathbf{P}})$ thus can be obtained as
\begin{align*}
\begin{split}
d(\mathbf{X},\mathcal{H}_{\mathbf{W},\mathbf{P}}) &= \frac{| \langle \log(\mathbf{X}) - \log(\mathbf{P}), \frac{ D\log_{\mathbf{P}}(\mathbf{W}) }{\| D\log_{\mathbf{P}}(\mathbf{W}) \|_F} \rangle_F |}{\| \log(\mathbf{X}) - \log(\mathbf{P}) \|_F} . \| \log(\mathbf{X}) - \log(\mathbf{P}) \|_F \\ &= \frac{| \langle \log(\mathbf{X}) - \log(\mathbf{P}),D\log_{\mathbf{P}}(\mathbf{W}) \rangle_F |}{\| D\log_{\mathbf{P}}(\mathbf{W}) \|_F}.  
\end{split}
\end{align*}

{\noindent \em Case 2: $\mathbf{Q}^* = \mathbf{P}$}. 

This case is trivial.   


\end{proof}

\section{Proof of Theorem~\ref{theorem:distance_to_SPD_hyperplanes_log_cholesky}}
\label{sec:appendix_distance_to_SPD_hyperplanes_log_cholesky}

\begin{proof}

Let $\mathcal{H}_{\mathbf{W},\mathbf{P}}$ be a SPD hypergyroplane 
in a gyrovector space $(\operatorname{Sym}_n^+,\oplus_{lc},\otimes_{lc})$,
$\mathbf{X} \notin \mathcal{H}_{\mathbf{W},\mathbf{P}}$,  
and $\mathbf{Q}^* \in \mathcal{H}_{\mathbf{W},\mathbf{P}}$ such that
\begin{equation*}
d(\mathbf{X},\mathbf{Q}^*) = \min_{\mathbf{Q} \in \mathcal{H}_{\mathbf{W},\mathbf{P}}} d(\mathbf{X},\mathbf{Q}) = d(\mathbf{X},\mathcal{H}_{\mathbf{W},\mathbf{P}}).   
\end{equation*}

The first part of the theorem can be proved using the same arguments as those in 
Appendix~\ref{sec:appendix_distance_to_SPD_hyperplanes_log_euclidean}.  
For the second part, we will only consider the case where $\mathbf{Q}^* \neq \mathbf{P}$, 
as the case where $\mathbf{Q}^* = \mathbf{P}$ is trivial (see Appendix~\ref{sec:appendix_distance_to_SPD_hyperplanes_log_euclidean}).  
We have
\begin{equation*}
\mathbf{Q}^* = \argmax_{\mathbf{Q} \in \mathcal{H}_{\mathbf{W},\mathbf{P}} \setminus \{ \mathbf{P} \}} \frac{ \langle \ominus_{lc} \mathbf{P} \oplus_{lc} \mathbf{Q}, \ominus_{lc} \mathbf{P} \oplus_{lc} \mathbf{X} \rangle }{ \| \ominus_{lc} \mathbf{P} \oplus_{lc} \mathbf{Q} \|. \| \ominus_{lc} \mathbf{P} \oplus_{lc} \mathbf{X} \| }. 
\end{equation*}

Let $\widetilde{\mathbf{Q}} = \ominus_{lc} \mathbf{P} \oplus_{lc} \mathbf{Q}$, 
$\widetilde{\mathbf{X}} = \ominus_{lc} \mathbf{P} \oplus_{lc} \mathbf{X}$. Then
\begin{equation*}
\widetilde{\mathbf{Q}} = \big( -\lfloor \varphi(\mathbf{P}) \rfloor + \lfloor \varphi(\mathbf{Q}) \rfloor + \mathbb{D}(\varphi(\mathbf{P}))^{-1} \mathbb{D}(\varphi(\mathbf{Q})) \big) \big( -\lfloor \varphi(\mathbf{P}) \rfloor + \lfloor \varphi(\mathbf{Q}) \rfloor + \mathbb{D}(\varphi(\mathbf{P}))^{-1} \mathbb{D}(\varphi(\mathbf{Q})) \big)^T,  
\end{equation*}
\begin{equation*}
\widetilde{\mathbf{X}} = \big( -\lfloor \varphi(\mathbf{P}) \rfloor + \lfloor \varphi(\mathbf{X}) \rfloor + \mathbb{D}(\varphi(\mathbf{P}))^{-1} \mathbb{D}(\varphi(\mathbf{X})) \big) \big( -\lfloor \varphi(\mathbf{P}) \rfloor + \lfloor \varphi(\mathbf{X}) \rfloor + \mathbb{D}(\varphi(\mathbf{P}))^{-1} \mathbb{D}(\varphi(\mathbf{X})) \big)^T.  
\end{equation*}

Using the definition of the SPD inner product, we get
\begin{equation*}
\langle \widetilde{\mathbf{Q}}, \widetilde{\mathbf{X}} \rangle = \langle -\lfloor \varphi(\mathbf{P}) \rfloor + \lfloor \varphi(\mathbf{Q}) \rfloor + \log( \mathbb{D}(\varphi(\mathbf{P}))^{-1} \mathbb{D}(\varphi(\mathbf{Q})) ), -\lfloor \varphi(\mathbf{P}) \rfloor + \lfloor \varphi(\mathbf{X}) \rfloor + \log( \mathbb{D}(\varphi(\mathbf{P}))^{-1} \mathbb{D}(\varphi(\mathbf{X})) ) \rangle_F.   
\end{equation*}

Therefore
\begin{equation*}
\mathbf{Q}^* = \argmax_{\mathbf{Q} \in \mathcal{H}_{\mathbf{W},\mathbf{P}} \setminus \{ \mathbf{P} \}} \frac{ \langle \mathbf{Z}_1,\mathbf{Z}_2 \rangle_F }{ \| \mathbf{Z}_1 \|_F . \| \mathbf{Z}_2 \|_F },
\end{equation*}
where $\mathbf{Z}_1 = -\lfloor \varphi(\mathbf{P}) \rfloor + \lfloor \varphi(\mathbf{Q}) \rfloor + \log( \mathbb{D}(\varphi(\mathbf{P}))^{-1} \mathbb{D}(\varphi(\mathbf{Q})) )$ and $\mathbf{Z}_2 = -\lfloor \varphi(\mathbf{P}) \rfloor + \lfloor \varphi(\mathbf{X}) \rfloor + \log( \mathbb{D}(\varphi(\mathbf{P}))^{-1} \mathbb{D}(\varphi(\mathbf{X})) )$.

By the definition of Log-Cholesky metrics,
\begin{align*}
\begin{split}
\langle \operatorname{Log}^{lc}_{\mathbf{P}}(\mathbf{Q}), \mathbf{W} \rangle_{\mathbf{P}} &= \langle \varphi(\mathbf{P}) \big( \varphi(\mathbf{P})^{-1} \operatorname{Log}^{lc}_{\mathbf{P}}(\mathbf{Q}) (\varphi(\mathbf{P})^{-1})^T \big)_{\frac{1}{2}}, \varphi(\mathbf{P}) \big( \varphi(\mathbf{P})^{-1} \mathbf{W} (\varphi(\mathbf{P})^{-1})^T \big)_{\frac{1}{2}} \rangle_{\varphi(\mathbf{P})}. 
\end{split}
\end{align*}

Note that
\begin{align*}
\begin{split}
\varphi(\mathbf{P})^{-1} \operatorname{Log}^{lc}_{\mathbf{P}}(\mathbf{Q}) (\varphi(\mathbf{P})^{-1})^T &= \varphi(\mathbf{P})^{-1} \Big( \varphi(\mathbf{P}) \big( \widetilde{\operatorname{Log}}_{\varphi(\mathbf{P})}(\varphi(\mathbf{Q})) \big)^T + \widetilde{\operatorname{Log}}_{\varphi(\mathbf{P})}(\varphi(\mathbf{Q})) \varphi(\mathbf{P})^T \Big) (\varphi(\mathbf{P})^{-1})^T \\ &= \big( \widetilde{\operatorname{Log}}_{\varphi(\mathbf{P})}(\varphi(\mathbf{Q})) \big)^T (\varphi(\mathbf{P})^{-1})^T + \varphi(\mathbf{P})^{-1} \widetilde{\operatorname{Log}}_{\varphi(\mathbf{P})}(\varphi(\mathbf{Q})),
\end{split}
\end{align*}
where $\widetilde{\operatorname{Log}}_{\mathbf{L}}(\mathbf{K}) = \lfloor \mathbf{K} \rfloor - \lfloor \mathbf{L} \rfloor + \mathbb{D}(\mathbf{L})\log(\mathbb{D}(\mathbf{L})^{-1}\mathbb{D}(\mathbf{K}))$ denotes the exponential map 
on the space of lower triangular matrices with positive diagonal entries~\cite{Lin_2019}.  

Hence
\begin{align*}
\begin{split}
\langle \operatorname{Log}^{lc}_{\mathbf{P}}(\mathbf{Q}), \mathbf{W} \rangle_{\mathbf{P}} &= \langle \varphi(\mathbf{P}) \Big( \big( \widetilde{\operatorname{Log}}_{\varphi(\mathbf{P})}(\varphi(\mathbf{Q})) \big)^T (\varphi(\mathbf{P})^{-1})^T + \varphi(\mathbf{P})^{-1} \widetilde{\operatorname{Log}}_{\varphi(\mathbf{P})}(\varphi(\mathbf{Q})) \Big)_{\frac{1}{2}}, \\ & \varphi(\mathbf{P}) \big( \varphi(\mathbf{P})^{-1} \mathbf{W} (\varphi(\mathbf{P})^{-1})^T \big)_{\frac{1}{2}} \rangle_{\varphi(\mathbf{P})} \\ &= \langle \varphi(\mathbf{P}) \varphi(\mathbf{P})^{-1} \widetilde{\operatorname{Log}}_{\varphi(\mathbf{P})}(\varphi(\mathbf{Q})), \varphi(\mathbf{P}) \big( \varphi(\mathbf{P})^{-1} \mathbf{W} (\varphi(\mathbf{P})^{-1})^T \big)_{\frac{1}{2}} \rangle_{\varphi(\mathbf{P})} \\ &= \langle \widetilde{\operatorname{Log}}_{\varphi(\mathbf{P})}(\varphi(\mathbf{Q})), \varphi(\mathbf{P}) \big( \varphi(\mathbf{P})^{-1} \mathbf{W} (\varphi(\mathbf{P})^{-1})^T \big)_{\frac{1}{2}} \rangle_{\varphi(\mathbf{P})}.   
\end{split}
\end{align*}

Let $\widetilde{\mathbf{W}} = \varphi(\mathbf{P}) \big( \varphi(\mathbf{P})^{-1} \mathbf{W} (\varphi(\mathbf{P})^{-1})^T \big)_{\frac{1}{2}}$. Then
\begin{align*}
\begin{split}
\langle \operatorname{Log}^{lc}_{\mathbf{P}}(\mathbf{Q}), \mathbf{W} \rangle_{\mathbf{P}} &= \langle \widetilde{\operatorname{Log}}_{\varphi(\mathbf{P})}(\varphi(\mathbf{Q})), \widetilde{\mathbf{W}} \rangle_{\varphi(\mathbf{P})} \\ &= \langle \lfloor \varphi(\mathbf{Q}) \rfloor - \lfloor \varphi(\mathbf{P}) \rfloor + \mathbb{D}(\varphi(\mathbf{P})) \log( \mathbb{D}(\varphi(\mathbf{P}))^{-1} \mathbb{D}(\varphi(\mathbf{Q})) ), \widetilde{\mathbf{W}} ) \rangle_{\varphi(\mathbf{P})} \\ &= \langle \lfloor \varphi(\mathbf{Q}) \rfloor - \lfloor \varphi(\mathbf{P}) \rfloor + \log( \mathbb{D}(\varphi(\mathbf{P}))^{-1} \mathbb{D}(\varphi(\mathbf{Q})) ), \lfloor \widetilde{\mathbf{W}} \rfloor + \mathbb{D}(\varphi(\mathbf{P}))^{-1} \mathbb{D}(\widetilde{\mathbf{W}}) \rangle_F.   
\end{split}
\end{align*}

Thus, for $\mathbf{Q} \in \mathcal{H}_{\mathbf{W},\mathbf{P}}$, we have
\begin{equation*}
\langle \lfloor \varphi(\mathbf{Q}) \rfloor - \lfloor \varphi(\mathbf{P}) \rfloor + \log( \mathbb{D}(\varphi(\mathbf{P}))^{-1} \mathbb{D}(\varphi(\mathbf{Q})) ), \lfloor \widetilde{\mathbf{W}} \rfloor + \mathbb{D}(\varphi(\mathbf{P}))^{-1} \mathbb{D}(\widetilde{\mathbf{W}}) \rangle_F = 0.  
\end{equation*}

The SPD gyrodistance $d(\mathbf{X},\mathcal{H}_{\mathbf{W},\mathbf{P}})$ is therefore given by
\begin{equation*}
d(\mathbf{X},\mathcal{H}_{\mathbf{W},\mathbf{P}}) = \frac{ | \langle -\lfloor \varphi(\mathbf{P}) \rfloor + \lfloor \varphi(\mathbf{X}) \rfloor + \log( \mathbb{D}(\varphi(\mathbf{P}))^{-1} \mathbb{D}(\varphi(\mathbf{X})) ) , \lfloor \widetilde{\mathbf{W}} \rfloor + \mathbb{D}(\varphi(\mathbf{P}))^{-1} \mathbb{D}(\widetilde{\mathbf{W}}) \rangle_F | }{ \| \lfloor \widetilde{\mathbf{W}} \rfloor + \mathbb{D}(\varphi(\mathbf{P}))^{-1} \mathbb{D}(\widetilde{\mathbf{W}}) \|_F },  
\end{equation*}
where $\widetilde{\mathbf{W}} = \varphi(\mathbf{P}) \big( \varphi(\mathbf{P})^{-1} \mathbf{W} (\varphi(\mathbf{P})^{-1})^T \big)_{\frac{1}{2}}$.


\end{proof}

\section{Proof of Theorem~\ref{theorem:distance_to_SPD_hyperplanes_affine_invariant}}
\label{sec:appendix_distance_to_SPD_hyperplanes_affine_invariant}

\begin{proof}

Note that
\begin{align}
\begin{split}
\langle \operatorname{Log}^{ai}_{\mathbf{P}}(\mathbf{Q}), \mathbf{W} \rangle_{\mathbf{P}} &\overset{(1)}{=} \langle \mathbf{P}^{-\frac{1}{2}} \operatorname{Log}^{ai}_{\mathbf{P}}(\mathbf{Q}) \mathbf{P}^{-\frac{1}{2}}, \mathbf{P}^{-\frac{1}{2}} \mathbf{W} \mathbf{P}^{-\frac{1}{2}} \rangle_F \\ &\overset{(2)}{=} \langle \mathbf{P}^{-\frac{1}{2}} \mathbf{P}^{\frac{1}{2}} \log(\mathbf{P}^{-\frac{1}{2}} \mathbf{Q} \mathbf{P}^{-\frac{1}{2}}) \mathbf{P}^{\frac{1}{2}} \mathbf{P}^{-\frac{1}{2}}, \mathbf{P}^{-\frac{1}{2}} \mathbf{W} \mathbf{P}^{-\frac{1}{2}} \rangle_F \\ &= \langle \log(\mathbf{P}^{-\frac{1}{2}} \mathbf{Q} \mathbf{P}^{-\frac{1}{2}}), \mathbf{P}^{-\frac{1}{2}} \mathbf{W} \mathbf{P}^{-\frac{1}{2}} \rangle_F,  
\end{split}
\end{align}
where (1) follows from the definition of Affine-Invariant metrics, 
and (2) follows from the expression of the exponential map associated with Affine-Invariant metrics.  

Thus, for $\mathbf{Q} \in \mathcal{H}_{\mathbf{W},\mathbf{P}}$, we have
\begin{equation}\label{eq:distance_spd_hyperplane_condition_ai}
\langle \log(\mathbf{P}^{-\frac{1}{2}} \mathbf{Q} \mathbf{P}^{-\frac{1}{2}}), \mathbf{P}^{-\frac{1}{2}} \mathbf{W} \mathbf{P}^{-\frac{1}{2}} \rangle_F = 0.  
\end{equation}

By the definition of the SPD pseudo-gyrodistance,
\begin{equation*}
\bar{d}(\mathbf{X},\mathcal{H}_{\mathbf{W},\mathbf{P}}) = \sin(\angle \mathbf{X} \mathbf{P} \bar{\mathbf{Q}})d(\mathbf{X},\mathbf{P}),
\end{equation*}
where
\begin{align*}
\begin{split}
\bar{\mathbf{Q}} &= \argmax_{\mathbf{Q} \in \mathcal{H}_{\mathbf{W},\mathbf{P}} \setminus \{ \mathbf{P} \}} \frac{ \langle \ominus_{ai} \mathbf{P} \oplus_{ai} \mathbf{Q}, \ominus_{ai} \mathbf{P} \oplus_{ai} \mathbf{X} \rangle }{ \| \ominus_{ai} \mathbf{P} \oplus_{ai} \mathbf{Q} \|. \| \ominus_{ai} \mathbf{P} \oplus_{ai} \mathbf{X} \| } \\ &\overset{(1)}{=} \argmax_{\mathbf{Q} \in \mathcal{H}_{\mathbf{W},\mathbf{P}} \setminus \{ \mathbf{P} \}} \frac{ \langle \operatorname{Log}^{ai}_{\mathbf{I}_n}(\ominus_{ai} \mathbf{P} \oplus_{ai} \mathbf{Q}), \operatorname{Log}^{ai}_{\mathbf{I}_n}(\ominus_{ai} \mathbf{P} \oplus_{ai} \mathbf{X}) \rangle_F }{ \| \operatorname{Log}^{ai}_{\mathbf{I}_n}(\ominus_{ai} \mathbf{P} \oplus_{ai} \mathbf{Q}) \|_F . \| \operatorname{Log}^{ai}_{\mathbf{I}_n}(\ominus_{ai} \mathbf{P} \oplus_{ai} \mathbf{X}) \|_F } \\ &= \argmax_{\mathbf{Q} \in \mathcal{H}_{\mathbf{W},\mathbf{P}} \setminus \{ \mathbf{P} \}} \frac{ \langle \log(\mathbf{P}^{-\frac{1}{2}} \mathbf{Q} \mathbf{P}^{-\frac{1}{2}}), \log(\mathbf{P}^{-\frac{1}{2}} \mathbf{X} \mathbf{P}^{-\frac{1}{2}}) \rangle_F }{ \| \log(\mathbf{P}^{-\frac{1}{2}} \mathbf{Q} \mathbf{P}^{-\frac{1}{2}}) \|_F . \| \log(\mathbf{P}^{-\frac{1}{2}} \mathbf{X} \mathbf{P}^{-\frac{1}{2}}) \mathbf{X} \|_F } 
\end{split}
\end{align*}

Our problem becomes the one of finding the minimum angle between 
the vector $\log(\mathbf{P}^{-\frac{1}{2}} \mathbf{X} \mathbf{P}^{-\frac{1}{2}})$ 
and the Euclidean hyperplane described by Eq.~(\ref{eq:distance_spd_hyperplane_condition_ai}).
Therefore, the SPD pseudo-gyrodistance $\bar{d}(\mathbf{X},\mathcal{H}_{\mathbf{W},\mathbf{P}})$ is given by
\begin{equation*}
\bar{d}(\mathbf{X},\mathcal{H}_{\mathbf{W},\mathbf{P}}) = \frac{| \langle \log(\mathbf{P}^{-\frac{1}{2}} \mathbf{X} \mathbf{P}^{-\frac{1}{2}}), \mathbf{P}^{-\frac{1}{2}} \mathbf{W} \mathbf{P}^{-\frac{1}{2}} \rangle_F |}{ \| \mathbf{P}^{-\frac{1}{2}} \mathbf{W} \mathbf{P}^{-\frac{1}{2}} \|_F }.
\end{equation*}

\end{proof}

\section{Proof of Corollary~\ref{corollary:distance_set_log_euclidean}}
\label{sec:appendix_distance_set_log_euclidean}


\begin{proof}

We first recall a result from~\citet{pennec:tel-00633163}.  

\begin{proposition}\label{propo:differential_matrix_log}
Let $\mathbf{P} \in \operatorname{Sym}_n^+$ and $\mathbf{W} \in \mathcal{T}_{\mathbf{P}}\operatorname{Sym}_n^+$. 
Then
\begin{equation*}
D\log_{\mathbf{P}}(\mathbf{W}) = \mathbf{O} ( D\log_{\pmb{\Sigma}}(\mathbf{O}^T \mathbf{W} \mathbf{O}) ) \mathbf{O}^T,
\end{equation*}
where $\mathbf{P} = \mathbf{O} \pmb{\Sigma} \mathbf{O}^T$ 
is the eigenvalue decomposition of $\mathbf{P}$.  
\end{proposition}

First, it is easy to see that
\begin{equation*}
\log(\mathbf{X}) = \diag(\log(\mathbf{X}_1),\ldots,\log(\mathbf{X}_N)) \text{ , } \log(\mathbf{P}) = \diag(\log(\mathbf{P}_1),\ldots,\log(\mathbf{P}_N)).
\end{equation*}

Let $\mathbf{P}_i = \mathbf{O}_i \pmb{\Sigma}_i \mathbf{O}_i^T$ be the eigenvalue decomposition of $\mathbf{P}_i,i=1,\ldots,N$. 
Then $\mathbf{P} = \mathbf{O} \pmb{\Sigma} \mathbf{O}^T$ where 
$\mathbf{O} = \diag(\mathbf{O}_1,\ldots,\mathbf{O}_N)$, 
and $\pmb{\Sigma} = \diag(\pmb{\Sigma}_1,\ldots,\pmb{\Sigma}_N)$. 
Note that
\begin{align}\label{eq:differential_matrix_log_decompose}
\begin{split}
D\log_{\mathbf{P}}(\mathbf{W}) &\overset{(1)}{=} \mathbf{O} ( D\log_{\pmb{\Sigma}}(\mathbf{O}^T \mathbf{W} \mathbf{O}) ) \mathbf{O}^T \\ &= \mathbf{O} \Big( D\log_{\pmb{\Sigma}}\big( \diag(\mathbf{O}_1^T \mathbf{W}_1 \mathbf{O}_1,\ldots,\mathbf{O}_N^T \mathbf{W}_N \mathbf{O}_N) \big) \Big) \mathbf{O}^T \\ &\overset{(2)}{=} \mathbf{O} \Big( \diag\big( D\log_{\pmb{\Sigma}_1}(\mathbf{O}_1^T \mathbf{W}_1 \mathbf{O}_1),\ldots,D\log_{\pmb{\Sigma}_N}(\mathbf{O}_N^T \mathbf{W}_N \mathbf{O}_N) \big) \Big) \mathbf{O}^T \\ &= \diag\Big( \mathbf{O}_1 \big(D\log_{\pmb{\Sigma}_1}(\mathbf{O}_1^T \mathbf{W}_1 \mathbf{O}_1)\big) \mathbf{O}_1^T,\ldots, \mathbf{O}_N \big(D\log_{\pmb{\Sigma}_N}(\mathbf{O}_N^T \mathbf{W}_N \mathbf{O}_N) \big) \mathbf{O}_N^T \Big) \\ &\overset{(3)}{=} \diag\big(D\log_{\mathbf{P}_1}(\mathbf{W}_1),\ldots,D\log_{\mathbf{P}_N}(\mathbf{W}_N)\big).
\end{split}
\end{align}

The derivation of Eq.~(\ref{eq:differential_matrix_log_decompose}) follows. 

(1) follows from Proposition~\ref{propo:differential_matrix_log}.  

(2) follows from the property of $D\log_{\pmb{\Sigma}}(.)$.  

(3) follows from Proposition~\ref{propo:differential_matrix_log}.  

According to Theorem~\ref{theorem:distance_to_SPD_hyperplanes_log_euclidean}, the SPD gyrodistance from $\mathbf{X}$ 
to a hyperplanes $\mathcal{H}_{\mathbf{W},\mathbf{P}}$ in a gyrovector space $(\operatorname{Sym}_{n \times N}^+,\oplus_{le},\otimes_{le})$ 
is given by
\begin{align*}
\begin{split}
d(\mathbf{X},\mathcal{H}_{\mathbf{W},\mathbf{P}}) &= \frac{| \langle \log(\mathbf{X}) - \log(\mathbf{P}),D\log_{\mathbf{P}}(\mathbf{W}) \rangle_F |}{\| D\log_{\mathbf{P}}(\mathbf{W}) \|_F} \\ &= \frac{| \sum_{i=1}^N \langle \log(\mathbf{X}_i) - \log(\mathbf{P}_i),D\log_{\mathbf{P}_i}(\mathbf{W}_i) \rangle_F |}{ \sqrt{ \sum_{i=1}^N \| D\log_{\mathbf{P}_i}(\mathbf{W}_i) \|_F^2} }.  
\end{split}
\end{align*} 

Note that each term in the expressions of the SPD gyrodistance and pseudo-gyrodistance given in 
Theorems~\ref{theorem:distance_to_SPD_hyperplanes_log_cholesky} 
and~\ref{theorem:distance_to_SPD_hyperplanes_affine_invariant} is formed
from the operation $\diag(.)$ defined in Corollary~\ref{corollary:distance_set_log_euclidean}. 
Then the results for LC and AI gyrovector spaces can be easily proved.  

\end{proof}

\end{document}